\documentclass{zlab}

\usepackage[T1]{fontenc}
\usepackage{tgpagella}
\usepackage{mathpazo}
\usepackage{inconsolata}

\usepackage{xspace}
\usepackage{graphicx,amsmath,amssymb,hyperref}
\usepackage{etoc}
\usepackage[authoryear,round]{natbib}
\usepackage{xcolor}
\usepackage{multirow}
\usepackage{makecell}
\usepackage{afterpage}
\usepackage{float}
\usepackage{wrapfig}
\usepackage{comment}
\usepackage{fancyvrb}
\usepackage{wasysym}
\usepackage{array}
\usepackage{longtable}
\usepackage{caption}
\usepackage{subcaption}
\usepackage{tcolorbox}
\usepackage{tikz}
\usetikzlibrary{positioning,arrows.meta,shapes.geometric,fit,calc}
\usepackage[nameinlink]{cleveref}

\newcommand{\pl}{\textcolor{paperlensblue}{\textbf{PaperScorer}}}
\newcommand{\systemfamilyname}{\pl}
\newcommand{\systemtextname}{\pl-Text}
\newcommand{\systemvisionname}{\pl-Vision}
\newcommand{\dataseticlrname}{\textit{ICLR}}
\newcommand{\dataseticlrshort}{\textit{ICLR}}
\newcommand{\datasetarxivname}{\textit{arXiv}}
\newcommand{\datasetarxivmediumname}{\textit{arXiv-m}}
\newcommand{\datasetarxiviclrname}{\textit{arXiv-ICLR}}

\newcommand{\systemfamily}{\systemfamilyname}
\newcommand{\systemtext}{\systemtextname}
\newcommand{\systemvision}{\systemvisionname}
\newcommand{\systemname}{\systemfamilyname}

\newcommand{\dataseticlr}{\dataseticlrname}

\newcommand{\dataset}{\dataseticlr}

\newcommand{\appref}[1]{\hyperref[#1]{Appendix~\ref*{#1}}}

\definecolor{darkblue}{rgb}{0, 0, 0.5}
\definecolor{trainyellow}{RGB}{255, 245, 204}
\definecolor{gaingreen}{RGB}{0, 100, 0}
\definecolor{lossred}{RGB}{170, 0, 0}
\definecolor{takeawayborder}{HTML}{2F5496}
\definecolor{paperlensblue}{HTML}{2F5496}
\definecolor{mainresultshighlight}{HTML}{CDDFF5}
\definecolor{takeawayhead}{HTML}{D9E7FF}
\definecolor{takeawaybg}{RGB}{248, 251, 255}
\newtcolorbox{resulttakeaways}[1][Main Takeaways]{
    colback=takeawaybg,
    colframe=takeawayborder,
    colbacktitle=takeawayhead,
    coltitle=black,
    fonttitle=\bfseries,
    title=#1,
    arc=2mm,
    boxrule=0.8pt,
    left=1.4mm,
    right=1.4mm,
    top=1mm,
    bottom=1mm
}

\hypersetup{
  colorlinks=true,
  urlcolor=darkblue,
  linkcolor=darkblue,
  citecolor=darkblue,
  filecolor=black,
  pdfborder={0 0 0}
}

\definecolor{link}{HTML}{0063BE}
\providecommand{\equationname}{Equation}
\providecommand{\sectionname}{Section}

\newcommand{\figref}[1]{%
  \figurename~\hyperref[#1]{\textcolor{link}{\ref*{#1}}}%
}
\newcommand{\tabref}[1]{%
  \tablename~\hyperref[#1]{\textcolor{link}{\ref*{#1}}}%
}
\newcommand{\eqrefc}[1]{%
  \equationname~\hyperref[#1]{\textcolor{link}{\ref*{#1}}}%
}
\newcommand{\secrefc}[1]{%
  \sectionname~\hyperref[#1]{\textcolor{link}{\ref*{#1}}}%
}

\newcommand{\paperlenswebsiteurl}{https://sachinkonan.github.io/PaperScorer.html}
\newcommand{\paperlensgithuburl}{https://github.com/zlab-princeton/PaperScorer}
\newcommand{\paperlenshfurl}{https://huggingface.co/collections/zlab-princeton/paperscorer}
\newcommand{\github}{\raisebox{-1.3pt}{\includegraphics[height=1.05em]{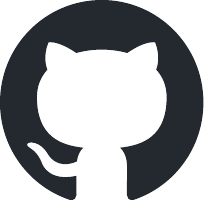}}\xspace}
\newcommand{\worldwideweb}{\raisebox{-1.3pt}{\includegraphics[height=1.05em]{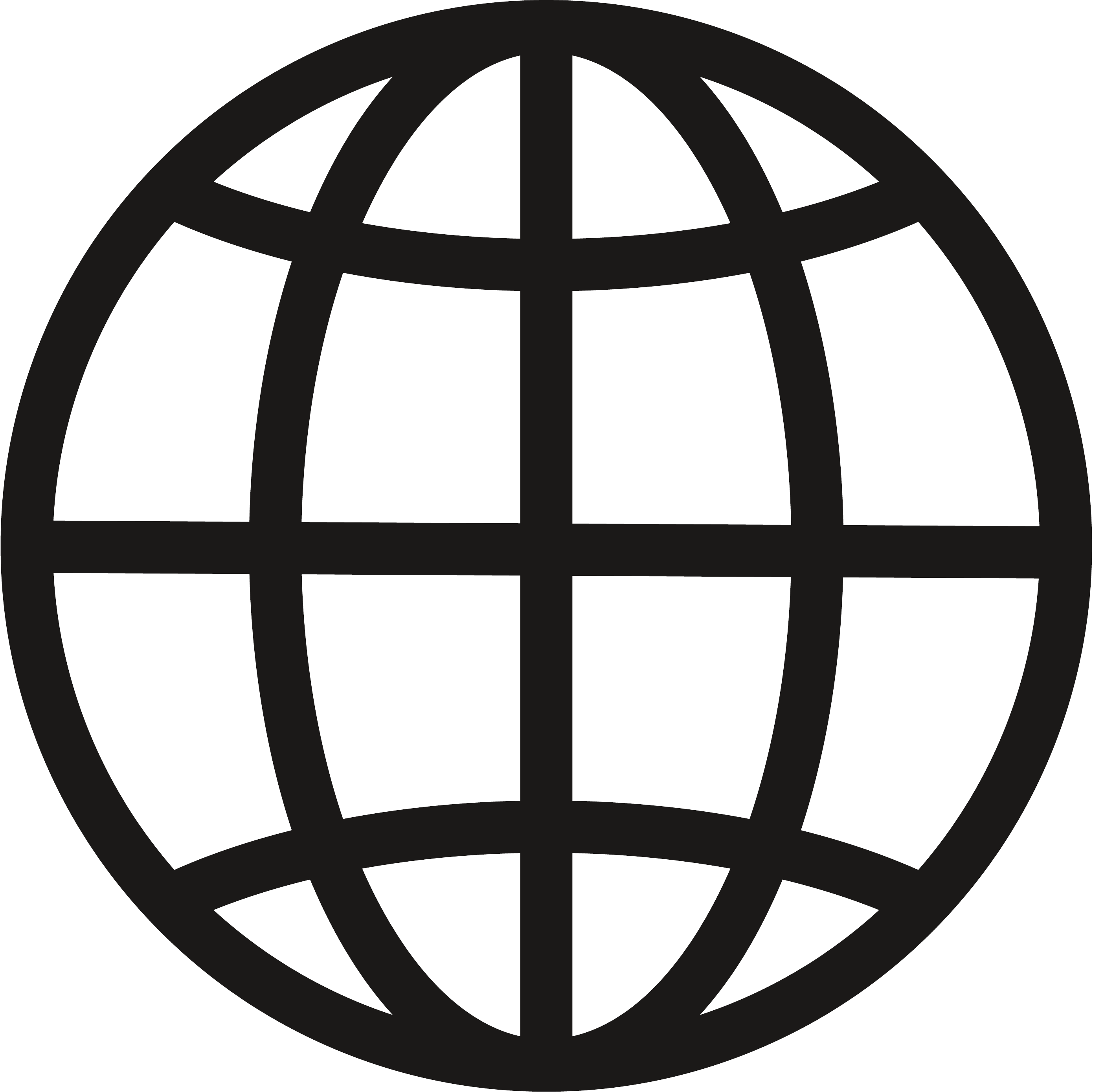}}\xspace}
\newcommand{\hf}{\raisebox{-1.3pt}{\includegraphics[height=1.05em]{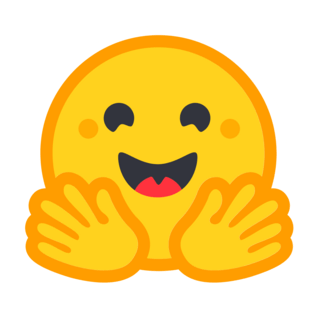}}\xspace}

\makeatletter
\fancypagestyle{firststyle}{
  \fancyhead[L]{}
  \fancyhead[C]{}
  \fancyhead[R]{}
  \fancyfoot[L]{\footerfont \the\correspondingauthor}
  \fancyfoot[R]{}
}
\makeatother

\DeclareTextFontCommand{\textbf}{\bfseries}
\title{\centering \fontsize{18}{24}\selectfont{\textcolor{paperlensblue}{PaperScorer}: How Predictable Is Paper Acceptance?}}

\author{
    \vspace{-.15cm}
    \parbox{\textwidth}{\centering
        Sachin Konan \quad
        Jonathan Liu \quad
        Zhuang Liu \quad
    }
    \\
    \vspace{.3cm}
    {\normalfont\fontsize{11}{15}\selectfont Princeton University\hspace{.1cm}}
}

\makeatletter
\renewcommand{\abscontent}{}%
\makeatother

\newenvironment{abstractblock}{%
  {\centering\large\bfseries Abstract\par}
  \vspace{0.2em}
  \begin{list}{}{%
      \setlength{\leftmargin}{2em}
      \setlength{\rightmargin}{2em}
      \setlength{\topsep}{0pt}
      \setlength{\parsep}{0pt}
  }
  \item[]
}{%
  \end{list}
  \par\normalfont\vspace{1em}
}

\newcommand{\abstractcontent}{%
\looseness=-1
Automated research is largely evaluated by benchmark performance, but the value of research rarely reduces to a single score. Humans judge a work's holistic contribution through peer review, which is hard to model because it is noisy and depends on many factors. However, what if, given only the paper, we could predict its peer-review decision directly? We introduce \systemfamily{}, a family of models that predict a paper's accept or reject decision from its extracted text or page screenshots, training only on the decision after removing author identity and formatting shortcuts. Vision models are the most predictive (68\% accuracy on ICLR papers and 77\% on ML papers from arXiv), beating text-only and frontier reasoning models. Unlike plain text, page images retain figures, layout, and presentation. This signal is robust, surviving smaller models and even compressing the whole paper into a single image. But accurate predictions are not evenly spread. \systemfamily{} is strongest on the lowest and highest rated papers, while papers rated in the middle stay unpredictable for our models and frontier systems alike. Additionally, its confidence is calibrated, so its most confident predictions are also its most accurate. Overall, \systemfamily{} shows how much of the decision a paper carries on its own, marking a first step toward AI-driven research evaluation.
}

\begin{document}

\begingroup
\makeatletter
\let\raggedright\centering
\makeatother
\vspace*{-0.9cm}
\maketitle
\endgroup

\setlength{\skip\footins}{1cm}

\vspace{-0.72cm}
{\centering\small
    \worldwideweb \href{\paperlenswebsiteurl}{\textbf{Website}} \quad
    \github \href{\paperlensgithuburl}{\textbf{Code}} \quad
    \hf \href{\paperlenshfurl}{\textbf{Hugging Face}}
\par}

\vspace{-0.28cm}

\begin{figure}[!ht]
    \centering
    \subcaptionbox{\systemfamily{} predicts a paper's decision from its text or visual view.\label{fig:teaser_flow}}[\textwidth]{%
        \includegraphics[width=\textwidth]{"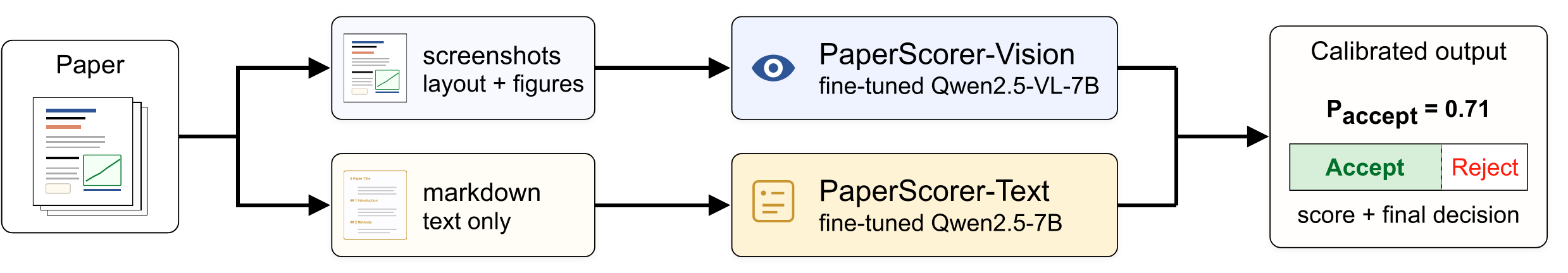"}}

    \vspace{0.6em}

    \subcaptionbox{Two anonymized corpora.\label{fig:teaser_corpora}}[0.286\textwidth]{%
        \includegraphics[width=0.93\linewidth]{"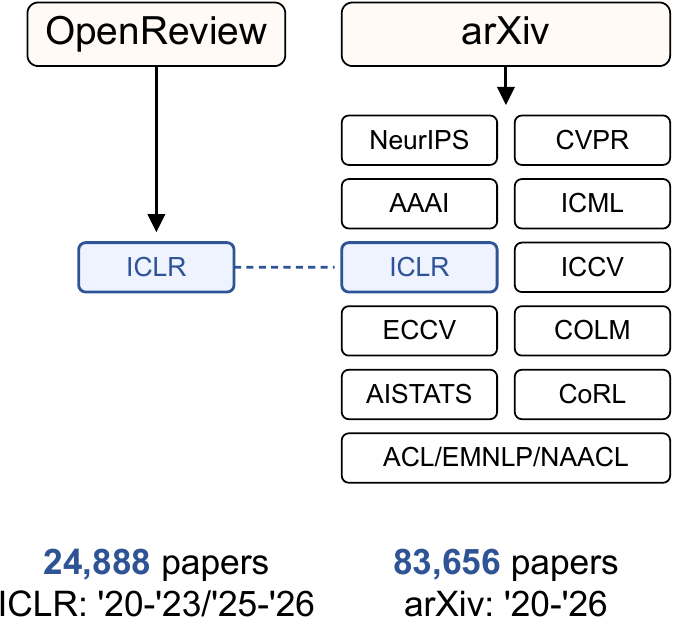"}}%
    \hfill
    \subcaptionbox{\systemfamily{} is an accurate decision predictor.\label{fig:teaser_predictor}}[0.66\textwidth]{%
        \includegraphics[width=0.93\linewidth]{"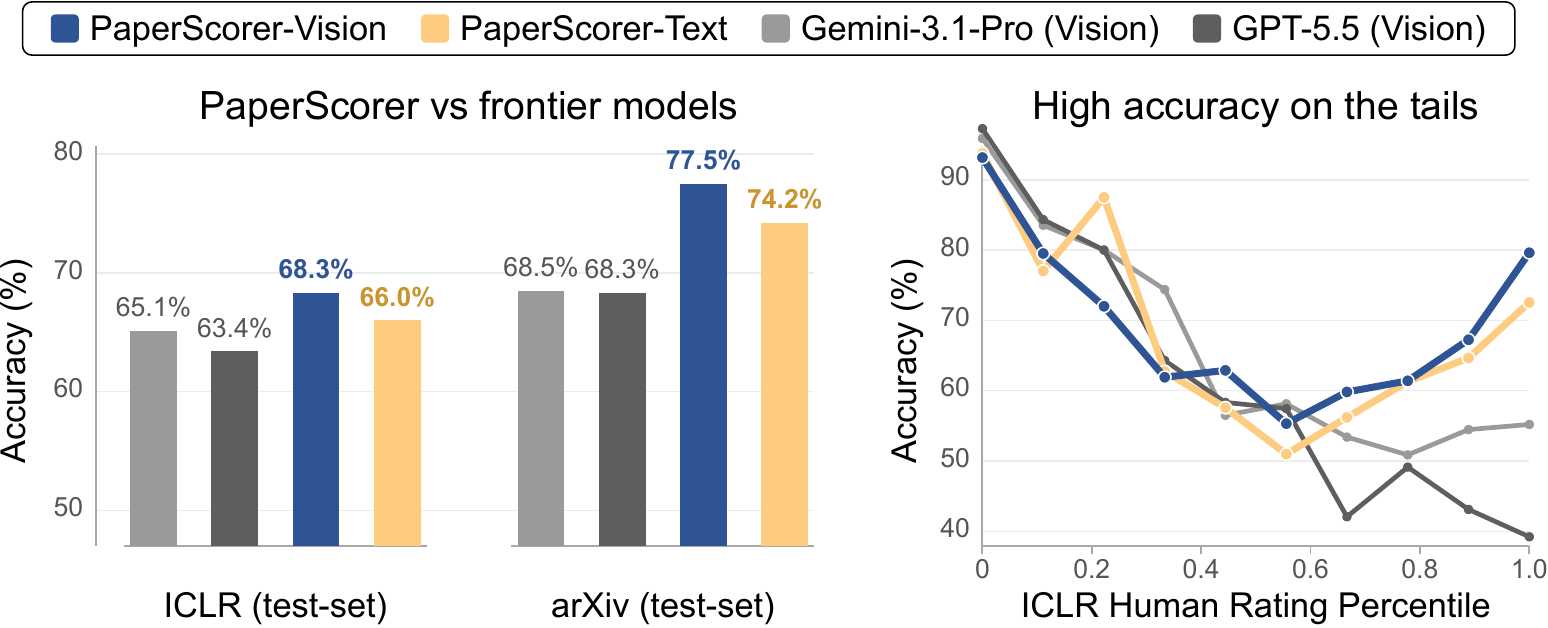"}}
    \caption{\textbf{(a)}~\systemfamily{} is a family of models trained to predict a paper's accept/reject decision from either a pure text or a visual view of the paper. \textbf{(b)}~For training and evaluation, we built two fully anonymized datasets, sourced from OpenReview and arXiv and labeled with accept/reject decisions. \textbf{(c)}~Across training recipes, we find fine-tuning on our large-scale corpora enables \systemfamily{} to exceed frontier models (GPT/Gemini), specifically on highly rated papers.}
    \label{fig:teaser}
\end{figure}

\vspace{-0.05cm}

\begin{abstractblock}
\abstractcontent
\end{abstractblock}

\newpage

\section{Introduction}
\label{sec:intro}

Increasingly capable large language model (LLM) agents are beginning to automate research end to end, proposing ideas, running experiments, and interpreting the results to recursively self-improve~\citep{lu2024ai,yamada2025ai,si2026towards,favaro2026recursive}. This loop needs a way to judge whether a result is good, and the standard judge is a benchmark. Optimizing benchmarks has driven much of the field's progress, from ImageNet in computer vision~\citep{deng2009imagenet} to MuJoCo in reinforcement learning~\citep{todorov2012mujoco}. Yet many contributions, such as findings or interpretability work, have no benchmark, and even where they exist, researchers disagree about which ones matter. This necessitates peer review.

However, peer review is noisy and poorly understood. A paper is typically judged by three or four reviewers and a meta-reviewer, whose assessments are combined into a final accept or reject decision \citep{yacoby_paper_review_process}. In the NeurIPS '14 and '21 consistency experiments, roughly half of papers accepted by one committee would not have been accepted by the other~\citep{beygelzimer2021consistency}. AI tools now add to this complexity~\citep{biswas2026ai}, as up to \textbf{17\%} of reviews were materially edited by LLMs~\citep{liang2024monitoring}, and LLM-generated feedback led \textbf{27\%} of treated reviewers to revise their reviews~\citep{thakkar2026llmfeedback}. A final decision therefore mixes \textit{paper signal} with \textit{noise} from reviewer assignment, interpretation, aggregation, and now the AI tools involved. Before automated systems can judge research the way peer review does, we need to know how much of the decision the paper itself determines. This motivates a simple question:
\smallskip
\begin{center}
\emph{From paper content alone, how predictable is the final decision?}
\end{center}
\smallskip

Measuring this signal requires more than fitting accept/reject labels. Without careful controls, a predictor can rely on shortcut signals rather than paper content \citep{zhang2025re}. Camera-ready formatting, author-identifying text, source-specific artifacts, and the natural majority-reject rate can all leak the label \citep{jin2024agentreview}. Even after removing shortcuts, final decisions remain noisy, so a useful model should report uncertainty, separating confident cases from papers whose outcomes are not predictable from content alone.

We introduce \systemfamily{}, a family of models that predicts binary acceptance decisions from anonymized papers. \systemfamily{} is trained on final outcomes rather than intermediate reviews or ratings, using anonymized datasets of machine learning (ML) papers. To test whether extracted text is sufficient, we compare two representations of each paper, markdown for text models and screenshots of each page for vision models. The text representation contains all text, including formulas and tables, but excludes figures, while vision preserves the full document. Our datasets contain fully anonymized papers from \dataseticlrname{}, which are openly hosted and shared on OpenReview, and from ML conferences on \datasetarxivname{}, all from '20--'26 and balanced 50/50 accept/reject to avoid a majority-reject shortcut.

Our main empirical result is that paper content contains recoverable decision signal, but this signal is limited. Our top vision model, \systemvision{}, achieves \textbf{68\%} accuracy on \dataseticlrname{}, improving over the best frontier baseline by \textbf{2.7\%}. On \datasetarxivname{}, \datasetarxivname{}-trained \systemvision{} achieves \textbf{77\%} accuracy, improving over frontier baselines by \textbf{9.0\%}. Beyond binary decisions, \systemfamily{} emits a continuous acceptance estimate. This score ranks accepted above rejected papers across thresholds, and on \dataseticlrname{}, it aligns with reviewer ratings better than reviewer-trained baselines. After calibration on validation data, confidence matches accuracy on our test datasets, so at $X\%$ confidence, accuracy is about $X\%$.

Our experiments isolate several factors underlying performance. First, \systemfamily{} achieves significantly higher accuracy with vision than with text on both datasets. Vision enables analysis of figures, layout, and full-document presentation, signal that our text extraction discards. Second, predictability is concentrated at the extremes of reviewer rating, but not symmetrically. Frontier models reliably flag only the lowest-rated papers, whereas \systemfamily{} is accurate on both low and high-rated papers. Across all evaluated models, however, accuracy remains near chance on papers with intermediate reviewer ratings, so for these papers the document alone gives our models little to go on. Third, performance improves with data scale but is largely invariant to model size or input resolution. A 3B model performs similarly to a 7B model, and compressing the whole paper into a single image only marginally degrades performance.

In summary, \systemfamily{} is the strongest paper decision predictor in our study, because a large, carefully controlled dataset lets straightforward supervised fine-tuning deliver higher accuracy and calibrated confidence, concentrated on the clear accepts and rejects. This matters because automated research makes research faster to iterate on, so submission volumes~\citep{su2025papercopilot} and reviewer burden~\citep{adam2025peerreviewcrisis} rise, while benchmark scores tell only part of a contribution's story, which means evaluation must scale in a way that reflects human judgment. At only 7B parameters, \systemfamily{} is a first step that measures how much of the human decision the paper alone carries, shows through calibration when to trust that estimate, and already plugs into agentic reviewing and literature search (\appref{sec:applications}).

    \section{Problem Formulation}
    \label{sec:probform}
    We study \textbf{content-only paper acceptance prediction}. Let $\mathcal{X}$ denote the space of observed paper artifacts and $\mathcal{Z}=\{0,1\}$ the final binary decision, with $z=1$ for acceptance and $z=0$ for rejection. Our goal is to learn a predictor $f_\theta(x)$ that returns, from the paper artifact alone, both a hard decision and a score whose higher values indicate a higher predicted probability of acceptance.

    \textbf{Peer review's decision process.} We view peer review as a latent chain where $X_0$ is the initial submission, $Y_1$ the initial reviews, $Y_2$ the rebuttal process, $X_{\mathrm{reb}}$ the revised paper, $R$ the revised review after rebuttal, $Y_3$ the meta-review, and $Z$ the final decision.
    \begin{equation*}
        X_0 \rightarrow Y_1 \rightarrow Y_2 \rightarrow X_{\mathrm{reb}} \rightarrow R \rightarrow Y_3 \rightarrow Z
    \end{equation*}
    Because public anonymous artifacts are most consistently available after rebuttal (\appref{appendix:dataset-notes} documents year-over-year availability), our practical task is $p(Z \mid X_{\mathrm{reb}})$ rather than $p(Z \mid X_0)$. In contrast with prior work~\citep{tyser2024reviewer2, liu2025deepreview}, reviews, ratings, and meta-reviews are \emph{not regressed on}.

    \textbf{Prediction target.} We frame this as a text-generation problem with a decision-only target, so prompting, SFT, and RL can all be studied under the same format.
    \begin{equation*}
        f_\theta(x) \rightarrow \boxed{\texttt{Accept}\mid\texttt{Reject}}
    \end{equation*}
    When token probabilities are available, we also compute an acceptance score $p_{\mathrm{accept}}(x)$ by normalizing the probabilities assigned to the two decision tokens. The target dataset is therefore a collection $\mathcal{D}=\{(x_i, z_i)\}_{i=1}^{N}$ of paper-decision pairs, where $x_i$ is derived from paper content alone and $z_i$ is the final conference decision.

    \textbf{Estimated Bayesian ceiling.} As the NeurIPS '14 / '21 consistency experiments~\citep{beygelzimer2021consistency} show, the decision carries some amount of randomness. To get a rough sense of where the ceiling on accuracy lies given this noise, we fit a Beta-Bernoulli model loosely to this single historical experiment. Let $p_i \in [0,1]$ denote the latent acceptance probability of paper $i$, with committee decisions drawn from independent Bernoulli distributions, implying the following values.
    \begin{equation*}
        \mathbb{E}[p]=0.26,\qquad
        \mathbb{E}[2p(1-p)]=0.23,\qquad
        \mathbb{E}[p^2]=\frac{0.52-0.23}{2}=0.145.
    \end{equation*}
    Matching these moments gives $p \sim \mathrm{Beta}(\alpha,\beta)$ with $(\alpha,\beta)\approx(0.39,1.10)$. Under this fitted model, the Bayes-optimal accuracy is approximately 83.3\% under the original acceptance distribution and 80.2\% on a balanced accept/reject evaluation. We use \textbf{80.2\%} only as a soft reference point, not a hard limit, since this estimate is loose and inherits the assumptions of one historical experiment. Full derivation is in \appref{appendix:bayes}.
    \section{PaperScorer: Dataset and Methods}

    This section describes the balanced \dataseticlrname{} and source-derived \datasetarxivname{} corpora
    with paired markdown and page screenshots, and the training recipes behind \systemfamily{}.

    \subsection{Datasets}

\begin{figure*}[h!]
        \centering
        \includegraphics[width=\textwidth]{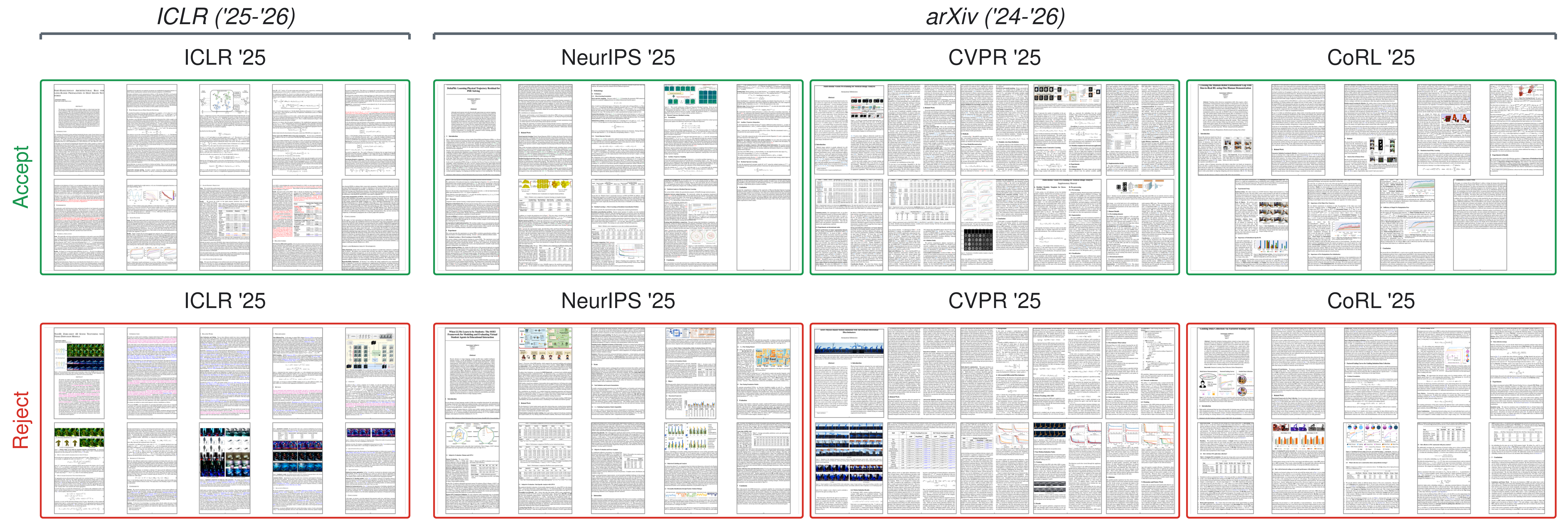}
        \vspace{-0.5em}
        \caption{\textbf{Dataset mosaic.} Screenshots of pages from \dataseticlrname{} and
        \datasetarxivname{}. \dataseticlrshort{} examples use latest-rebuttal OpenReview PDFs, while
        \datasetarxivname{} examples are derived from LaTeX source, recompiled with venue-specific
        anonymization macros. Each example tiles multiple pages for visualization; in practice,
        \systemvision{} processes individual pages. Notably, \dataseticlrshort{} papers can contain
        rebuttal-era revisions (colored text), whereas source-rendered \datasetarxivname{} papers do not.}
        \label{fig:dataset_mosaic}
        \vspace{-0.8em}
    \end{figure*}

    \systemfamily{} uses two balanced, anonymous accept/reject sources, both split 85/5/10
    across train, validation, and test, and we evaluate on the most recent years of each test split. \dataseticlrname{} contains 25K submissions from OpenReview,
    evaluated on '25--'26. \datasetarxivname{}
    contains LaTeX-source-derived proceedings papers, evaluated on '24--'26 to include
    biannual venues like ECCV. \datasetarxivname{} training uses
    a large 83K set (\datasetarxivname{}) and a smaller 25K set proportional to \dataseticlrshort{}
    (\datasetarxivmediumname{}). All datasets are balanced 50/50, and \Cref{fig:dataset_mosaic} shows examples from both.

    A paper-decision dataset is useful only if the model sees the same evidence that a final
    decision maker could have seen. We therefore require three properties before a paper
    enters the corpus:
    \begin{enumerate}[leftmargin=1.5em,itemsep=0.08em,topsep=0.2em]
        \item \textbf{Anonymity.} The exported artifact should not reveal
        authors, affiliations, acknowledgments, source repositories, or camera-ready status
        that would leak the final review outcome to the model.
        \item \textbf{Main-conference scope.} Papers must be full main-conference
        submissions, excluding workshops, short papers, industry tracks, withdrawn records,
        and other items outside the target decision process.
        \item \textbf{Consistent processing of accepts and rejects.} Accept and reject artifacts should differ
        through natural paper content, not through formatting shortcuts, rendering,
        or availability cues.
    \end{enumerate}


    \textbf{Data sourcing.}
    We build \dataseticlrshort{} from OpenReview because ICLR releases both accepted and
    rejected submissions with review metadata. The challenge is obtaining the anonymized
    artifact rather than a post-review PDF. Across six recent years, latest-rebuttal PDFs
    give the best availability for anonymous artifacts (\appref{appendix:dataset-notes}), although
    they can contain rebuttal-era revisions in the artifact itself.
    We therefore use ICLR '20--'23, '25, and '26 and exclude ICLR '24, where accepts are only camera-ready.

    \begin{figure*}[h!]
        \centering
        \includegraphics[width=0.88\textwidth]{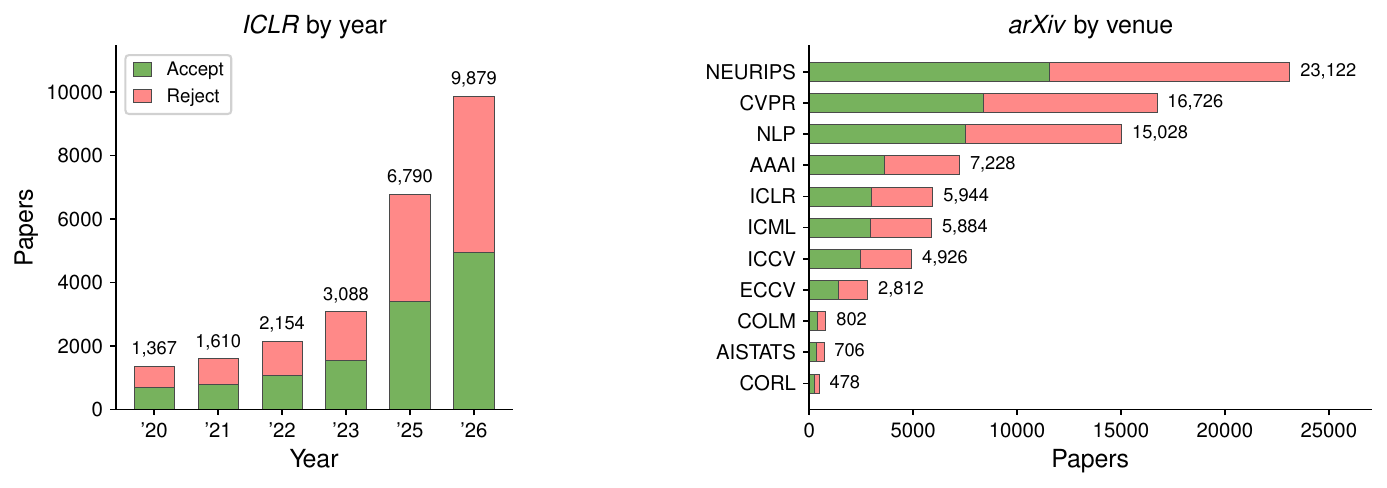}
        \vspace{-0.45em}
        \caption{\textbf{Dataset coverage.} We summarize the balanced \systemfamily{} pools by \dataseticlrshort{} year and \datasetarxivname{} venue after anonymous recompilation and filtering. \dataseticlrshort{} coverage leans toward later years, while \datasetarxivname{} coverage is largest for NeurIPS, CVPR, and ACL-style NLP venues, with smaller balanced slices from other ML conferences.}
        \label{fig:dataset_overview}
        \vspace{-0.85em}
    \end{figure*}

    ICLR is only one machine-learning venue, so we also build a LaTeX-source derived \datasetarxivname{} corpus
    for broader training and evaluation. \datasetarxivname{} does not provide reject labels, but title
    matches to conference proceedings provide high-precision accepted papers
    \citep{su2025papercopilot}. We therefore define an \emph{implied reject} as a
    LaTeX-compiling paper that uses a covered venue template, satisfies the main-conference
    filters, and does not appear in that venue's proceedings. This gives a noisier but more diverse source
    whose accept class is high fidelity and whose reject class tests whether the learned
    decision boundary survives weak negative labels.

    We use \datasetarxivname{} as a testbed for two transfer questions. First, within \datasetarxivname{}, we ask whether
    diverse supervision improves classification on held-out \datasetarxivname{} papers, paying particular
    attention to accept recall because accepted papers have the strongest provenance. Second,
    across sources, we compare \datasetarxivname{}-trained models on the high-fidelity \dataseticlrshort{} set and
    \dataseticlrshort{}-trained models on \datasetarxivname{}. This isolates whether diversity improves generalization or
    whether clean same-source labels remain more valuable than a larger weakly labeled pool.

    \vspace{0.05em}
    \textbf{Data filtering and normalization.}
    \dataseticlrshort{} is primarily filtered by year and artifact availability, then balanced
    within OpenReview year. \datasetarxivname{} requires more reconstruction. Starting from
    434,579 ML \datasetarxivname{} papers, we pull 434,443 LaTeX source tarballs in bulk from
    arXiv's requester-pays S3 feed and extract their \texttt{.tex} contents with NeMo
    Curator's \texttt{ArxivIterator}/\texttt{ArxivExtractor}~\citep{nemo_curator}.
    We then detect venue style files, compile anonymous review-mode PDFs, and remove residual identifiers. We drop
    non-main-conference, workshop, withdrawn, industry-track, and short-format papers. For
    \datasetarxivname{}, we additionally require a successful anonymous render, a references section,
    and between six and 30 main content pages to ensure consistency across papers. Full pipeline
    details for both sources are documented in \appref{appendix:data_pipeline}.

    The coverage plot in \Cref{fig:dataset_overview} reflects this sampling choice. \dataseticlrshort{}
    starts with 40K ICLR submissions and subsamples rejects to 25K while keeping the annual
    distribution. For \datasetarxivname{}, submission year is more nebulous: implied rejects have
    upload dates, while proceedings-matched accepts have conference years and post-review uploads.
    We therefore do not year-balance \datasetarxivname{}, and instead subsample 101,659 filtered
    papers to an 83,656 balanced set that keeps the starting venue distribution.

    Both sources pass through the same MinerU \citep{wang2024mineru} normalization before export. We
    remove headers, line-numbering asides, author-identifying spans, and all content in and after
    the references. Text models receive markdown with excluded regions omitted, while vision models
    receive page screenshots with the same regions whitened. Text inputs use the full markdown up to a
    pre-defined $24{,}480$ max context length, while for vision, each page is rendered and downscaled to at most $1{,}003{,}520$ pixels
    ($1280{\times}28^2$), roughly $1{,}240$ vision tokens. A typical paper runs about ten pages (median
    $10$, up to $33$), so the vision input averages roughly $12{,}420$ tokens, about half the max context length. Accepted papers still contain more content, figures, and citations after cleanup, reflecting differences in content rather than artifact mismatches.

    \begin{wraptable}[9]{r}{0.40\textwidth}
    \vspace{-1.2em}
    \centering
    \footnotesize
    \providecommand{\modtext}{\textcolor{black!50}{\scriptsize\textsc{Text}}}
    \providecommand{\modvision}{\textbf{\scriptsize\textsc{Vision}}}
    \setlength{\tabcolsep}{4.0pt}
    \renewcommand{\arraystretch}{1.04}
    \begin{tabular}{@{}lcccc@{}}
        \toprule
        & \multicolumn{2}{c}{\textbf{\dataseticlrshort{} (test)}} & \multicolumn{2}{c}{\textbf{\datasetarxivname{} (test)}} \\
        \cmidrule(lr){2-3} \cmidrule(l){4-5}
        \textbf{Train Setting} & \modvision & \modtext & \modvision & \modtext \\
        \midrule
        Natural train & -9.2 & -34.5 & -43.0 & -49.8 \\
        + val. thresh. & -6.0 & +27.9 & -10.8 & +8.7 \\
        \rowcolor{trainyellow} \textbf{Balanced train} & \textbf{+1.8} & \textbf{-6.6} & \textbf{-3.2} & \textbf{-1.6} \\
        \bottomrule
    \end{tabular}
    \vspace{0.35em}
    \caption{\textbf{Training-prior bias.} Accept$-$reject recall gap after each prior/threshold rule; lower absolute value is better.}
    \label{tab:training_prior_discrepancy}
    \vspace{-1.1em}
\end{wraptable}

    \noindent\textbf{Why balanced training and evaluation?}\label{sec:balanced_training_evaluation}\label{sec:balanced_evaluation}\label{sec:ratio_diagnostics}
    Conference decisions are naturally imbalanced, so raw accuracy can reward a predictor that
    mostly rejects papers. Therefore, we use a balanced test set, whereby accuracy equally weights
    accept and reject recall. This still supports natural-rate interpretation: for samples drawn
    from the same curated domain, class-conditional recalls from a balanced test can be reweighted
    by the known accept prior. In \appref{appendix:prior-diagnostics}, this predicts natural-rate
    accuracy to within a few percent on \dataseticlrshort{}/\datasetarxivname{}.

    The training prior is a separate choice. Natural-rate training matches the conference base rate,
    but the native decision rule then predicts rejection too often, since rejects are the majority class. \Cref{tab:training_prior_discrepancy}
    summarizes this as the gap between accept and reject recall. Natural-rate \dataseticlrshort{} and \datasetarxivname{} models
    predict rejection far more often than acceptance under their native threshold, and tuning the threshold on validation data can shift it too far the other way.
    Balanced-trained models are closer to zero, so we use balanced training for the headline \systemfamily{} models. This removes the excess rejection predictions without per-dataset threshold tuning.

    \subsection{Methods}

    \textbf{Chosen models.}
    We use Qwen2.5 text and Qwen2.5-VL vision backbones \citep{qwen2.5,bai2025qwen2} because our original deployment target was \dataseticlrshort{}, and Qwen2.5 is the latest Qwen generation that plausibly predates public ICLR '25 decisions (text in December '24, VL on January 26, '25). This makes \dataseticlrshort{} a clean deployment setting by the same cutoff logic, while later Qwen3/3.5 models are more likely to have seen those results; \datasetarxivname{} was added later and did not drive backbone selection. We also retrain a Qwen3 text backbone on \dataseticlrshort{} and find that swapping the base model moves accuracy little (\appref{appendix:backbone_choice}).

    \noindent Our main comparison is against DeepReview-7B/14B \citep{liu2025deepreview} (trained through ICLR '24), Gemini-3.1-Pro (January '25 cutoff), and GPT-5.5 (December '25 cutoff) reasoning baselines. GPT-5.5's cutoff postdates the public release of ICLR '25 decisions, so we probe it for memorization directly (\appref{appendix:memory_probe}). It recognizes many of our test papers, but the decisions it claims to recall are not reliable. Once its accept bias is accounted for, they are at chance. It also cannot regenerate abstracts from titles (ROUGE-L \citep{lin-2004-rouge} at noise level), and its accuracy on the papers it recognizes matches its accuracy on those it does not, including on ICLR '25. These probes provide no clear evidence that recalled acceptance decisions explain its baseline performance.

    \noindent We additionally compare against Jev \citep{typesafe2026jev}, a much cheaper model claimed to be well calibrated. It returns an acceptance probability from text, so we give it the same input as \systemtext{} and compare calibration in \Cref{sec:where_signal_lies}.

    \textbf{Baseline Prompting.}
    For main results, DeepReview runs in its native Standard Mode with four simulated
    reviewers, self-verification, and a meta-review decision. Gemini/GPT baselines use one
    venue-generic AI-Scientist-style prompt \citep{lu2024ai,yamada2025ai} that requests a
    1--10 rating and an Accept/Reject decision. Mini and large frontier variants run
    at high reasoning effort. The generated decision is the default hard label, and the
    rating is a ranking score, not a probability. Qwen rows use the answer-only
    supervised fine-tuning (SFT) prompt or the reasoning reinforcement-learning (RL)
    prompt (\appref{appendix:prompting}).

    We also evaluate frontier coding-agent harnesses with the same rating-plus-decision task. Each Codex or Claude Code run receives an isolated paper workspace with stripped markdown, page screenshots, instructions, and a schema validator that enforces a well-formed \texttt{PREDICTIONS.json}; we log the full trajectory in \texttt{session.jsonl}. Agents run at the same high reasoning effort, inspect text and images selectively, and stay siloed without internet (prompt and layout in \appref{appendix:claude_review_prompts}).

    \textbf{Baseline Correction Measures.}
    Under the base prompt, several models strongly favor one decision class. We also tested
    more critical prompt variants, but bringing predicted acceptance rates closer to 50\%
    did not reliably improve accuracy. We therefore retain the base prompt, the same one used to
    start SFT/RL training, for the main comparisons. These base prompts and their critical variants are defined in
    \appref{appendix:prompt_ablations}.

    For each baseline, we also tune a threshold on its score using a small held-out sample of papers, and
    use the thresholded score as the hard label when it improves accuracy. Out-of-the-box thresholds are skewed: under its default threshold, GPT predicts
    rejection much more often than acceptance, and Gemini leans the same way, though less strongly. Yet both still show relatively high AUC, suggesting the predicted
    rating is well-ordered while the chosen decision threshold is biased. Calibrating the
    threshold generally helps baseline numbers, in agreement with the findings of
    \citet{lu2024ai}.

    \textbf{SFT.}
    Supervised fine-tuning is our primary method for studying data representation, training
    distribution, scaling, and calibration. We fine-tune Qwen2.5 text models at 3B, 7B, and
    14B scale, and Qwen2.5-VL models at 3B and 7B scale, using LLaMA-Factory
    \citep{zheng2024llamafactoryunifiedefficientfinetuning}. Let $P_\theta(\cdot \mid x)$
    denote the model's next-token distribution at the decision token for paper $x$. In the Qwen2.5
    vocabulary, \texttt{Accept} and \texttt{Reject} are individual tokens, so we read off the
    probability of each and normalize over the two:
    \[
        p_{\mathrm{accept}}=\frac{P_\theta(\texttt{Accept}\mid x)}{P_\theta(\texttt{Accept}\mid x)+P_\theta(\texttt{Reject}\mid x)}.
    \]
    After SFT, almost all decision mass lies on these two tokens (validated across all models),
    so $p_{\mathrm{accept}}$ behaves as a usable binary accept probability. Full training
    settings are documented in \appref{appendix:training_settings}.
    The SFT sweeps vary training source and scale, class prior, text versus screenshot inputs,
    mixed text+vision inputs, and suffixes such as handcrafted paper statistics or generated reviews
    (\appref{appendix:data_mixture}).

    \textbf{RL.}
    For reinforcement-learning variants, we fine-tune the policy with Group Relative Policy
    Optimization (GRPO) \citep{Guo_2025} using SkyRL \citep{cao2025skyrl} on the same paper-only accept/reject task. The
    policy uses the reasoning prompt, samples a group of completions for each paper, and
    updates using within-group relative rewards rather than a learned reward model, so the
    model can reason before its final label. We score each completion $y$ for paper $x$ with:
    \[
        r(y,x)=0.8\cdot\mathbf{1}\!\left[d(y)=d^\star(x)\right]
              +0.2\cdot\mathbf{1}\!\left[\textsc{format}(y)\right],
    \]
    where $d(y)$ extracts the boxed Accept/Reject decision with keyword fallback, $d^\star(x)$ is the ground-truth decision, and $\textsc{format}(y)$ checks that the completion has the required \texttt{<think>{\ldots}</think>} block followed by \texttt{<answer>{\textbackslash}boxed\{Accept|Reject\}</answer>}. The format term matters early in training, when many completions skip the boxed answer (details in \appref{appendix:rl}).

    \section{Results}\label{sec:results}

    We first define our metrics (\Cref{sec:eval_metrics}), then set a simple paper features baseline (\Cref{sec:tabular_baseline}), and finally report main results against prompting and coding-agent baselines (\Cref{sec:main_results}).

    \subsection{Evaluation Metrics}\label{sec:eval_metrics}

    Each paper has a true decision $z\in\mathcal{Z}$, and each system produces
    a hard prediction $\hat z\in\mathcal{Z}$ and an accept score $s\in[0,1]$, where higher means
    more likely to be accepted. AUC evaluates each system's ranking without requiring its scores to represent
    calibrated probabilities. For \systemfamily{}, $s$ is the accept probability $p_{\mathrm{accept}}$, while for frontier models $s$ is a normalized 1--10 rating.

    \begin{table}[H]
        \centering
        \vspace{-0.45em}
        \footnotesize
        \setlength{\tabcolsep}{2.2pt}
        \renewcommand{\arraystretch}{1.04}
        \setlength{\aboverulesep}{1.8pt}\setlength{\belowrulesep}{1.8pt}
        \newcommand{\metricssubgrouphead}[1]{%
            \addlinespace[0.20ex]
            \rowcolor{mainresultshighlight}\multicolumn{2}{c}{\textbf{#1}} \\
            \addlinespace[0.2em]
        }
        \begin{tabular}{@{}>{\raggedright\arraybackslash}m{0.29\columnwidth}@{\hspace{0.7em}}>{\raggedright\arraybackslash}m{0.64\columnwidth}@{}}
            \toprule
            \textbf{Metric} & \multicolumn{1}{c}{\textbf{Definition and role in evaluation}} \\
            \midrule
            \metricssubgrouphead{Primary metrics}
            \textsuperscript{$\star$}Accuracy (Acc) &
            Fraction of papers whose predicted decision matches the truth. On our balanced 50/50 test set, accepts and rejects count equally, so always-reject scores only $50\%$. \\
            \cmidrule(lr){1-2}
            \textsuperscript{$\star$}Area under the ROC curve (AUC) &
            The model produces a score $s$, and across thresholds we plot the fraction of accepts called accept ($y$) against the fraction of rejects called accept ($x$); AUC is the area under this curve, the chance a random accept outscores a random reject. \\
            \midrule
            \metricssubgrouphead{Secondary diagnostics}
            \textsuperscript{$\diamond$}Accept / reject recall ($R_A$, $R_R$) &
            Model's ability to recall all instances of each binary label. Shows whether errors lean toward accepts or rejects. \\
            \cmidrule(lr){1-2}
            \textsuperscript{$\diamond$}Predicted acceptance rate ($p_A$) &
            Fraction of all papers the model calls accept. Near $50\%$ indicates that the model does not have an inherent bias toward the accept or reject class. \\
            \cmidrule(lr){1-2}
            \textsuperscript{$\diamond$}Rating correlation ($\rho_R$) &
            Spearman correlation between the accept score $s$ and each paper's average reviewer rating, measuring alignment with human judgments. \\
            \cmidrule(lr){1-2}
            \textsuperscript{$\diamond$}Expected calibration error (ECE) &
            Average gap between confidence and accuracy across confidence bins on the balanced test. Low ECE means $X\%$-confidence predictions are right about $X\%$ of the time. \\
            \bottomrule
        \end{tabular}
        \vspace{0.05em}
        \caption{\textbf{Evaluation diagnostics.} Primary metrics (\textsuperscript{$\star$}) steer training and data decisions. Secondary diagnostics (\textsuperscript{$\diamond$}) drive behavioral analysis. An ideal predictor is accurate (Acc), ranks accepts above rejects (AUC), and is well calibrated.}
        \label{tab:evaluation_metrics}
        \vspace{-0.25em}
    \end{table}

    When studying \systemfamily{} reliability, we ask whether confidence tracks realized
    accuracy after post-hoc calibration. The decision confidence is $c=p_{\mathrm{accept}}$ for an Accept
    prediction and $c=1-p_{\mathrm{accept}}$ for a Reject prediction. We bin papers by $c$, such that in
    each bin $B_m$, $\mathrm{Acc}(B_m)$ is the fraction of correct predictions and
    $\mathrm{Conf}(B_m)$ is the average confidence. ECE is the weighted gap
    $|\mathrm{Acc}(B_m)-\mathrm{Conf}(B_m)|$ on the balanced test; \Cref{sec:where_signal_lies} also checks the
    acceptance probability at real acceptance rates. Frontier ratings are 1--10 review-style scores, not
    accept/reject probabilities, so we report calibration only for \systemfamily{} and Jev.

    Our primary metrics ask whether a system both makes the right binary decision and ranks
    accepted papers above rejected papers. The goal is to design a predictor that makes
    human-like decisions while producing a score that orders papers the way peer review
    does. Accuracy tests whether predictions match final decisions without hiding
    accept/reject collapse, while AUC tests whether accepted papers receive higher scores
    than rejected papers independent of a fixed threshold. This matters because the same
    model may be used on conferences or years outside its training distribution, where
    post-hoc threshold tuning may not be available. Secondary diagnostics separate class bias
    from rating alignment and calibration.

    \subsection{A Simple Paper Features Baseline}\label{sec:tabular_baseline}

\begin{figure*}[h!]
        \centering
        \subcaptionbox{Top feature weights on logistic regression of paper features.\label{fig:feature_probe_top_weights}}[0.54\textwidth]{%
            \parbox[c][2.30in][c]{\linewidth}{%
                \centering
\scriptsize
\setlength{\tabcolsep}{2.2pt}
\renewcommand{\arraystretch}{1.16}
\begin{tabular}{@{}>{\centering\arraybackslash}p{0.224\linewidth}>{\centering\arraybackslash}p{0.224\linewidth}@{\hspace{10pt}}>{\centering\arraybackslash}p{0.224\linewidth}>{\centering\arraybackslash}p{0.224\linewidth}@{}}
\toprule
\multicolumn{2}{c}{\textbf{\textit{ICLR}}} & \multicolumn{2}{c}{\textbf{\textit{arXiv}}} \\
\cmidrule(lr){1-2}\cmidrule(l){3-4}
Accept & Reject & Accept & Reject \\
\midrule
Paper length (+0.47) & Sentence length (-0.18) & Paper length (+0.48) & Sentence length (-0.24) \\
Abstract sentence count (+0.23) & Abstract words (-0.14) & Math notation/1k words (+0.19) & Recent citation share (-0.15) \\
Vocabulary richness (+0.21) & Rare-word ratio (-0.14) & Ablation mentions (+0.14) & Sections per pg. (-0.15) \\
Figure refs. per pg. (+0.18) & Graph-topic terms (-0.08) & Cross-refs. per pg. (+0.11) & Title words (-0.09) \\
Appendix refs. per pg. (+0.13) & Sections per pg. (-0.08) & Appendix refs. per pg. (+0.10) & RL terms/1k words (-0.07) \\
\bottomrule
\end{tabular}

            }%
        }
        \hfill
        \subcaptionbox{Top 3 accept-feature distributions.\label{fig:feature_probe_density_lines}}[0.43\textwidth]{%
            \parbox[c][2.30in][c]{\linewidth}{%
                \centering
                \includegraphics[width=0.94\linewidth]{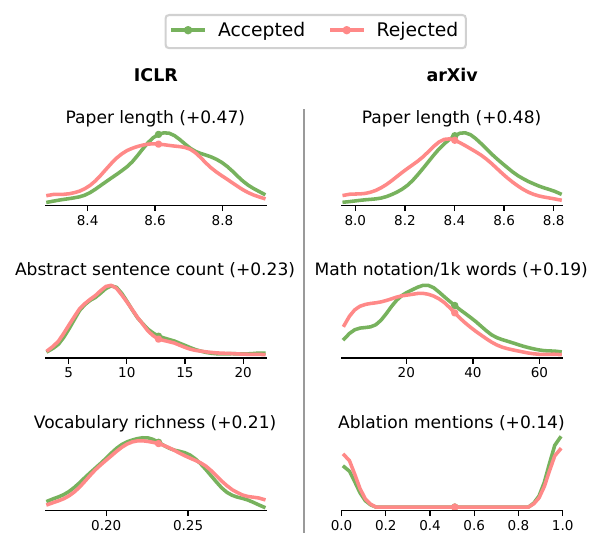}
            }%
        }
        \caption{\textbf{Feature-probe audit.} \textbf{(a)} The table reports top standardized
        logistic-regression weights from separate \dataseticlrshort{}- and \datasetarxivname{}-trained
        models, with weights in parentheses. \textbf{(b)} The plots show accepted and rejected paper
        distributions for the top three accept-associated features within each source on held-out
        \dataseticlrshort{} and \datasetarxivname{} papers. In feature names, \textit{per pg.} means
        per estimated body page. Full feature definitions are in \appref{app:tabular_features}.}
        \label{fig:feature_probe_joint_audit}
        \vspace{-0.55em}
    \end{figure*}

    Before training any large model, we ask how much of the accept/reject decision is recoverable from simple, hand-crafted features. We fit two full-data logistic-regression models, one on \dataseticlrshort{} and one on
    \datasetarxivname{}, using paper features such as length, references, math notation, topic
    terms, and rhetorical counts. The models exclude source identifiers, venue labels, ratings,
    citations, author names, and body-page counts, since page metadata is not comparable between
    the \dataseticlrshort{} and \datasetarxivname{} text exports. These simple models reach 61.1\% accuracy on \dataseticlrshort{} and 62.5\% on \datasetarxivname{}.
    Their coefficients reveal which readily measured paper features are associated with acceptance. \Cref{fig:feature_probe_joint_audit} shows that
    accept features generally have to do with paper length, references, and figures, while reject features are
    more lexical (full inventory in \appref{appendix:feature-probe}).

    \subsection{Main Results}\label{sec:main_results}

    Table~\ref{tab:main_results} summarizes performance on the \systemfamily{} \dataseticlrshort{} and
    \datasetarxivname{} test sets. We start by analyzing baselines, then study the data and
    training effects behind \systemfamily{}.

\begin{table}[H]
\centering
\definecolor{paacceptbias}{HTML}{2F7D45}
\definecolor{parejectbias}{HTML}{B24A3A}
\definecolor{mainresultsneutralband}{HTML}{E6E6E6}
\definecolor{mainresultsbaselineband}{HTML}{E6E6E6}
\definecolor{mainresultssftband}{HTML}{D6DEEC}
\definecolor{mainresultsrlband}{HTML}{F3D2D2}
\definecolor{mainresultssftrow}{HTML}{ECF0F7}
\definecolor{mainresultsrlrow}{HTML}{FBE9E9}
\definecolor{mainresultssftname}{HTML}{2F5496}
\definecolor{mainresultsrlname}{HTML}{B23A3A}
\definecolor{mainresultshighlight}{HTML}{CDDFF5}
\newcommand{\mainresultsgrouphead}[2]{%
  \addlinespace[0.28ex]
  \rowcolor{#1}\multicolumn{12}{c}{\textbf{#2}} \\
  \addlinespace[-0.35ex]
}
\newcommand{\mainresultssubgrouphead}[1]{%
  \addlinespace[0.20ex]
  \multicolumn{12}{l}{\scriptsize\itshape #1} \\
  \addlinespace[-0.45ex]
}
\newcommand{\mainresultsmodelindent}{\hspace{0.65em}}
\newcommand{\mainresultspadeltalight}[1]{\fontsize{4.1}{4.2}\selectfont #1}
\newcommand{\mainresultspadeltamedium}[1]{\fontseries{b}\fontsize{4.1}{4.2}\selectfont #1}
\newcommand{\mainresultspadeltastrong}[1]{\bfseries\boldmath\fontsize{4.1}{4.2}\selectfont #1}
\newcommand{\mainresultspabiaslight}[4]{\textcolor{#1}{\makebox[1.46em][l]{\mainresultspadeltalight{#2#3.#4}}}}
\newcommand{\mainresultspabiasmedium}[4]{\textcolor{#1}{\makebox[1.46em][l]{\mainresultspadeltamedium{#2#3.#4}}}}
\newcommand{\mainresultspabiasstrong}[4]{\textcolor{#1}{\makebox[1.46em][l]{\mainresultspadeltastrong{#2#3.#4}}}}
\newcommand{\mainresultsnum}[2]{\makebox[1.45em][r]{#1}\makebox[0.78em][l]{.#2}}
\newcommand{\mainresultsnumbf}[2]{\makebox[1.45em][r]{\textbf{#1}}\makebox[0.78em][l]{\textbf{.#2}}}
\newcommand{\mainresultsnumul}[2]{\makebox[1.45em][r]{\underline{#1}}\makebox[0.78em][l]{\underline{.#2}}}
\newcommand{\mainresultspanum}[2]{\makebox[1.42em][r]{#1}\makebox[0.48em][l]{.#2}}
\newcommand{\mainresultspanumbf}[2]{\makebox[1.42em][r]{\textbf{#1}}\makebox[0.48em][l]{\textbf{.#2}}}
\newcommand{\mainresultspanumul}[2]{\makebox[1.42em][r]{\underline{#1}}\makebox[0.48em][l]{\underline{.#2}}}
\newcommand{\mainresultszerodec}[1]{.#1}
\newcommand{\mainresultszerodecbf}[1]{\textbf{.#1}}
\newcommand{\mainresultszerodecul}[1]{\underline{.#1}}
\newcommand{\mainresultssigndec}[2]{\makebox[0.42em][r]{#1}\makebox[0.86em][l]{.#2}}
\newcommand{\mainresultssigndecbf}[2]{\makebox[0.42em][r]{\textbf{#1}}\makebox[0.86em][l]{\textbf{.#2}}}
\newcommand{\mainresultssigndecul}[2]{\makebox[0.42em][r]{\underline{#1}}\makebox[0.86em][l]{\underline{.#2}}}
\newcommand{\mainresultsar}[2]{\makebox[3.35em][c]{#1/#2}}
\newcommand{\mainresultspaheader}{\makebox[3.90em][c]{\textsuperscript{$\diamond$}$p_A$}}
\newcommand{\mainresultspa}[3]{\makebox[3.90em][l]{\csname #1\endcsname{#2}{#3}}}
\newcommand{\mainresultspawithbias}[8]{\makebox[3.90em][l]{\csname #1\endcsname{#2}{#3}\hspace{0.42em}\csname #5\endcsname{#4}{#6}{#7}{#8}}}
\newcommand{\modtext}{\textcolor{black!50}{\scriptsize\textsc{Text}}}
\newcommand{\modvision}{\textbf{\scriptsize\textsc{Vision}}}
\newcommand{\modboth}{\scriptsize\textsc{Both}}
\scriptsize
\setlength{\tabcolsep}{2.15pt}
\renewcommand{\arraystretch}{1.075}
\resizebox{\textwidth}{!}{%
\begin{tabular}{lccccccc!{\hspace{6pt}}cccc}
\toprule
& & & \multicolumn{5}{c}{\textbf{\dataseticlrshort{}}} & \multicolumn{4}{c}{\textbf{\datasetarxivname{}}} \\
\cmidrule(lr){4-8} \cmidrule(lr){9-12}
\textbf{Method} & \textbf{Train} & \textbf{Input} & \textsuperscript{$\star$}Acc & \textsuperscript{$\star$}AUC & \textsuperscript{$\diamond$}$R_A/R_R$ & \mainresultspaheader & \textsuperscript{$\diamond$}$\rho_R$ & \textsuperscript{$\star$}Acc & \textsuperscript{$\star$}AUC & \textsuperscript{$\diamond$}$R_A/R_R$ & \mainresultspaheader \\
\midrule
\mainresultsgrouphead{mainresultsbaselineband}{Baselines}
\mainresultssubgrouphead{Mini Frontier Reasoning}
\mainresultsmodelindent Gemini-3-Flash & - & \modvision & \mainresultsnum{65}{6} & \mainresultszerodec{672} & \mainresultsar{67}{65} & \mainresultspawithbias{mainresultspanumbf}{50}{1}{paacceptbias}{mainresultspabiasstrong}{+}{0}{1} & \mainresultssigndec{+}{39} & \mainresultsnum{62}{4} & \mainresultszerodec{653} & \mainresultsar{50}{75} & \mainresultspawithbias{mainresultspanum}{37}{3}{parejectbias}{mainresultspabiasmedium}{-}{12}{7} \\
\mainresultsmodelindent Gemini-3-Flash & - & \modtext & \mainresultsnum{63}{0} & \mainresultszerodec{643} & \mainresultsar{65}{61} & \mainresultspawithbias{mainresultspanum}{52}{3}{paacceptbias}{mainresultspabiasstrong}{+}{2}{3} & \mainresultssigndec{+}{27} & \mainresultsnum{61}{2} & \mainresultszerodec{655} & \mainresultsar{51}{71} & \mainresultspawithbias{mainresultspanum}{39}{6}{parejectbias}{mainresultspabiasmedium}{-}{10}{4} \\
\mainresultsmodelindent GPT-5.4-Mini & - & \modvision & \mainresultsnum{57}{0} & \mainresultszerodec{650} & \mainresultsar{30}{84} & \mainresultspawithbias{mainresultspanum}{22}{6}{parejectbias}{mainresultspabiaslight}{-}{27}{4} & \mainresultssigndec{+}{30} & \mainresultsnum{62}{6} & \mainresultszerodec{677} & \mainresultsar{79}{46} & \mainresultspawithbias{mainresultspanum}{65}{7}{paacceptbias}{mainresultspabiaslight}{+}{15}{7} \\
\mainresultsmodelindent GPT-5.4-Mini & - & \modtext & \mainresultsnum{60}{0} & \mainresultszerodec{632} & \mainresultsar{50}{70} & \mainresultspawithbias{mainresultspanum}{39}{4}{parejectbias}{mainresultspabiasmedium}{-}{10}{6} & \mainresultssigndec{+}{30} & \mainresultsnum{58}{6} & \mainresultszerodec{635} & \mainresultsar{43}{74} & \mainresultspawithbias{mainresultspanum}{34}{0}{parejectbias}{mainresultspabiaslight}{-}{16}{0} \\
\mainresultssubgrouphead{Large Frontier Reasoning}
\mainresultsmodelindent Gemini-3.1-Pro & - & \modvision & \mainresultsnum{65}{1} & \mainresultszerodec{690} & \mainresultsar{49}{81} & \mainresultspawithbias{mainresultspanum}{34}{4}{parejectbias}{mainresultspabiaslight}{-}{15}{6} & \mainresultssigndec{+}{38} & \mainresultsnum{68}{5} & \mainresultszerodec{736} & \mainresultsar{52}{85} & \mainresultspawithbias{mainresultspanum}{33}{2}{parejectbias}{mainresultspabiaslight}{-}{16}{8} \\
\mainresultsmodelindent Gemini-3.1-Pro & - & \modtext & \mainresultsnum{62}{8} & \mainresultszerodec{673} & \mainresultsar{42}{83} & \mainresultspawithbias{mainresultspanum}{29}{6}{parejectbias}{mainresultspabiaslight}{-}{20}{4} & \mainresultssigndec{+}{36} & \mainresultsnum{67}{7} & \mainresultszerodec{710} & \mainresultsar{69}{66} & \mainresultspawithbias{mainresultspanumbf}{50}{6}{paacceptbias}{mainresultspabiasstrong}{+}{0}{6} \\
\mainresultsmodelindent GPT-5.5 & - & \modvision & \mainresultsnum{63}{4} & \mainresultszerodec{692} & \mainresultsar{87}{40} & \mainresultspawithbias{mainresultspanum}{73}{6}{paacceptbias}{mainresultspabiaslight}{+}{23}{6} & \mainresultssigndec{+}{40} & \mainresultsnum{68}{3} & \mainresultszerodec{743} & \mainresultsar{58}{78} & \mainresultspawithbias{mainresultspanum}{39}{3}{parejectbias}{mainresultspabiasmedium}{-}{10}{7} \\
\mainresultsmodelindent GPT-5.5 & - & \modtext & \mainresultsnum{64}{4} & \mainresultszerodec{681} & \mainresultsar{65}{64} & \mainresultspawithbias{mainresultspanumul}{50}{2}{paacceptbias}{mainresultspabiasstrong}{+}{0}{2} & \mainresultssigndec{+}{37} & \mainresultsnum{66}{0} & \mainresultszerodec{695} & \mainresultsar{65}{67} & \mainresultspawithbias{mainresultspanumul}{48}{6}{parejectbias}{mainresultspabiasstrong}{-}{1}{4} \\
\mainresultssubgrouphead{Frontier Reasoning Agents}
\mainresultsmodelindent GPT-5.5 (Codex Agent) & - & \modboth & \mainresultsnum{55}{3} & \mainresultszerodec{667} & \mainresultsar{15}{95} & \mainresultspawithbias{mainresultspanum}{10}{0}{parejectbias}{mainresultspabiaslight}{-}{40}{0} & \mainresultssigndec{+}{39} & \mainresultsnum{56}{2} & \mainresultszerodec{683} & \mainresultsar{19}{93} & \mainresultspawithbias{mainresultspanum}{13}{0}{parejectbias}{mainresultspabiaslight}{-}{37}{0} \\
\mainresultsmodelindent Sonnet-4.6 (Claude Agent) & - & \modboth & \mainresultsnum{61}{2} & \mainresultszerodec{711} & \mainresultsar{38}{85} & \mainresultspawithbias{mainresultspanum}{26}{6}{parejectbias}{mainresultspabiaslight}{-}{23}{4} & \mainresultssigndec{+}{44} & \mainresultsnum{60}{7} & \mainresultszerodec{739} & \mainresultsar{35}{87} & \mainresultspawithbias{mainresultspanum}{23}{8}{parejectbias}{mainresultspabiaslight}{-}{26}{2} \\
\mainresultssubgrouphead{Frontier Decision Models}
\mainresultsmodelindent Jev & - & \modtext & \mainresultsnum{63}{4} & \mainresultszerodec{682} & \mainresultsar{81}{46} & \mainresultspawithbias{mainresultspanum}{67}{9}{paacceptbias}{mainresultspabiaslight}{+}{17}{9} & \mainresultssigndec{+}{37} & \mainresultsnum{60}{6} & \mainresultszerodec{659} & \mainresultsar{78}{44} & \mainresultspawithbias{mainresultspanum}{66}{8}{paacceptbias}{mainresultspabiaslight}{+}{16}{8} \\
\mainresultssubgrouphead{Review-Trained}
\mainresultsmodelindent DeepReview-7B & - & \modtext & \mainresultsnum{59}{7} & \mainresultszerodec{618} & \mainresultsar{56}{64} & \mainresultspawithbias{mainresultspanum}{46}{3}{parejectbias}{mainresultspabiasstrong}{-}{3}{7} & \mainresultssigndec{+}{22} & \mainresultsnum{58}{7} & \mainresultszerodec{620} & \mainresultsar{36}{82} & \mainresultspawithbias{mainresultspanum}{27}{1}{parejectbias}{mainresultspabiaslight}{-}{22}{9} \\
\mainresultsmodelindent DeepReview-14B & - & \modtext & \mainresultsnum{64}{9} & \mainresultszerodec{667} & \mainresultsar{70}{60} & \mainresultspawithbias{mainresultspanum}{55}{3}{paacceptbias}{mainresultspabiasmedium}{+}{5}{3} & \mainresultssigndec{+}{38} & \mainresultsnum{61}{6} & \mainresultszerodec{699} & \mainresultsar{50}{73} & \mainresultspawithbias{mainresultspanum}{38}{0}{parejectbias}{mainresultspabiasmedium}{-}{12}{0} \\
\midrule
\mainresultsgrouphead{mainresultssftband}{PaperScorer SFT}
\mainresultssubgrouphead{Base Instruct Models}
\mainresultsmodelindent Qwen2.5-VL-7B & - & \modvision & \mainresultsnum{54}{1} & \mainresultszerodec{551} & \mainresultsar{76}{32} & \mainresultspawithbias{mainresultspanum}{71}{7}{paacceptbias}{mainresultspabiaslight}{+}{21}{7} & \mainresultssigndec{+}{10} & \mainresultsnum{55}{9} & \mainresultszerodec{571} & \mainresultsar{89}{23} & \mainresultspawithbias{mainresultspanum}{82}{8}{paacceptbias}{mainresultspabiaslight}{+}{32}{8} \\
\mainresultsmodelindent Qwen2.5-7B & - & \modtext & \mainresultsnum{49}{9} & \mainresultszerodec{496} & \mainresultsar{100}{0} & \mainresultspawithbias{mainresultspanum}{99}{9}{paacceptbias}{mainresultspabiaslight}{+}{49}{9} & \mainresultssigndec{-}{00} & \mainresultsnum{53}{0} & \mainresultszerodec{519} & \mainresultsar{74}{32} & \mainresultspawithbias{mainresultspanum}{70}{9}{paacceptbias}{mainresultspabiaslight}{+}{20}{9} \\
\mainresultssubgrouphead{\dataseticlrshort{}-trained}
\mainresultsmodelindent \textbf{\textcolor{mainresultssftname}{PaperScorer}}-Vision & \dataseticlrshort{} & \modvision & \cellcolor{mainresultshighlight}\mainresultsnumbf{68}{3} & \cellcolor{mainresultshighlight}\mainresultszerodecbf{736} & \cellcolor{mainresultshighlight}\mainresultsar{69}{67} & \cellcolor{mainresultshighlight}\mainresultspawithbias{mainresultspanum}{50}{9}{paacceptbias}{mainresultspabiasstrong}{+}{0}{9} & \mainresultssigndecbf{+}{48} & \mainresultsnum{64}{2} & \mainresultszerodec{719} & \mainresultsar{50}{78} & \mainresultspawithbias{mainresultspanum}{35}{4}{parejectbias}{mainresultspabiasmedium}{-}{14}{6} \\
\mainresultsmodelindent \textbf{\textcolor{mainresultssftname}{PaperScorer}}-Text & \dataseticlrshort{} & \modtext & \mainresultsnumul{66}{0} & \mainresultszerodecul{721} & \mainresultsar{63}{69} & \mainresultspawithbias{mainresultspanum}{46}{7}{parejectbias}{mainresultspabiasstrong}{-}{3}{3} & \mainresultssigndecul{+}{47} & \mainresultsnum{59}{5} & \mainresultszerodec{723} & \mainresultsar{28}{91} & \mainresultspawithbias{mainresultspanum}{18}{5}{parejectbias}{mainresultspabiaslight}{-}{31}{5} \\
\mainresultssubgrouphead{arXiv-trained}
\mainresultsmodelindent \textbf{\textcolor{mainresultssftname}{PaperScorer}}-Vision & \datasetarxivmediumname{} & \modvision & \mainresultsnum{59}{8} & \mainresultszerodec{676} & \mainresultsar{41}{79} & \mainresultspawithbias{mainresultspanum}{31}{1}{parejectbias}{mainresultspabiaslight}{-}{18}{9} & \mainresultssigndec{+}{35} & \mainresultsnum{71}{5} & \mainresultszerodecul{825} & \mainresultsar{59}{84} & \mainresultspawithbias{mainresultspanum}{36}{6}{parejectbias}{mainresultspabiasmedium}{-}{13}{4} \\
\mainresultsmodelindent \textbf{\textcolor{mainresultssftname}{PaperScorer}}-Text & \datasetarxivmediumname{} & \modtext & \mainresultsnum{62}{6} & \mainresultszerodec{656} & \mainresultsar{65}{60} & \mainresultspawithbias{mainresultspanum}{52}{3}{paacceptbias}{mainresultspabiasstrong}{+}{2}{3} & \mainresultssigndec{+}{33} & \mainresultsnum{71}{7} & \mainresultszerodec{787} & \mainresultsar{63}{80} & \mainresultspawithbias{mainresultspanum}{40}{8}{parejectbias}{mainresultspabiasmedium}{-}{9}{2} \\
\mainresultsmodelindent \textbf{\textcolor{mainresultssftname}{PaperScorer}}-Vision & \datasetarxivname{} & \modvision & \mainresultsnum{60}{7} & \mainresultszerodec{653} & \mainresultsar{52}{69} & \mainresultspawithbias{mainresultspanum}{41}{8}{parejectbias}{mainresultspabiasmedium}{-}{8}{2} & \mainresultssigndec{+}{31} & \cellcolor{mainresultshighlight}\mainresultsnumbf{77}{5} & \cellcolor{mainresultshighlight}\mainresultszerodecbf{843} & \cellcolor{mainresultshighlight}\mainresultsar{75}{80} & \cellcolor{mainresultshighlight}\mainresultspawithbias{mainresultspanum}{46}{4}{parejectbias}{mainresultspabiasstrong}{-}{3}{6} \\
\mainresultsmodelindent \textbf{\textcolor{mainresultssftname}{PaperScorer}}-Text & \datasetarxivname{} & \modtext & \mainresultsnum{62}{0} & \mainresultszerodec{656} & \mainresultsar{56}{68} & \mainresultspawithbias{mainresultspanum}{43}{9}{parejectbias}{mainresultspabiasmedium}{-}{6}{1} & \mainresultssigndec{+}{33} & \mainresultsnumul{74}{2} & \mainresultszerodec{815} & \mainresultsar{72}{76} & \mainresultspawithbias{mainresultspanum}{47}{3}{parejectbias}{mainresultspabiasstrong}{-}{2}{7} \\
\midrule
\mainresultsgrouphead{mainresultsrlband}{PaperScorer RL}
\mainresultssubgrouphead{Base Instruct Models}
\mainresultsmodelindent Qwen2.5-VL-7B & - & \modvision & \mainresultsnum{50}{1} & \mainresultszerodec{488} & \mainresultsar{100}{1} & \mainresultspawithbias{mainresultspanum}{99}{5}{paacceptbias}{mainresultspabiaslight}{+}{49}{5} & \mainresultssigndec{+}{01} & \mainresultsnum{50}{2} & \mainresultszerodec{529} & \mainresultsar{100}{0} & \mainresultspawithbias{mainresultspanum}{99}{8}{paacceptbias}{mainresultspabiaslight}{+}{49}{8} \\
\mainresultsmodelindent Qwen2.5-7B & - & \modtext & \mainresultsnum{50}{1} & \mainresultszerodec{509} & \mainresultsar{99}{1} & \mainresultspawithbias{mainresultspanum}{99}{3}{paacceptbias}{mainresultspabiaslight}{+}{49}{3} & \mainresultssigndec{+}{04} & \mainresultsnum{51}{1} & \mainresultszerodec{534} & \mainresultsar{100}{2} & \mainresultspawithbias{mainresultspanum}{98}{9}{paacceptbias}{mainresultspabiaslight}{+}{48}{9} \\
\mainresultssubgrouphead{RL-trained}
\mainresultsmodelindent \textbf{\textcolor{mainresultsrlname}{PaperScorer}}-Vision (RL) & \dataseticlrshort{} & \modvision & \mainresultsnum{55}{0} & \mainresultszerodec{517} & \mainresultsar{78}{32} & \mainresultspawithbias{mainresultspanum}{73}{1}{paacceptbias}{mainresultspabiaslight}{+}{23}{1} & \mainresultssigndec{+}{03} & \mainresultsnum{60}{3} & \mainresultszerodec{557} & \mainresultsar{84}{37} & \mainresultspawithbias{mainresultspanum}{73}{6}{paacceptbias}{mainresultspabiaslight}{+}{23}{6} \\
\mainresultsmodelindent \textbf{\textcolor{mainresultsrlname}{PaperScorer}}-Text (RL) & \dataseticlrshort{} & \modtext & \mainresultsnum{56}{1} & \mainresultszerodec{517} & \mainresultsar{99}{13} & \mainresultspawithbias{mainresultspanum}{93}{2}{paacceptbias}{mainresultspabiaslight}{+}{43}{2} & \mainresultssigndec{+}{03} & \mainresultsnum{58}{5} & \mainresultszerodec{596} & \mainresultsar{95}{22} & \mainresultspawithbias{mainresultspanum}{86}{3}{paacceptbias}{mainresultspabiaslight}{+}{36}{3} \\
\bottomrule
\end{tabular}%
}
\vspace{0.8em}
\caption{\textbf{Main results.} We compare prompting baselines, trained reviewers, and \systemfamily{} variants on \dataseticlrshort{} and \datasetarxivname{}, using Acc/AUC as primary correctness and ranking metrics. \systemfamily{} rows use 7B checkpoints. Bold/underline mark first/second per column, where first/second for $p_A$ is closest to the balanced 50\% accept prior (colored deltas show the signed deviation). We report $\rho_R$ only for \dataseticlrshort{}, where every paper carries a reviewer rating. \systemfamily{} improves Acc/AUC over frontier systems, with vision strongest overall and models doing best on the source they were trained on.}
\label{tab:main_results}
\end{table}

    \vspace{-0.6em}

    \textbf{Baseline performance.}\label{inf_results}
    We compare base Qwen instruct models, mini and large frontier Gemini/GPT reasoning models,
    decision models, coding agents, and DeepReview. Untuned Qwen2.5 predicts Accept for 97--99\% of papers
    (\appref{appendix:prompt_ablations}). Smaller high-reasoning models are
    also skewed natively, with Gemini-3-Flash accepting almost every paper and GPT-5.4-Mini rejecting almost every paper. Large frontier
    systems are more accurate and run at high reasoning (median $\sim$2{,}000 reasoning tokens/request;
    Table~\ref{tab:frontier_reasoning_tokens}), but their accept/reject recalls are skewed. Gemini-3.1-Pro accepts only 30--34\% of papers in three of four settings,
    while GPT-5.5 swings between accepting 39\% and 74\%.

    The coding-agent rows test whether interactive access to both modalities changes this pattern.
    While Claude Code has the strongest rating correlation and the agents have among the strongest AUCs in the baseline block, their Acc is lower than
    the best non-agent frontier baselines. Like Jev, coding agents can rank papers in a way
    that agrees with reviewer ratings, but still turn that ranking into skewed accept/reject decisions. DeepReview
    generates review-style judgments and a meta-review decision; its 14B model improves over 7B but still trails
    the best frontier model and \systemfamily{}.

    \textbf{Binary SFT recovers a stronger decision signal.}\label{sec:binary_prediction_powerful}
    The central result in Table~\ref{tab:main_results} is that direct accept/reject supervision gives
    the strongest decision predictor and ranking signal. \looseness=-1 On \dataseticlrshort{},
    \systemvision{} reaches $68.3$ Acc, $2.7\%$ ahead of the best frontier vision model, Gemini-3-Flash. On \datasetarxivname{}, the matched
    \datasetarxivname{}-trained \systemvision{} reaches $77.5$, a larger $9.0\%$ ahead of the best frontier baseline.
    On both datasets, \systemvision{} significantly beats both large frontier models in Acc and AUC
    (one-sided paired bootstrap, $p<.05$; \appref{appendix:statistical_significance}). These gains
    are not a threshold artifact, since \systemvision{} keeps accept and reject recall
    balanced, while the frontier systems rank papers well but natively skew their decisions.

    \begin{figure}[H]
        \centering
        \includegraphics[width=\textwidth]{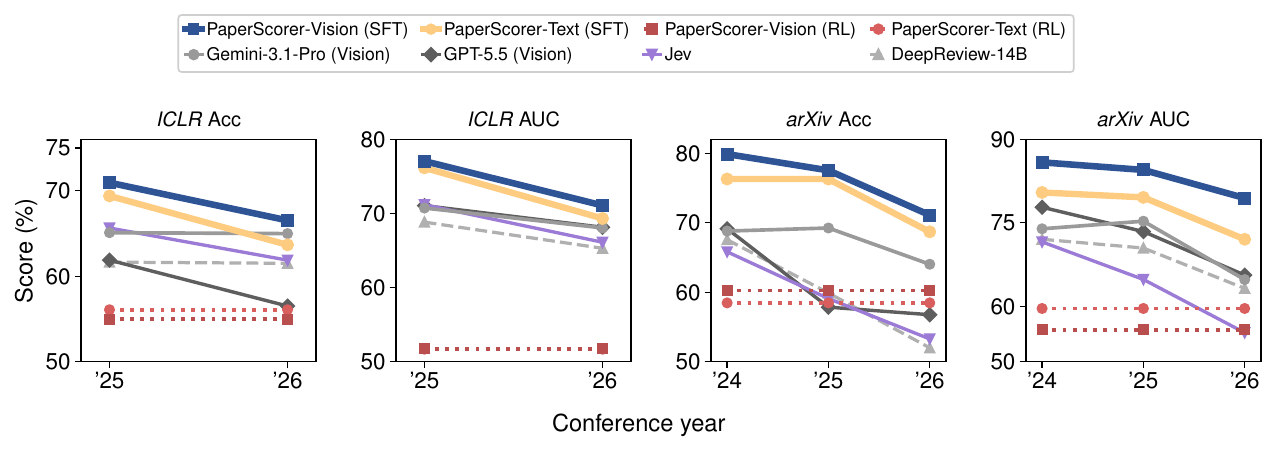}
        \caption{\textbf{Binary SFT is strongest over time.} \systemtext{} and \systemvision{} use matched-source training. Across years, supervised \systemfamily{} leads Acc/AUC, vision is strongest, and RL trails prompting frontier models.}
        \label{fig:sft_prompting_rl_yearwise}
    \end{figure}

    Figure~\ref{fig:sft_prompting_rl_yearwise} shows that the modality and method ordering is stable
    across years, with \systemvision{} best in every matched-source year, prompting
    intermediate, and RL weaker. These results indicate that papers carry recoverable decision signal, although learning that signal through long reasoning traces is difficult with only a single, noisy
    accept/reject bit as reward. Supervising directly on the final decision works better, and vision keeps
    the layout and presentation cues that a markdown extract would strip.

    \looseness=-1 Accuracy is not constant over time, and Acc and AUC drop sharply from '25 to '26 across \systemfamily{}, frontier models, and DeepReview, suggesting a shared change in the evaluation distribution. The early-cycle \datasetarxivname{} '26 slice is also small, with 198 test papers from AAAI, CVPR, and ICLR '26. Each cycle also introduces new methods, benchmarks, and reviewer norms that reset what counts as a strong paper, so some recalibration is unavoidable for any predictor whose training pool is more than a year stale. Training only through '25 transfers poorly to '26 (\appref{appendix:data_mixture}), so we fold '26 into training and expect any deployed predictor to be periodically retrained as new conferences land.

    \looseness=-1 Beyond hard decisions, both \systemtext{} and \systemvision{} reach top AUC on both datasets, ranking accepts above rejects regardless of threshold. The score also tracks human ratings; on \dataseticlrshort{}, \systemvision{}'s $p_{\mathrm{accept}}$ correlates with reviewer ratings more closely than DeepReview and frontier baselines, despite training only on final decisions. \datasetarxivname{} ratings are sparser ($292$/$\sim$$1{,}415$ test papers, skewed toward NeurIPS and ICLR accepts); on that biased subset coding agents correlate with ratings better than \systemfamily{} but still make worse decisions, so correlation with reviewer ratings does not convert into correct accept/reject calls.

\begin{figure}[t!]
        \centering
        \captionsetup[subfigure]{skip=14pt}
        \subcaptionbox{\textbf{Model scale.} 3B/14B differences to 7B, matched on training dataset. \colorbox{mainresultshighlight}{Shaded} $\Delta$Acc cells are not significant at a 95\% paired bootstrap CI.\label{fig:model_scale_deltas}}[0.55\textwidth]{%
            \parbox[c][1.85in][c]{\linewidth}{%
\centering
\footnotesize
\newcommand{\modelscalemodtext}{\textcolor{black!50}{\footnotesize\textsc{Text}}}
\newcommand{\modelscalemodvision}{\textbf{\footnotesize\textsc{Vision}}}
\setlength{\tabcolsep}{2pt}
\renewcommand{\arraystretch}{1.18}
\begin{tabular}{@{}l@{\hspace{14pt}}l@{\hspace{14pt}}l@{\hspace{14pt}}>{\centering\arraybackslash}m{3.4em}@{\hspace{10pt}}>{\centering\arraybackslash}m{3.4em}@{}}
\toprule
\textbf{Size} & \textbf{Train} & \textbf{Input} & \makecell{\textbf{\dataseticlrshort{}}\\$\Delta$\textbf{Acc}} & \makecell{\textbf{\datasetarxivname{}}\\$\Delta$\textbf{Acc}} \\
\midrule
3B & \dataseticlrshort{} & \modelscalemodvision & \cellcolor{mainresultshighlight}-1.2 & \cellcolor{mainresultshighlight}+1.0 \\
3B & \dataseticlrshort{} & \modelscalemodtext & \cellcolor{mainresultshighlight}+1.1 & \cellcolor{mainresultshighlight}+0.9 \\
3B & \datasetarxivmediumname{} & \modelscalemodvision & \cellcolor{mainresultshighlight}-1.0 & +3.1 \\
3B & \datasetarxivmediumname{} & \modelscalemodtext & \cellcolor{mainresultshighlight}-1.7 & \cellcolor{mainresultshighlight}-0.7 \\
14B & \dataseticlrshort{} & \modelscalemodtext & \cellcolor{mainresultshighlight}+1.0 & \cellcolor{mainresultshighlight}0.0 \\
\bottomrule
\end{tabular}

            }%
        }
        \hfill
        \subcaptionbox{Acc vs train data size on \dataseticlrshort{} and \datasetarxivname{}.\label{fig:in_domain_data_scaling}}[0.42\textwidth]{%
            \parbox[c][1.85in][c]{\linewidth}{%
                \centering
                \includegraphics[width=\linewidth]{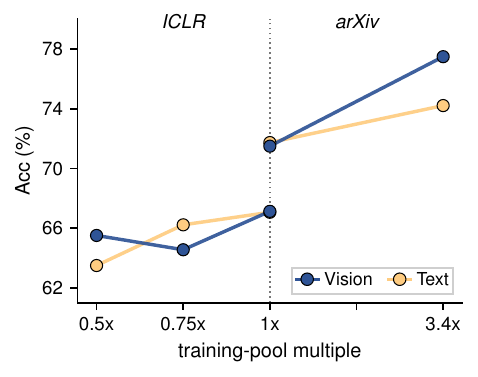}
            }%
        }
        \caption{\textbf{Scaling effects.} Decision accuracy is largely invariant to model size, with 3B models matching 7B Acc, while data scale drives in-domain accuracy, most visibly on \datasetarxivname{} where the 3.4$\times$ pool boost dominates.}
        \label{fig:model_data_scaling}
    \end{figure}

    \noindent\textbf{Model and data scale.} Model scale changes decision accuracy only modestly once the SFT objective is fixed (Figure~\ref{fig:model_data_scaling}a). Relative to matched 7B checkpoints, 3B and 14B variants barely move Acc, with no consistent direction across modality or source. A paired 95\% bootstrap CI reveals that nearly every change in Acc is statistically insignificant (shaded in Figure~\ref{fig:model_scale_deltas}). Vision is similarly competitive at smaller scale, so in Figure~\ref{fig:model_data_scaling}b the largest \datasetarxivname{} Acc gain comes from data scale rather than parameter scale.

    For \dataseticlrshort{}, we vary dataset size from 50\% to 75\% of the full pool. For \datasetarxivname{}, the \datasetarxivmediumname{} and \datasetarxivname{} sets give us a 3.4$\times$ multiple. More data helps for both datasets, and Figure~\ref{fig:model_data_scaling}b shows that the dominant \datasetarxivname{} gain is data scale, not parameter scale, as the curve climbs steeply from \datasetarxivmediumname{} to \datasetarxivname{}, well past the spread of the 3B/7B/14B parameter sweep. \dataseticlrshort{} is much smaller, so taking 50\% and 75\% of this population compresses the visible slope, and the curves should not be read as comparable in absolute terms.

    \vspace{1em}

    \begin{wrapfigure}[10]{r}{0.47\textwidth}
        \vspace{-1.12em}
        \centering
        \includegraphics[width=\linewidth]{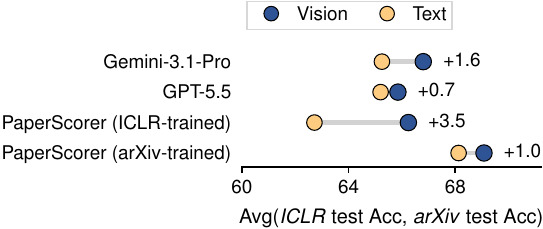}
        \vspace{-1.35em}
        \caption{\textbf{Vision is strongest on average.} Points show the benefit of using vision over text on both datasets.}
        \label{fig:vision_text_dumbbell}
        \vspace{-0.35em}
    \end{wrapfigure}
    \noindent\textbf{Vision is strongest.} It improves over text in the average \dataseticlrshort{}/\datasetarxivname{} view
    (Figure~\ref{fig:vision_text_dumbbell}) and in almost every dataset-specific comparison (\Cref{tab:vision_text_significance}).
    For \systemfamily{}, the Vision row beats its Text row on both datasets, and a paired
    bootstrap confirms the gain ($p=.012$ and $.005$ after Holm--Bonferroni correction
    \citep{abdi2010holm}). For the frontier families, the vision--text gaps are small and none is significant once their
    decisions are thresholded (\appref{appendix:statistical_significance}).
    Screenshots preserve complete figures and tables, equation layout, and visual polish, all of which
    markdown extraction discards.

    Second, the signal largely survives heavy compression, since tiling all ten \systemvision{} pages into a
    single 2380$\times$1512 image cuts vision tokens by about $63\%$ (roughly $12{,}420$ to $4{,}590$),
    and only drops performance on \dataseticlrshort{} by 2.2\% (Table~\ref{tab:panel_ablation}),
    a small change relative to the $>60\%$ token cut.

    To investigate vision's consistent advantage and the small cost of compression, we asked the
    base VL backbone to perform OCR on each input. On per-page screenshots it recovers ${\sim}80\%$
    of the content verbatim against the markdown extract. On the compressed panel, OCR retention drops
    to ${\sim}30\%$. Compression sharply reduces recoverable text while preserving layout, figure placement, and overall
    visual presentation. Prior work supports this direction, as diagram count and quality correlate
    with citations \citep{marshall2021numberqualitydiagramsscholarly}, and paper organization shapes
    peer-review outcomes \citep{jung2026drivespaperacceptanceprocesscentric}.

    \begin{resulttakeaways}[Conjecture: vision works through paper gestalt]
        The paper-gestalt hypothesis \citep{huang2018deeppapergestalt} proposed that a paper's visual
        appearance can predict acceptance, but the original evidence mixed visual presentation with
        author and venue cues. Our anonymized, matched text/vision comparisons support it, and much of the
        performance survives compression that sharply reduces recoverable text. This suggests that visual
        presentation contributes to prediction, though we cannot isolate it from text that remains readable.
    \end{resulttakeaways}

    \begin{table}[t!]
        \centering
        \small
        \setlength{\tabcolsep}{4.5pt}
        \renewcommand{\arraystretch}{1.08}
        \captionsetup{skip=0.55em}
        \begin{tabular}{@{}lcccccc@{}}
\toprule
\textbf{Input} & \textbf{Vision toks.} & \textbf{$\Delta$ toks.} & \textbf{Acc} & \textbf{$\Delta$ Acc} & \textbf{AUC} & \textbf{$\Delta$ AUC} \\
\midrule
Per-page \systemvision{} & $\sim$12,420 & -- & 68.3 & -- & .736 & -- \\
Panel \systemvision{} & 4,590 & -63\% & 66.1 & -2.2 & .704 & -.032 \\
\bottomrule
\end{tabular}

        \caption{\textbf{Panel-token tradeoff (\dataseticlrshort{}).} On \dataseticlrshort{}, the \dataseticlrshort{}-trained panel model tiles ten pages into one 2380$\times$1512 image, preserving most of the per-page vision result at about a third of the vision-token cost. We did not train an \datasetarxivname{} panel model, so this ablation is \dataseticlrshort{} only.}
        \label{tab:panel_ablation}
        \vspace{-1.25em}
    \end{table}


    \section{Where does this signal lie?}
    \label{sec:where_signal_lies}
    Having shown that paper content carries a recoverable, well-ranked decision signal, we now ask
    \emph{where} that signal lives, across confidence ranges, papers, and
    conferences.

    \begin{figure}[H]
        \centering
        \includegraphics[width=\textwidth]{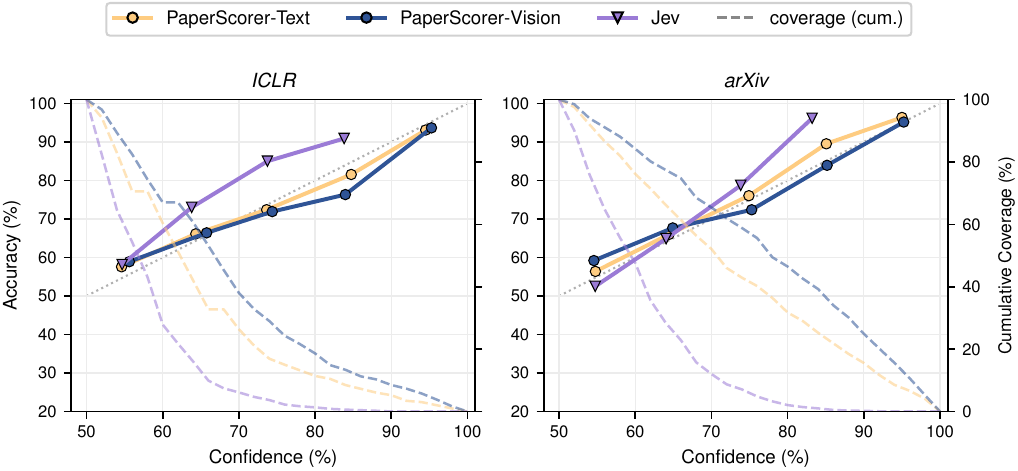}
        \vspace{-0.4em}
        \caption{\textbf{Calibration on both datasets.} Reliability (confidence vs.\ accuracy, $y{=}x$ dotted) on the balanced \dataseticlrshort{} (left) and \datasetarxivname{} (right) test sets for \systemfamily{} text and vision and for Jev, after temperature scaling. Dashed lines show the cumulative coverage (share of test papers at or above each confidence).}
        \label{fig:calibration}
    \end{figure}

    \vspace{-1.15em}
    \noindent\textbf{Confidence tracks accuracy.}
    Like most modern neural networks \citep{Guo_2017}, ours are mildly overconfident, with raw accept
    probabilities higher than their accuracy warrants, so we correct them with temperature
    scaling \citep{Guo_2017}. For each paper $i$ we take the model's \emph{accept log-odds}
    $\ell_i=\log P(\text{Accept})-\log P(\text{Reject})$ from the decision-token log-probabilities, and fit
    a temperature $\tau$ for the calibrated accept probability $p_{\mathrm{cal}}$ by minimizing its
    negative log-likelihood, where $z_i\in\{0,1\}$ is the decision:
    \[
        p_{\mathrm{cal}}=\sigma\big(\ell_i/\tau+\log(\pi/(1-\pi))\big),\qquad
        \mathcal{L}(\tau)=-\sum_i \big[z_i\log p_{\mathrm{cal}}+(1-z_i)\log(1-p_{\mathrm{cal}})\big].
    \]
    Here, $\pi$ is the acceptance rate of the scored papers. We fit $\tau$ on a natural-rate validation
    split with $\pi$ set to its acceptance rate, so the shift handles the base rate and $\tau$ only sets
    sharpness. On our balanced test sets $\pi=0.5$ and the shift vanishes. Neither changes a
    decision (Accept iff $\ell_i\ge 0$, i.e.\ $p_{\mathrm{cal}}\ge\pi$), so Table~\ref{tab:main_results} is unchanged.
    Confidence is $c=p_{\mathrm{cal}}$ for an Accept and $1-p_{\mathrm{cal}}$ for a Reject. Jev's scores already sit near the real acceptance rate, so we fit its $\tau$ without the shift and apply the correction in reverse on our 50/50 test sets.

    After scaling, both \systemfamily{} models track the $y{=}x$ line on both datasets (Figure~\ref{fig:calibration}), so their confidence matches their accuracy, and Jev, corrected the same way, is roughly calibrated too (ECE $.05$ and $.02$, vs.\ $.02$--$.03$ for ours). The difference is how often each model is confident (dashed coverage curves). At 70\% confidence or more, \systemfamily{} covers about a third of \dataseticlrshort{} papers and over half of \datasetarxivname{} papers, at nearly 80\% and 85\% accuracy, while Jev covers only 6\% and 12\%.

    \textbf{Deployment at the real acceptance rate.} Most papers are rejected, and a venue's rate $\pi$ can be estimated from previous years. Adding $\log(\pi/(1-\pi))$ to $\ell$ gives acceptance probabilities at that rate, cutting calibration error from .13 to .02--.03. At \dataseticlrshort{}'s rate, Accept calls are right 47\% of the time and Reject calls 82\% (\appref{appendix:calibration}).

    \begin{figure*}[h!]
        \centering
        \vspace{6pt}
        \subcaptionbox{Accuracy by rating percentile (\dataseticlrshort{}).\label{fig:selective_rating}}[0.48\textwidth]{%
            \includegraphics[width=\linewidth]{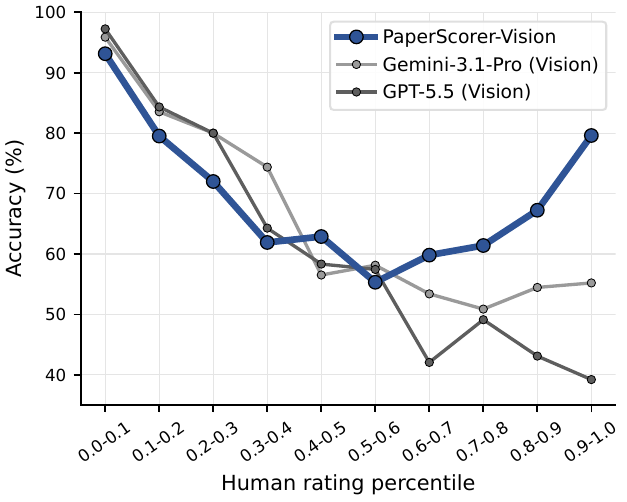}}
        \hfill
        \subcaptionbox{Rating $\times$ confidence $\to$ accuracy (\dataseticlrshort{}).\label{fig:selective_surface}}[0.48\textwidth]{%
            \includegraphics[width=0.82\linewidth]{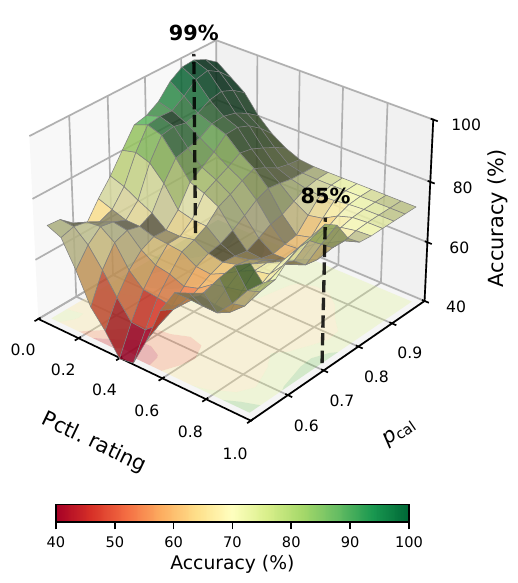}}
        \caption{\textbf{Accuracy of \systemvision{} across reviewer rating.}
        Rating percentile is a paper's percentile rank in $[0,1]$ among the papers of its own
        conference and year.
        \textbf{(a)}~Frontier models recognize low-rated papers but fail on the high-rated ones, whereas
        \systemvision{} is accurate at both tails, while all models are weak in the middle. \textbf{(b)}~The
        same shape as a confidence/rating surface, with a deep dip in the $[0.4,0.5]$ rating slice and the
        high-accuracy, high-confidence region concentrated on the tail papers.}
        \label{fig:selective_reliability}
    \end{figure*}

    \noindent\textbf{Selective reliability.}
    \dataseticlrshort{} has human ground-truth ratings for each paper, \looseness=-1 so we can measure accuracy at each level of the human rating
    (Figure~\ref{fig:selective_reliability}a). Every model is most accurate at the lowest-rated papers and
    near chance in the middle of the rating range. The models differ at the top end, where \systemvision{} stays
    accurate on the highest-rated papers while GPT and Gemini fall off, so the frontier models accurately
    predict the lowest-rated papers but fail on the highest-rated ones. Conditioning jointly on rating and confidence
    (Figure~\ref{fig:selective_reliability}b) shows the same shape, with accuracy peaking on the
    confident extremes and collapsing in the middle, where none of the evaluated models finds enough signal.

\begin{figure}[t]
        \centering
        \includegraphics[width=0.86\textwidth]{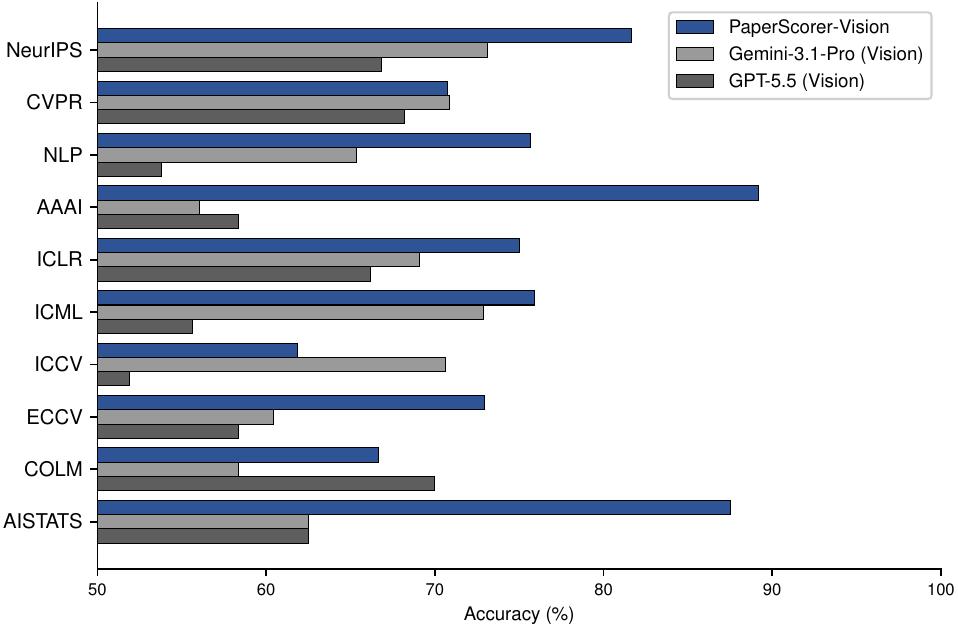}
        \vspace{-0.35em}
        \caption{\textbf{Accuracy by venue.} The best frontier vision models (Gemini-3.1-Pro, GPT-5.5)
        against \systemvision{} on \datasetarxivname{} by venue. \systemfamily{} leads on all high-coverage venues but CVPR and ICCV, where Gemini ties or leads.}
        \label{fig:frontier_by_conference}
        \vspace{-0.9em}
    \end{figure}

    \noindent\textbf{Performance across conferences.}
    Accuracy varies sharply by venue (Figure~\ref{fig:frontier_by_conference}).
    \looseness=-1 On six of the seven highest-coverage venues, \systemvision{} matches or beats the best
    frontier model (Gemini-3.1-Pro or GPT-5.5). The exceptions are the vision-heavy CV venues, where the frontier
    models perform particularly well. Gemini ties \systemfamily{} on CVPR and is the most accurate model on ICCV,
    the only high-coverage venue where a frontier model wins. AAAI is the other extreme, with \systemfamily{} near $90$ while
    both frontier models fall close to chance, the widest gap. AAAI accepts and rejects run equally long, so the gap is not a length effect;
    \systemfamily{} ranks them more cleanly here than anywhere else. In-domain training therefore gives \systemfamily{} a clear edge everywhere except the visual CV
    papers, where strong general vision models compete.

    \section{Related Work}

    \looseness=-1 \textbf{Paper acceptance prediction and dataset construction.}
    Paper outcome prediction predates modern language models, with early work using bibliometric, lexical, and handcrafted signals \citep{yan2011citation, su2024words, xu2025forecite}. Recent datasets and methods move closer to paper-level accept/reject prediction, including PeerRead~\citep{kang2018dataset}, Deep Paper Gestalt~\citep{huang2018deeppapergestalt}, DeepReview~\citep{liu2025deepreview}, ScholarPeer~\citep{goyal2026scholarpeercontextawaremultiagentframework}, and Sakana's AI Reviewer~\citep{lu2024ai}. The main challenge is label provenance, since prior sources mix arXiv, LaTeX, and OpenReview artifacts~\citep{liu2025deepreview}, compare camera-ready accepts against anonymized rejects~\citep{lu2024ai}, or use workshop papers as reject proxies~\citep{huang2018deeppapergestalt}. These choices can make labels recoverable from source, venue, or revision artifacts (audited in \appref{appendix:dataset-shortcuts}); we instead construct datasets in which the label is recoverable only from the paper's content.

    \textbf{Review generation and reviewer agents.}
    A parallel line of work studies critique generation, review quality, reviewer agreement, and multi-agent review simulation rather than direct final-decision prediction. Reviewer2~\citep{tyser2024reviewer2} analyzes reviewer behavior and agreement. DeepReview~\citep{liu2025deepreview}, ReviewRL~\citep{zeng2025reviewrl}, and OpenReviewer~\citep{idahl2025openreviewer} generate review-style outputs, while AgentReview~\citep{jin2024agentreview}, the Stanford Agentic Reviewer~\citep{zhang2025agenticreviewer}, MARG~\citep{d2024marg}, MAMORX~\citep{wang2024mamorx}, and ScholarPeer~\citep{goyal2026scholarpeercontextawaremultiagentframework} move toward agentic reviewing. These systems need reliable decision priors because generated critiques can be persuasive, yet when conditioned on the wrong prior they can lead to drastically different comments and decisions. We therefore ask whether a calibrated acceptance probability improves agentic reviewing (\appref{sec:application}).


    \textbf{AI-assisted scientific workflows.}
    Automated reviewing is also becoming part of larger scientific workflows. Systems such as The AI Scientist v1/v2~\citep{lu2024ai,yamada2025ai}, CycleResearcher~\citep{weng2024cycle}, FunSearch~\citep{romera2024mathematical}, AlphaEvolve~\citep{deepmind2025alphaevolve}, and execution-grounded research agents~\citep{si2026towards} use model-based evaluation to filter ideas, judge generated artifacts, or steer search. AstaBench~\citep{bragg2026astabench} formalizes evaluation for the full idea-to-paper setting. In these pipelines, reviewer models act as selection mechanisms rather than summarizers, so systematic accept bias or poor calibration can shape which research directions are pursued. This motivates \systemname{} as a calibrated decision prior for both retrieval and review assistance (\appref{sec:applications}).

    \vspace{-0.5em}
    \section{Conclusion and Applications}
    We introduce \systemfamily{}, a family of text and vision models that predict a paper's final accept/reject decision from its content alone. Tested on \dataseticlrshort{} and source-derived \datasetarxivname{} papers, supervised binary prediction recovers a strong paper-level signal, and 7B vision \systemfamily{} models outperform frontier reasoning and DeepReview baselines on decision accuracy and ranking, while producing calibrated acceptance probabilities. We find that paper artifacts contain substantial recoverable decision signal. Screenshots of the paper preserve layout, figure, and presentation cues that markdown extraction tends to discard.

    \underline{\systemfamily{} is not a replacement for peer review or a universal measure of scientific value}. Its best use is as a calibrated estimate of acceptance probability that larger scientific workflows can use as a prior, which we demonstrate in two settings (\appref{sec:applications}). In literature retrieval, combining dense semantic retrieval scores from an embedding model with \systemvision{}'s acceptance score surfaces papers that are both semantically relevant and likely to be accepted. On the LitSearch benchmark, this improves ground-truth citation recall in the top-10 results by \textbf{8.7\%}. In agentic reviewing, conditioning Claude Code on the \systemvision{} prior raises decision accuracy from 63.0\% to 70.6\% and improves alignment with human reviews, mainly by correcting the agent's incorrect acceptance predictions.

    \section{Limitations and Future Work}
    We cannot yet say \emph{why} the model favors one paper over another. Our attention maps and section-edit tests (\appref{appendix:where_signal}) show what it reads, but they do not \emph{establish causality}, and they rely on last-layer attention from only a few papers. The idea that vision performs better by capturing a paper's overall look remains a conjecture, supported but not settled by our compression and modality results. The model also weakens on the newest papers, where our data is thin. By '26, some decisions reflect AI help as much as human judgment, and dropping '26 from training lowers accuracy (\appref{appendix:data_mixture}). That year is also unusual, with shorter rebuttals, the narrowest accept-reject rating gap we have seen, and \datasetarxivname{} labels from only a few venues. This should ease as more '26 papers appear, and \systemfamily{} can be retrained through our pipeline.

    \textbf{What the score captures.} A high $p_{\mathrm{accept}}$ predicts the accept/reject outcome; it is not a measure of novelty or long-term impact, and it comes with no explanation of \emph{why} (though it does have correlations to reviewer rating and citation count). On impact, the one slice with both full reviewer ratings and citations, ICLR~'25, is encouraging, as $p_{\mathrm{accept}}$ tracks later citations about as well as the human review score does ($\rho\approx.25$ for both, \appref{appendix:statistical_significance}), though one conference-year and small sample make this a preliminary result. The model itself provides no explanation for its prediction, so any reasoning for a decision must be generated afterward, by the model or an agentic reviewer. Conditioning a reviewer on $p_{\mathrm{accept}}$ raises agreement with human reviews (\appref{sec:agent_review_application}), but such post-hoc rationales can hallucinate. A more robust path likely needs multi-agent generation with verification loops.

    \textbf{Future work.} The clearest next steps are expanding high-fidelity datasets across more venues, improving borderline-paper modeling, and developing richer supervision than a single binary outcome, especially for cases where reviewer disagreement and venue-specific norms matter.

    \section*{Acknowledgements}
    This work was supported by Princeton Research Computing resources, including the Della high-performance computing cluster. We additionally thank Harish Krishnakumar and Linrong Cai for their help in experimental execution, Taiming Lu and Haozhe Chen for helpful discussions during the project, and Shardul Kothapalli for conversations that led to the literature search application.

    \bibliographystyle{plainnat}
    \bibliography{main}

\newpage
\appendix

\counterwithin{table}{section}
\counterwithin{figure}{section}
\renewcommand{\thetable}{\thesection.\arabic{table}}
\renewcommand{\thefigure}{\thesection.\arabic{figure}}
\setcounter{section}{0}
\setcounter{table}{0}
\setcounter{figure}{0}

{\huge\bfseries Appendix}\par
\vspace{0.5cm}
\noindent This appendix provides supporting derivations, dataset documentation, extended evaluation, and training/inference details for the main paper:
\begin{itemize}[noitemsep,topsep=2pt,parsep=0pt,partopsep=0pt,leftmargin=*]\raggedright
    \item \S~\ref{appendix:bayes} derives the Bayesian-optimal accuracy ceiling that anchors the 80.2\% reference point in \S\ref{sec:probform}.
    \item \S~\ref{app:data} documents data provenance and the artifact-shortcut audit across our datasets and prior work.
    \item \S~\ref{app:tabular_features} catalogs handcrafted feature probes and accept/reject distributions across ICLR and \datasetarxivname{}.
    \item \S~\ref{appendix:memory_probe} tests whether frontier models can recall ground-truth decisions through label-recall and content-completion probes.
    \item \S~\ref{appendix:backbone_choice} compares Qwen2.5 and Qwen3 text backbones on \dataseticlrshort{} alongside their knowledge cutoffs.
    \item \S~\ref{app:training_variations} reports data-mixture and input-representation ablations plus the reverse-order and prior diagnostics.
    \item \S~\ref{app:generalization} reports cross-source transfer between \dataseticlrshort{} and \datasetarxivname{}, including the \datasetarxiviclrname{} subset.
    \item \S~\ref{app:prompting} records venue-specific prompts and prompt-correction studies for base open-source and frontier models.
    \item \S~\ref{app:training} covers SFT settings, the GRPO policy, and post-hoc temperature calibration.
    \item \S~\ref{appendix:statistical_significance} reports paired bootstrap intervals, rating/citation diagnostics, and a frontier repeatability audit.
    \item \S~\ref{sec:applications} details the LitSearch reranking and agentic-reviewing applications of the calibrated \systemfamily{} prior, including the Claude Code reviewer scaffolding and Codex grading rubric.
    \item \S~\ref{app:signal} analyzes section-level attention maps and the four-commit Codex edit experiments.
\end{itemize}

\begingroup
\hypersetup{linkcolor=black}
\etoctoccontentsline{part}{}
\etocstandardlines
\etocsettocstyle{}{}
\etocsetnexttocdepth{subsection}
\localtableofcontents
\endgroup

\newpage

    \section{Bayesian Optimal Accuracy Derivation}\label{appendix:bayes}

    We follow the NeurIPS '14 / '21 consistency experiment~\citep{beygelzimer2021consistency}, where each paper $i$ is reviewed by two committees $j \in \{1,2\}$ with independent decisions conditional on a latent acceptance probability $p_i \sim F$ (step~\ref{eq:bayes_setup}). Here $F$ is the (unknown) population distribution of per-paper acceptance probabilities, whose first two moments we estimate from the consistency experiment below. The experiment's reported acceptance rate of ${\approx}\,26\%$ fixes the first moment (step~\ref{eq:bayes_Ep}), and its disagreement rate of ${\approx}\,23\%$ corresponds to $\mathbb{E}[2p(1-p)]$ (step~\ref{eq:bayes_disagree}), which expands into the second moment and variance (steps~\ref{eq:bayes_Ep2},~\ref{eq:bayes_var}). Matching these to a Beta family (steps~\ref{eq:bayes_beta},~\ref{eq:bayes_ab}) gives $(\alpha,\beta)\approx(0.39,1.10)$. Under the natural acceptance distribution, the Bayes-optimal predictor accepts when $p>0.5$, giving $\mathrm{Acc}^{\star}\approx 83.3\%$ (step~\ref{eq:bayes_acc_natural}).

    \begin{align*}
        D_j^{(i)} &\sim \mathrm{Bernoulli}(p_i), \qquad j \in \{1,2\} \tag{a}\label{eq:bayes_setup}\\
        \mathbb{E}[p] &= 0.26 \tag{b}\label{eq:bayes_Ep}\\
        \mathbb{E}[2p(1-p)] &= 0.23 \tag{c}\label{eq:bayes_disagree}\\
        2\mathbb{E}[p] - 2\mathbb{E}[p^2] &= 0.23 \;\Longrightarrow\; \mathbb{E}[p^2] = \frac{0.52 - 0.23}{2} = 0.145 \tag{d}\label{eq:bayes_Ep2}\\
        \mathrm{Var}(p) &= 0.145 - 0.26^2 \approx 0.077 \tag{e}\label{eq:bayes_var}\\
        p &\sim \mathrm{Beta}(\alpha,\beta), \qquad \tfrac{\alpha}{\alpha+\beta} = 0.26, \;\; \tfrac{\alpha(\alpha+1)}{(\alpha+\beta)(\alpha+\beta+1)} = 0.145 \tag{f}\label{eq:bayes_beta}\\
        (\alpha,\beta) &\approx (0.39, 1.10) \tag{g}\label{eq:bayes_ab}\\
        \mathrm{Acc}^{\star} &= \mathbb{E}[\max(p,1-p)] \approx 83.3\% \tag{h}\label{eq:bayes_acc_natural}\\[3pt]
        p \mid D{=}1 &\sim \mathrm{Beta}(1.39, 1.10), \quad p \mid D{=}0 \sim \mathrm{Beta}(0.39, 2.10) \tag{i}\label{eq:bayes_post}\\
        P(\mathrm{accept}\mid p) &= \tfrac{\beta p}{\beta p + \alpha(1-p)} = 0.5 \;\;\text{at}\;\; p = \tfrac{\alpha}{\alpha+\beta} = 0.26 \tag{j}\label{eq:bayes_postaccept}\\
        \mathrm{Acc}^{\star}_{\mathrm{balanced}} &\approx 80.2\% \tag{k}\label{eq:bayes_acc_balanced}
    \end{align*}

    Our evaluation uses a balanced 50/50 pool of accepted and rejected papers, so the relevant ceiling differs. Beta-Bernoulli conjugacy gives the conditional posteriors in step~\ref{eq:bayes_post}, and under a uniform 50/50 class prior the posterior accept probability (step~\ref{eq:bayes_postaccept}) equals $0.5$ exactly at $p=\alpha/(\alpha+\beta)=0.26$. The Bayes-optimal rule in the balanced setting therefore predicts accept when $p>0.26$. Averaging over the two conditional Beta distributions yields the balanced ceiling in step~\ref{eq:bayes_acc_balanced}, the value reported in the main text.

\section{Data Curation Details}\label{app:data}

    \subsection{Shortcuts in Prior Datasets}\label{appendix:dataset-shortcuts}

    Existing accept/reject datasets often mix artifacts in ways that make labels recoverable from source, venue, or revision artifacts rather than from the paper's content. For instance, Deep Paper Gestalt~\citep{huang2018deeppapergestalt} pairs accepted CVPR papers against workshop papers as a reject pool, but the workshop format alone (shorter page count, distinct headers) is enough to shortcut the prediction problem without ever reading the paper. We compare three public datasets, DeepReview '24--'25~\citep{liu2025deepreview}, Sakana's AI Scientist '22~\citep{lu2024ai}, and our raw ICLR '24 (before normalization), against the selected \systemfamily{} years. We tested several signals defined in \Cref{tab:dataset_shortcut_signals}, and \Cref{tab:dataset_shortcuts} reports accept/reject percentages (A/R) for each.

    \begin{table}[H]
        \centering
        \footnotesize
        \setlength{\tabcolsep}{4pt}
        \renewcommand{\arraystretch}{1.1}
        \begin{tabular}{@{}p{0.30\textwidth}p{0.66\textwidth}@{}}
            \toprule
            \textbf{Signal} & \textbf{Definition} \\
            \midrule
            Markdown / LaTeX source & Native source artifacts (e.g.\ \texttt{\textbackslash begin\{equation\}}, \texttt{\textbackslash cite}) survive in the text, indicating the dataset retained LaTeX source rather than rendering to PDF. \\
            Author names visible & Non-anonymous author block present before the abstract. \\
            Affil.\ before abstract & Institutional affiliation visible in the front matter before the abstract. \\
            Has ack. & An Acknowledgments section heading is present anywhere in the paper. \\
            Affil.\ in ack. & Institutional affiliation strings appear inside the Acknowledgments section. \\
            GitHub in abstract & A GitHub URL appears in the abstract text. \\
            \bottomrule
        \end{tabular}
        \caption{\textbf{Identity-leaking signals tested in the artifact shortcut audit.} Each signal is a heuristic that, if positively correlated with the accept/reject label, lets a classifier shortcut the prediction problem from formatting alone.}
        \label{tab:dataset_shortcut_signals}
    \end{table}

    Among the comparators, DeepReview '24 carries source for 78.7\% of accepts but only 41.2\% of rejects, a strong leakage shortcut. AI Scientist '22 has 100\% of accept papers with visible author names against essentially zero reject papers, because the rejects come from anonymized OpenReview submissions while the accepts are camera-ready PDFs. Our selected years stay at most 0.92\% A / 1.73\% R on any signal. Our audit shows that sourcing papers from the latest pre-camera-ready rebuttal artifact does not suffer from any of these shortcuts.

    \begin{table}[H]
        \centering
        \footnotesize
        \setlength{\tabcolsep}{3pt}
        \renewcommand{\arraystretch}{1.1}
        \begin{tabular*}{\textwidth}{@{\extracolsep{\fill}}lccccc}
            \toprule
            \textbf{Signal} & \shortstack{\textbf{DeepReview}\\\textbf{'24}} & \shortstack{\textbf{DeepReview}\\\textbf{'25}} & \shortstack{\textbf{AI Scientist}\\\textbf{'22}} & \shortstack{\textbf{Ours raw}\\\textbf{'24}} & \shortstack{\textbf{Ours selected}\\\textbf{years}} \\
            \midrule
            Markdown / LaTeX source & 78.7 / 41.2 & 53.0 / 37.8 & 0.0 / 0.0 & 0.0 / 0.0 & 0.01 / 0.0 \\
            Author names visible & 0.0 / 0.0 & 0.0 / 0.0 & 100.0 / 0.6 & 99.7 / 0.4 & 0.0 / 0.0 \\
            Affil.\ before abs. & 0.7 / 0.2 & 0.4 / 0.2 & 53.8 / 0.0 & 51.1 / 0.04 & 0.0 / 0.0 \\
            Has ack. & 16.4 / 2.8 & 4.9 / 2.8 & 74.5 / 2.8 & 73.9 / 1.7 & 0.92 / 1.73 \\
            Affil.\ in ack. & 4.4 / 1.1 & 1.3 / 0.6 & 48.9 / 0.9 & 31.8 / 0.1 & 0.07 / 0.20 \\
            GitHub in abs. & 4.0 / 0.5 & 0.5 / 0.5 & 0.0 / 0.0 & 22.5 / 1.1 & 0.0 / 0.0 \\
            \bottomrule
        \end{tabular*}
        \caption{\textbf{Artifact shortcut audit across public paper datasets.} We measure accept/reject percentages (A/R) of source, author, affiliation, acknowledgment, and code-leakage signals. The selected \systemfamily{} years avoid these shortcuts, while raw ICLR '24 shows why camera-ready accepts remain problematic.}
        \label{tab:dataset_shortcuts}
    \end{table}

    \subsection{Data Provenance and Ensuring Anonymity}\label{appendix:dataset-notes}

    \textbf{Artifact availability across OpenReview venues.}
    We audit which submission, rebuttal, camera-ready, and reject artifacts are public across the
    OpenReview venues and years we considered, and use the result to decide which year/version of
    each paper to keep. Yellow cells in \Cref{tab:artifact_availability} mark the coverage used in
    \dataset{}: for ICLR we keep '20--'23 (initial-submission and rebuttal artifacts) and '25--'26
    (latest-rebuttal PDFs that are public without the camera-ready transition), and we exclude '24
    entirely. NeurIPS, ICML, and COLM publish only accepts on OpenReview, so we only source ICLR submissions from OpenReview.

\begin{table}[b!]
        \centering
        \scriptsize
        \setlength{\tabcolsep}{3pt}
        \renewcommand{\arraystretch}{1.02}
        \begin{tabular*}{\textwidth}{@{\extracolsep{\fill}}llcccccccccc}
\toprule
\multirow{2}{*}{\textbf{Decision}} & \multirow{2}{*}{\textbf{Artifact}} & \multicolumn{5}{c}{\textbf{ICLR}} & \textbf{NeurIPS} & \textbf{ICML} & \textbf{ICML} & \textbf{COLM} \\
\cmidrule(lr){3-7}
& & \textbf{'17--'19} & \textbf{'20--'23} & \textbf{'24} & \textbf{'25} & \textbf{'26} & \textbf{'21--'25} & \textbf{'23--'24} & \textbf{'25} & \textbf{'24--'25} \\
\midrule
\multirow{3}{*}{\textbf{Accepts}}
& Initial         & Y & Y & N & N & N$^\dagger$ & 21--22 & N & N & N \\
& Latest rebuttal & Y & \cellcolor{trainyellow}Y & N & \cellcolor{trainyellow}Y & \cellcolor{trainyellow}Y$^\dagger$ & Y & N & N & N \\
& Camera-ready    & Y & Y & Y & Y & Y & 21--22 & Y & Y & Y \\
\midrule
\multirow{3}{*}{\textbf{Rejects}}
& Initial         & Y & Y & N & N & N$^\dagger$ & N & N & N & N \\
& Latest rebuttal & Y & \cellcolor{trainyellow}Y & Y$^\ast$ & \cellcolor{trainyellow}Y & \cellcolor{trainyellow}Y$^\dagger$ & Y & N & Y & N \\
& Public rejects  & Y & Y & Y & Y & Y & Opt. & N & Opt. & N \\
\bottomrule
        \end{tabular*}
        \caption{\textbf{OpenReview artifact availability.} We audit which submission, rebuttal, camera-ready, and reject artifacts are public across venues and years. Yellow cells denote the year/artifact coverage used in \dataset{}; `Y' = public, `N' = unavailable, and `Opt.' = author-opt-in rejects.}
        \label{tab:artifact_availability}
    \end{table}

\begin{figure}[H]
        \centering
        \includegraphics[width=0.92\textwidth]{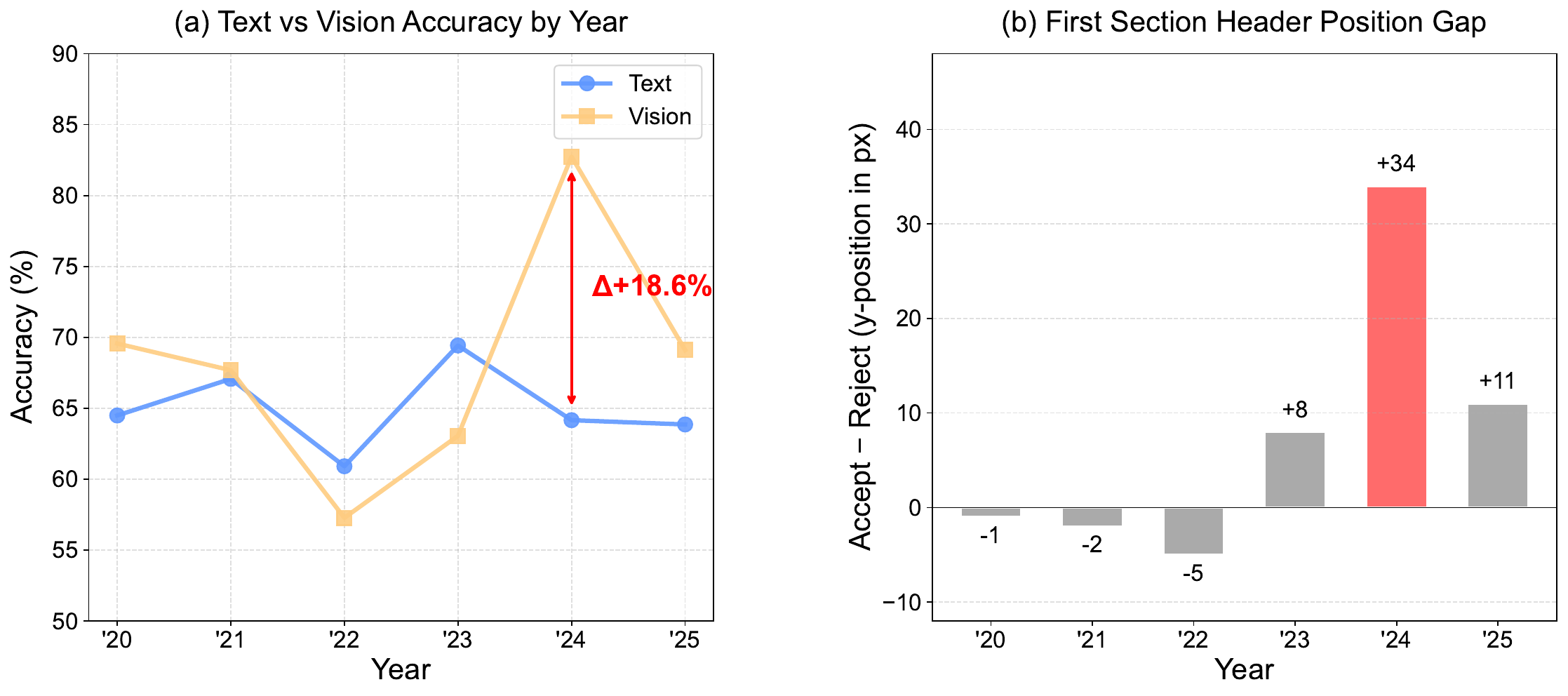}
        \caption{\textbf{Why normalized ICLR '24 remains shortcut-prone for vision.} We strip identity-bearing spans from text and whiten the same regions in rendered pages, but the camera-ready transition still shifts page layout and section start positions. Text mostly tolerates this cleanup; vision does not.}
        \label{fig:2024_bias}
    \end{figure}

    \textbf{Why we exclude ICLR '24.}
    For '24, neither the initial submission nor the latest rebuttal PDF is publicly available, so the only artifact we can recover is the camera-ready version of accepts paired with the latest rebuttal of rejects. We attempted to recover an anonymized version of ICLR '24 by performing manual anonymization on the camera-ready PDF. We strip identity-bearing spans from text and whiten the same regions in rendered pages, but the camera-ready transition still shifts page layout and section start positions (\Cref{fig:2024_bias}). Vision picks this up. Even after normalization, the model remains unusually strong on '24, indicating that residual layout cues survive whiteout. MinerU confirms the source of the problem. Across years, the mean $y$-position of the introduction start is roughly matched between accepts and rejects, except in '24 where the gap is notable. This is consistent with author-block expansion, GitHub insertion, and whitespace changes after masking, which is why '24 is excluded from the benchmark and training pool.

    \textbf{ICLR submission metadata across years.}
    Beyond artifact availability, the venues themselves change over time. \Cref{tab:iclr_metadata} compares page limits, rating scales, and artifact availability across ICLR '20--'26. The rating scale shifts repeatedly --- 4-point in '20, 10-point in '21, then 6-point from '22 onward --- so we normalize all reviewer ratings into percentile space before computing any rating-aligned metric. Page limits also creep up across the rebuttal and camera-ready windows; we account for this with explicit length filters per year.

    \begin{table}[H]
        \centering
        \footnotesize
        \setlength{\tabcolsep}{2pt}
        \renewcommand{\arraystretch}{1.1}
        \begin{tabular*}{\textwidth}{@{\extracolsep{\fill}}lccccccc}
            \toprule
            \textbf{Signal} & \textbf{'20} & \textbf{'21} & \textbf{'22} & \textbf{'23} & \textbf{'24} & \textbf{'25} & \textbf{'26} \\
            \midrule
            Source & CFP+OR & CFP+OR & CFP+OR & CFP+OR & CFP+OR & CFP+OR & Guide+DL \\
            Min len. & -- & -- & -- & -- & -- & 6 & None \\
            Max len. & 10 & 8 & 9 & 9 & 9 & 10 & 9 \\
            Max reb./CR len. & 10 & 10 & 10 & 10 & 11 & 11 & 11 \\
            Rating scale & 4-pt & 10-pt & 6-pt & 6-pt & 6-pt & 6-pt & 6-pt \\
            Orig. sub. avail. & Y & Y & Y & Y & N & N & N \\
            Latest reb. avail. & Y & Y & Y & Y & N & Y & Y \\
            Camera-ready avail. & Y & Y & Y & Y & Y & Y & N \\
            \bottomrule
        \end{tabular*}
        \caption{\textbf{ICLR year-by-year submission metadata.} We compare page limits, rating scales, and artifact availability to justify the \dataset{} year selection. Rebuttal/camera-ready limits include Acknowledgments, Ethics, and Reproducibility; CFP = Call for Papers, OR = OpenReview, and DL = downloaded during rebuttal before camera-ready.}
        \label{tab:iclr_metadata}
    \end{table}

    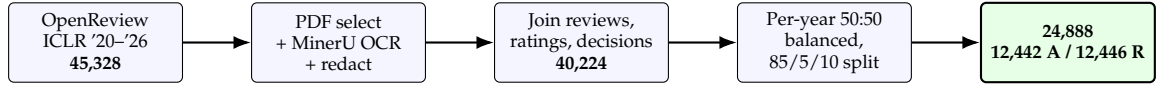
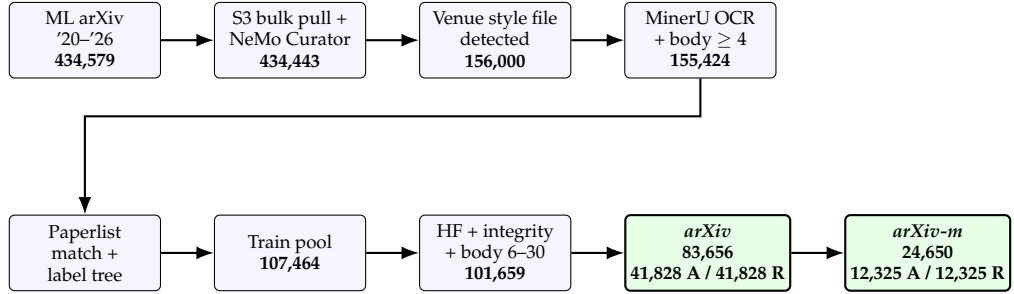
\begin{figure}[H]
        \centering
        \captionsetup[subfigure]{justification=raggedright,singlelinecheck=false}
        \begin{subfigure}{\linewidth}
            \centering
            \scriptsize
            \begin{tikzpicture}[
                node distance=10mm and 9mm,
                stage/.style={draw, rectangle, rounded corners=2pt, minimum width=2.3cm, minimum height=1.05cm, align=center, fill=blue!4, inner sep=2pt, font=\scriptsize},
                final/.style={stage, fill=green!10, thick},
                arr/.style={->, >=Latex, thick}
            ]
            \node[stage] (i1) {OpenReview\\ ICLR '20--'26\\ \textbf{45{,}328}};
            \node[stage, right=of i1] (i2) {PDF select\\ + MinerU OCR\\ + redact};
            \node[stage, right=of i2] (i3) {Join reviews,\\ ratings, decisions\\ \textbf{40{,}224}};
            \node[stage, right=of i3] (i4) {Per-year 50:50\\ balanced,\\ 85/5/10 split};
            \node[final, right=of i4] (i5) {\textbf{24{,}888}\\ \textbf{12{,}442 A / 12{,}446 R}};
            \draw[arr] (i1) -- (i2);
            \draw[arr] (i2) -- (i3);
            \draw[arr] (i3) -- (i4);
            \draw[arr] (i4) -- (i5);
            \end{tikzpicture}
            \caption{\textbf{\dataseticlrshort{} pipeline.} Decisions for both accepts and rejects are fully available on OpenReview, so the pipeline is simple --- download OpenReview PDFs and metadata, extract markdown via MinerU, normalize, and subsample. The final \textbf{24{,}888} balanced papers (\textbf{12{,}442 accept / 12{,}446 reject}, per-year 1:1 over '20--'23 and '25--'26, with '24 excluded) are used in training.}
            \label{fig:data_pipeline_iclr}
        \end{subfigure}

        \vspace{6mm}

        \begin{subfigure}{\linewidth}
            \centering
            \scriptsize
            \begin{tikzpicture}[
                node distance=8mm and 7mm,
                stage/.style={draw, rectangle, rounded corners=2pt, minimum width=2.0cm, minimum height=1.0cm, align=center, fill=blue!4, inner sep=2pt, font=\scriptsize},
                final/.style={stage, fill=green!10, thick},
                arr/.style={->, >=Latex, thick}
            ]
            \node[stage] (a1) {ML arXiv\\ '20--'26\\ \textbf{434{,}579}};
            \node[stage, right=of a1] (a2) {S3 bulk pull +\\ NeMo Curator\\ \textbf{434{,}443}};
            \node[stage, right=of a2] (a3) {Venue style file\\ detected\\ \textbf{156{,}000}};
            \node[stage, right=of a3] (a5) {MinerU OCR\\ + body $\ge 4$\\ \textbf{155{,}424}};
            \node[stage, below=18mm of a1] (a6) {Paperlist\\ match +\\ label tree};
            \node[stage, right=of a6] (a7) {Train pool\\ \textbf{107{,}464}};
            \node[stage, right=of a7] (a8) {HF + integrity\\ + body 6--30\\ \textbf{101{,}659}};
            \node[final, right=of a8] (a9) {\textbf{\datasetarxivname{}}\\ \textbf{83{,}656}\\ \textbf{41{,}828 A / 41{,}828 R}};
            \node[final, right=of a9] (a10) {\textbf{\datasetarxivmediumname{}}\\ \textbf{24{,}650}\\ \textbf{12{,}325 A / 12{,}325 R}};
            \draw[arr] (a1) -- (a2);
            \draw[arr] (a2) -- (a3);
            \draw[arr] (a3) -- (a5);
            \draw[arr] (a5.south) -- ++(0,-5mm) -| (a6.north);
            \draw[arr] (a6) -- (a7);
            \draw[arr] (a7) -- (a8);
            \draw[arr] (a8) -- (a9);
            \draw[arr] (a9) -- (a10);
            \end{tikzpicture}
            \caption{\textbf{\datasetarxivname{} pipeline.} Decisions are derived externally from public paperlists~\citep{su2025papercopilot}. The dominant drop is venue-template detection, since without a recognized venue style file we cannot trust the venue tag and cannot assign a label. Per-venue 50:50 balancing produces the \textbf{\datasetarxivname{}} dataset (\textbf{83{,}656} papers, \textbf{41{,}828 A / 41{,}828 R}) spanning NeurIPS, CVPR, NLP, AAAI, ICLR, ICML, and other ML venues. Proportional per-venue downsampling produces the smaller \textbf{\datasetarxivmediumname{}} dataset (\textbf{24{,}650} papers, \textbf{12{,}325 A / 12{,}325 R}) used throughout the paper for compute-matched ablations.}
            \label{fig:data_pipeline_arxiv}
        \end{subfigure}

        \caption{\textbf{\dataseticlrshort{} and \datasetarxivname{} data pipelines.} The \dataseticlrshort{} pipeline is short because OpenReview publishes both accept and reject decisions for every submission. The \datasetarxivname{} pipeline is long because both the venue and the decision must be recovered from each paper's arXiv source tarball.}
        \label{fig:data_pipeline}
    \end{figure}

    \subsection{Our Data Pipeline}\label{appendix:data_pipeline}

    We run two independent workflows that produce two independent datasets. The \dataseticlrname{} workflow downloads ICLR submissions and decisions directly from OpenReview, runs MinerU, redacts identity, and subsamples. The \datasetarxivname{} workflow recovers both the venue and the decision from each paper's arXiv source tarball and matches against public paperlists. \Cref{fig:data_pipeline_iclr,fig:data_pipeline_arxiv} show the two funnels, and \Cref{tab:dataset_canonical_counts} summarizes the final canonical training pools used throughout the paper.

    \begin{table}[H]
        \centering
        \footnotesize
        \setlength{\tabcolsep}{6pt}
        \renewcommand{\arraystretch}{1.1}
        \begin{tabular}{@{}lrrrr@{}}
            \toprule
            \textbf{Source} & \textbf{Total} & \textbf{Accept} & \textbf{Reject} & \textbf{Train / Val / Test} \\
            \midrule
            \dataseticlrshort{} & 24{,}888 & 12{,}442 & 12{,}446 & 21{,}141 / 1{,}252 / 2{,}495 \\
            \datasetarxivname{} & 83{,}656 & 41{,}828 & 41{,}828 & 71{,}124 / 4{,}174 / 8{,}358 \\
            \datasetarxivmediumname{} & 24{,}650 & 12{,}325 & 12{,}325 & 20{,}962 / 1{,}232 / 2{,}456 \\
            \bottomrule
        \end{tabular}
        \caption{\textbf{Canonical training datasets.} \dataseticlrshort{} is per-year 50:50 balanced across '20--'23 and '25--'26 ('24 excluded). \datasetarxivname{} is per-venue 50:50 balanced across NeurIPS, CVPR, NLP, AAAI, ICLR, ICML, and other ML venues. \datasetarxivmediumname{} is a proportional per-venue subsample of \datasetarxivname{} used throughout the paper for compute-matched ablations.}
        \label{tab:dataset_canonical_counts}
    \end{table}

    \paragraph{\dataseticlrname{}.}
    We pull every ICLR submission in OpenReview for '20--'26 (45{,}328 papers) and, per year, the latest publicly available pre-camera-ready PDF. PDFs run through MinerU to obtain markdown and per-page images, then we redact identity-bearing spans (author block, affiliations, acknowledgments, GitHub links). After joining with reviews and decisions and applying an 85/5/10 stratified split balanced per (year, decision) cell, we obtain 24{,}888 papers ('20: 1{,}367; '21: 1{,}610; '22: 2{,}154; '23: 3{,}088; '25: 6{,}790; '26: 9{,}879).

    \paragraph{\datasetarxivname{}.}
    We start from 434{,}579 ML arXiv papers spanning '20--'26 by update date. We pull source directly from arXiv's requester-pays S3 mirror (\texttt{s3://arxiv/src/}) for every paper from January~'20 through our 05/01/'26 cutoff, and use NeMo Curator's vendored \texttt{ArxivIterator}~\citep{nemo_curator} to walk each monthly tar and recover the per-paper \texttt{.tex} contents. We then scan every \texttt{.tex} against our tracked-venue templates and keep only the tarballs that match a known venue style file. The workflow detects the submission venue from the LaTeX source via these venue-specific style files, strips identity-bearing spans directly from the LaTeX source (author block, affiliations, acknowledgments, GitHub links) and recompiles the PDF, runs MinerU, and matches each in-pool paper against every venue paperlist in the public Paper Copilot index~\citep{su2025papercopilot} within a $Y{-}5$ to $Y{+}2$ year window. Papers that template for a venue with ground-truth but find no paperlist match are tagged as implied rejects and kept, unless they trigger a workshop, short-paper, or no-ground-truth filter that drops them. Per-venue 50:50 balancing with a stratified 85/5/10 split yields the \datasetarxivname{} dataset of 83{,}656 papers spanning NeurIPS, CVPR, ACL, AAAI, ICLR, ICML, and other ML venues. Throughout the paper we use the full \datasetarxivname{} set. For compute-matched ablations, we use the subsample \datasetarxivmediumname{}, produced by proportional per-(venue, label) downsampling of \datasetarxivname{} to 24{,}650 papers.

\section{Simple Feature Analysis}\label{app:tabular_features}\label{appendix:feature-probe}

\begin{longtable}{@{}>{\raggedright\arraybackslash}p{0.22\textwidth}>{\raggedright\arraybackslash}p{0.58\textwidth}rr@{}}
\caption{\textbf{Programmatic feature inventory.} We compute each probe feature from the concatenated human-side paper text, truncated to the first 80,000 characters. In feature names, \textit{per pg.} means per estimated body page.}\label{tab:feature_inventory}\\
\toprule
\textbf{Feature} & \textbf{Definition} & $\mathbf{w}_\mathrm{ICLR}$ & $\mathbf{w}_\mathrm{arXiv}$ \\
\midrule
\endfirsthead
\toprule
\textbf{Feature} & \textbf{Definition} & $\mathbf{w}_\mathrm{ICLR}$ & $\mathbf{w}_\mathrm{arXiv}$ \\
\midrule
\endhead
Ablation mentions & Set to 1 if lowercase text contains ablation, otherwise 0. & +0.09 & +0.14 \\
Abstract sent. length & Average sentence length (in words) of the abstract. & +0.12 & $-$0.01 \\
Abstract sentence count & Number of sentences in the abstract. & +0.23 & $-$0.01 \\
Abstract words & Number of whitespace-delimited words in the abstract. & $-$0.14 & +0.04 \\
Algorithms per pg. & Count case-sensitive matches of the pattern Algorithm followed by whitespace and a digit, then divide by the estimated body-page proxy. & $-$0.03 & +0.03 \\
Appendix refs. per pg. & Count case-sensitive matches of Appendix followed by whitespace and a capital letter, then divide by the estimated body-page proxy. & +0.13 & +0.10 \\
Baseline mention & Set to 1 if lowercase text contains baseline, otherwise 0. & $\,$0.00 & +0.07 \\
Benchmark mention & Set to 1 if lowercase text contains benchmark, otherwise 0. & $-$0.01 & +0.03 \\
Citation-year count & Detect four-digit year strings in the extracted text, keep years in [1990, 2026], and compute log(1 + number of retained year strings). & +0.10 & $-$0.01 \\
Code mentions & In lowercase text, count regex matches of \texttt{github}, \texttt{code} followed by \texttt{avail}(\texttt{able}), \texttt{open} followed by \texttt{sourc}(\texttt{e}), or \texttt{implementation} followed by \texttt{avail}(\texttt{able}). & +0.04 & +0.06 \\
Confidence words & In lowercase text, count word-boundary matches of these confidence terms: clearly, obviously, significantly, substantially, dramatically, remarkably, notably, undoubtedly, certainly, demonstrate, prove, establish, confirm, superior. Divide by max(total word count, 1) and multiply by 1,000. & +0.03 & +0.02 \\
Contribution mentions & Count lowercase regex matches of words beginning with \texttt{contrib} (e.g.\ \emph{contribute}, \emph{contribution}). & $\,$0.00 & $-$0.03 \\
Cross-refs. per pg. & Count case-sensitive references to Figure, Fig., Table, Section, Appendix, Eq., or Equation followed by optional whitespace, an optional opening parenthesis or bracket, and an uppercase letter or digit; divide by the estimated body-page proxy. & +0.08 & +0.11 \\
CV topic terms & Count case-insensitive matches of these computer-vision terms: image, vision, convolutional, CNN, object detection, segmentation, pixel, ResNet, ViT, visual, ImageNet, CIFAR, recognition. Divide by max(total word count, 1) and multiply by 1,000. & +0.06 & $-$0.02 \\
Efficiency topic terms & Count case-insensitive matches of these efficiency/compression terms: pruning, quantization, distillation, compression, efficient, lightweight, mobile, edge, inference speed, FLOP. Divide by max(total word count, 1) and multiply by 1,000. & +0.01 & $-$0.02 \\
Equation refs. per pg. & Count case-sensitive equation-reference patterns: Eq or Equation followed by optional punctuation, whitespace, an optional opening parenthesis or bracket, and a digit, plus literal LaTeX \textbackslash{}eqref or \textbackslash{}ref occurrences; divide by the estimated body-page proxy. & $-$0.01 & +0.01 \\
Ethics indicator & Set to 1 if lowercase text contains ethic or broader impact, otherwise 0. & +0.01 & +0.03 \\
Fairness topic terms & Count case-insensitive matches of these fairness/ethics terms: fairness, bias, ethical, demographic, equit, disparate, protected attribute, discrimination. Divide by max(total word count, 1) and multiply by 1,000. & $\,$0.00 & $-$0.01 \\
Federated terms & Count case-insensitive matches of these federated/privacy/distributed-learning terms: federated, distributed learning, privacy, differential privacy, communication efficient. Divide by max(total word count, 1) and multiply by 1,000. & $-$0.07 & $-$0.06 \\
Figure refs. per pg. & Count case-sensitive matches of Figure or Fig followed by optional punctuation, optional whitespace, and a digit; divide by the estimated body-page proxy. & +0.18 & +0.10 \\
Generative topic terms & Count case-insensitive matches of these generative-modeling terms: GAN, generative, VAE, variational, diffusion, autoencoder, latent space, generation, synthesis, sample quality, FID. Divide by max(total word count, 1) and multiply by 1,000. & +0.03 & $\,$0.00 \\
Graph-topic terms & Count case-insensitive matches of these graph-learning terms: graph neural, GNN, node, edge, adjacency, message passing, graph convolution, knowledge graph, link prediction. Divide by max(total word count, 1) and multiply by 1,000. & $-$0.08 & $-$0.01 \\
Hedge words & In lowercase text, count word-boundary matches of these hedge terms: might, could, may, possibly, perhaps, likely, suggests, appears, seemingly, arguably, potentially, uncertain. Divide by max(total word count, 1) and multiply by 1,000. & $\,$0.00 & $-$0.02 \\
Intro fraction & If an Introduction heading is detected, count words after that heading up to the next Markdown heading line beginning with \#, and divide by max(total word count, 1); otherwise use 0. & +0.01 & +0.01 \\
Limitations indicator & Set to 1 if lowercase text contains limitation, otherwise 0. & $-$0.03 & $-$0.06 \\
Lists per pg. & Count Markdown bullet lines that begin with optional whitespace and - or *, plus numbered-list lines that begin with optional whitespace, digits, then . or ); divide by the estimated body-page proxy. & $-$0.03 & $-$0.03 \\
Long-word ratio & Strip punctuation from whitespace tokens, keep alphabetic words, and compute the share whose length is greater than 8 characters. & $-$0.03 & +0.01 \\
Math notation/1k words & Count pairs of dollar-sign math delimiters as floor(number of dollar signs / 2), divide by max(total word count, 1), and multiply by 1,000. & +0.06 & +0.19 \\
Meta-learning topic terms & Count case-insensitive matches of these meta-learning/transfer/domain-adaptation terms: meta learning, few shot, zero shot, transfer learning, domain adaptation, multi task. Divide by max(total word count, 1) and multiply by 1,000. & $-$0.01 & +0.02 \\
Multimodal terms & Count case-insensitive matches of these multimodal/vision-language terms: multimodal, cross modal, vision language, CLIP, image text, visual question. Divide by max(total word count, 1) and multiply by 1,000. & $-$0.01 & $\,$0.00 \\
Negative words & In lowercase text, count word-boundary matches of these negative terms: fail, limitation, drawback, weakness, shortcoming, disadvantage, unable, cannot, lack, insufficient, poor, worse, degrade, suffer. Divide by max(total word count, 1) and multiply by 1,000. & +0.04 & +0.01 \\
NLP topic terms & Count case-insensitive matches of these NLP/language-modeling terms: NLP, natural language, text classification, sentiment, translation, language model, transformer, BERT, GPT, token, vocabulary, corpus, parsing, NER, named entity. Divide by max(total word count, 1) and multiply by 1,000. & +0.04 & +0.03 \\
Novelty mentions & Count lowercase regex matches of the word novel. & +0.01 & +0.01 \\
Optimization topic terms & Count case-insensitive matches of these optimization terms: optimization, convergence, SGD, Adam, learning rate, gradient descent, momentum, regularization, batch size, weight decay. Divide by max(total word count, 1) and multiply by 1,000. & $-$0.02 & $-$0.02 \\
Paper length & Count whitespace-delimited words in the extracted text and compute log(1 + word count). & +0.47 & +0.48 \\
Positive words & In lowercase text, count word-boundary matches of these positive terms: outperform, improve, achieve, superior, advantage, benefit, effective, efficient, robust, strong, excellent, promising, remarkable, impressive, advance. Divide by max(total word count, 1) and multiply by 1,000. & +0.02 & $\,$0.00 \\
Proposal mentions & Count lowercase regex matches of words beginning with propos. & $-$0.05 & $-$0.01 \\
Rare-word ratio & From the first 5,000 lowercased alphabetic words, compute the number of unique words that appear exactly once divided by max(number of unique words, 1). & $-$0.14 & $-$0.06 \\
Recent citation share & Detect four-digit year strings in the extracted text, keep years in [1990, 2026], and compute count(years >= 2024) divided by max(number of retained year strings, 1). & +0.09 & $-$0.15 \\
Related work & Set to 1 if lowercase text contains related work, otherwise 0. & +0.02 & +0.09 \\
Reproducibility indicator & Set to 1 if lowercase text matches reproduc, code followed later by avail, or open followed by one character then sourc, otherwise 0. & $-$0.02 & $-$0.06 \\
RL terms/1k words & Count case-insensitive matches of these reinforcement-learning terms: reinforcement learning, reward, policy gradient, MDP, Q-learning, actor critic, environment, agent, exploration, exploitation, Atari, MuJoCo. Divide by max(total word count, 1) and multiply by 1,000. & $-$0.02 & $-$0.07 \\
Sections per pg. & Count Markdown headings that start a line with one to three hash marks followed by whitespace, then divide by the estimated body-page proxy. & $-$0.08 & $-$0.15 \\
Self-references/1k words & Count these lowercase phrases: our method, our approach, our model, our framework, our algorithm, our technique, our work, our paper, and our contribution. Divide by max(total word count, 1) and multiply by 1,000. & +0.01 & +0.08 \\
Self-supervised topic terms & Count case-insensitive matches of these self-supervised/contrastive-learning terms: self supervised, contrastive, SimCLR, BYOL, MoCo, pretext task, representation learning, unsupervised pre. Divide by max(total word count, 1) and multiply by 1,000. & +0.03 & +0.02 \\
Sentence length & Split the extracted text on periods, exclamation marks, and question marks, keep spans longer than 10 characters, and compute total whitespace-delimited word count divided by max(number of retained spans, 1). & $-$0.18 & $-$0.24 \\
Sentence-length variation & Split the extracted text into retained sentences as above, compute the standard deviation of sentence word counts, and divide by max(avg\_sentence\_length, 1). & +0.02 & +0.04 \\
Significance mentions & Set to 1 if lowercase text matches statistic followed later by signific, p followed by one character then value, confidence interval, standard deviation, or error bar, otherwise 0. & $-$0.01 & $-$0.03 \\
Speech topic terms & Count case-insensitive matches of these speech/audio terms: speech, audio, acoustic, spectrogram, ASR, TTS, voice, phoneme, waveform. Divide by max(total word count, 1) and multiply by 1,000. & +0.01 & $\,$0.00 \\
Table refs. per pg. & Count case-sensitive matches of Table followed by whitespace and a digit, then divide by the estimated body-page proxy. & $-$0.03 & +0.06 \\
Theory markers & Count case-sensitive matches of Theorem, Lemma, Proposition, Corollary, Definition, and Proof. & +0.03 & +0.06 \\
Theory topic terms & Count case-insensitive matches of these theory/proof/bound terms: theorem, proof, lemma, proposition, corollary, bound, complexity, PAC, VC dimension, Rademacher, concentration inequality. Divide by max(total word count, 1) and multiply by 1,000. & +0.07 & +0.01 \\
Title colon & Set to 1 if the extracted title contains a colon, otherwise 0. & +0.05 & +0.05 \\
Title length & Use metadata title when available, otherwise the first Markdown H1 heading, and count characters. & $-$0.07 & +0.01 \\
Title number & Set to 1 if the extracted title contains any digit, otherwise 0. & +0.06 & $\,$0.00 \\
Title question mark & Set to 1 if the extracted title contains a question mark, otherwise 0. & +0.01 & +0.01 \\
Title words & Use metadata title when available, otherwise the first Markdown H1 heading, and count whitespace-delimited title words. & $-$0.03 & $-$0.09 \\
Training details & Count how many of these phrases appear case-insensitively in the text: learning rate, batch size, epoch, GPU, TPU, A100, V100, training time. & $-$0.05 & $-$0.07 \\
Vocabulary richness & From the first 5,000 lowercased alphabetic words, compute number of unique words divided by max(number of sampled words, 1). & +0.21 & $\,$0.00 \\
Word length & Strip punctuation from whitespace tokens, keep tokens containing only alphabetic characters, and average their character lengths, using 0 when no alphabetic words are found. & $-$0.07 & $-$0.03 \\
\bottomrule
\end{longtable}

The main text reports the top five accept- and reject-associated features from separate ICLR and \datasetarxivname{} feature probes in \Cref{fig:feature_probe_joint_audit}. \Cref{tab:feature_inventory} lists every feature in the probe with its definition. The columns $w_{\mathrm{ICLR}}$ and $w_{\mathrm{arXiv}}$ are the signed standardized coefficients from logistic-regression probes trained on all features against the ICLR and \datasetarxivname{} decision label, respectively. The probe excludes source identifiers, direct venue labels, ratings, citations, author names, and non-comparable body-page counts. The strongest individual signal --- paper length on both sources --- saturates at $w \approx +0.48$, an order of magnitude below what \systemfamily{} achieves end-to-end. Reject-side magnitudes are smaller still ($|w| \le 0.25$), and most features sit between $-0.05$ and $+0.10$, indicating that simple structural shortcuts do not cleanly separate accept from reject.

    \section{Do Frontier Models Memorize the Test Decisions?}\label{appendix:memory_probe}

    GPT-5.5's December '25 knowledge cutoff postdates the public release of ICLR '25 decisions and most of our \datasetarxivname{} test years, so a natural worry is that its baseline accuracy comes from remembering outcomes rather than judging papers. Gemini-3.1-Pro (January '25 cutoff) predates both. We test this in two steps. First, we ask whether the models remember our test papers at all. Second, and more importantly, we ask whether remembering a paper makes the model any more accurate on it.

    \paragraph{Do the models remember the papers?}
    We sample 200 papers per corpus (100 accepted, 100 rejected) from the papers on which GPT-5.5 is evaluated in \Cref{tab:main_results}, so that every probed paper also has a prediction from the main table. Each paper is probed with five \emph{label-recall} cues that ask the model whether it remembers the paper's decision:
    \begin{itemize}[noitemsep,topsep=2pt,leftmargin=*]
        \item \texttt{title} --- title only.
        \item \texttt{title\_abstract} --- title plus abstract.
        \item \texttt{mc} --- multiple choice among accept, reject, and a distractor, given the title and abstract.
        \item \texttt{continuation} --- the first sentence after the abstract, asking whether the model recognizes the paper.
        \item \texttt{web\_search\_framing} --- a prompt phrased as if the model could look up the venue and outcome.
    \end{itemize}
    For each cue we report the \emph{claim rate}, the fraction of papers where the model says it remembers the decision, and the \emph{outcome accuracy when claimed}. A sixth \emph{content-recall} cue asks the model to continue the abstract from its title, scored by ROUGE-L \citep{lin-2004-rouge}. Gemini-3.1-Pro was probed on an earlier 50-paper draw per corpus with the same cues.

    \Cref{tab:memory_probe_cues} shows that GPT-5.5 does recognize many of our test papers, and that recognition follows its cutoff exactly. On \dataseticlrshort{}, the \texttt{title} cue is claimed for 30\% of '25 papers but only 7\% of '26 papers, and on \datasetarxivname{} for 44\% of '24 papers but 17\% of '25 and 23\% of '26. Gemini-3.1-Pro claims recall for at most 4\% of \dataseticlrshort{} papers under any cue. Recognition is not the same as remembering the decision, however. When GPT-5.5 claims to recall a decision it is right only 50--73\% of the time, and this is inflated by an accept bias: it says ``accept'' for 65\% of the \dataseticlrshort{} papers it claims to remember, so balanced accuracy over its claimed decisions is 44\%, no better than chance. Content recall is absent: mean ROUGE-L on abstract continuation is $\le 0.004$ on every slice and no paper reaches $0.5$.

    \paragraph{Does remembering a paper help?}
    If GPT-5.5 were retrieving remembered decisions, it would be more accurate on the papers it recognizes. We split each probed corpus into papers it claimed to recall under at least one cue and papers it never claimed, and compare its accuracy from \Cref{tab:main_results} on the two groups (\Cref{tab:memory_probe_tuned}). Recognized papers are no easier on either dataset (permutation $p = .28$ and $.88$), including on ICLR '25, the only \dataseticlrshort{} year inside its cutoff, and accuracy does not rise with the number of cues under which a paper is recognized. Papers whose recalled decision happened to be correct do score higher, but this is circular, since those are the papers where the model's impression already matched the outcome.

    We conclude that GPT-5.5 recognizes a substantial fraction of our test papers but does not retrieve their decisions, and that its baseline accuracy in \Cref{tab:main_results} is not explained by memorization.

    \begin{table}[H]
        \centering
        \footnotesize
        \setlength{\tabcolsep}{4pt}
        \renewcommand{\arraystretch}{1.08}
        \begin{tabular}{@{}llcccccc@{}}
            \toprule
            & & \multicolumn{2}{c}{\textbf{\dataseticlrshort{}}} & \multicolumn{3}{c}{\textbf{\datasetarxivname{}}} \\
            \cmidrule(lr){3-4}\cmidrule(lr){5-7}
            \textbf{Cue} & \textbf{Model} & '25 & '26 & '24 & '25 & '26 \\
            \midrule
            \texttt{title} & GPT-5.5 & 30 / 79 & 7 / 50 & 44 / 82 & 17 / 33 & 23 / 43 \\
            \texttt{title} & Gemini-3.1-Pro & 0 / 0 & 0 / 0 & 5 / 100 & 0 / 0 & 0 / 0 \\
            \texttt{title\_abstract} & GPT-5.5 & 22 / 65 & 4 / 75 & 46 / 83 & 13 / 50 & 10 / 33 \\
            \texttt{title\_abstract} & Gemini-3.1-Pro & 0 / 0 & 0 / 0 & 15 / 67 & 0 / 0 & 0 / 0 \\
            \texttt{mc} & GPT-5.5 & 33 / 65 & 0 / 0 & 73 / 78 & 16 / 59 & 0 / 0 \\
            \texttt{mc} & Gemini-3.1-Pro & 0 / 0 & 0 / 0 & 30 / 83 & 0 / 0 & 0 / 0 \\
            \texttt{continuation} & GPT-5.5 & 52 / 54 & 25 / 48 & 60 / 66 & 39 / 48 & 43 / 31 \\
            \texttt{continuation} & Gemini-3.1-Pro & 0 / 0 & 0 / 0 & 5 / 100 & 0 / 0 & 0 / 0 \\
            \texttt{web\_search\_framing} & GPT-5.5 & 44 / 51 & 25 / 48 & 52 / 70 & 42 / 53 & 13 / 50 \\
            \texttt{web\_search\_framing} & Gemini-3.1-Pro & 4 / 100 & 0 / 0 & 10 / 100 & 0 / 0 & 0 / 0 \\
            \midrule
            \multicolumn{7}{l}{\textit{Content recall (abstract completion) --- ROUGE-L mean / frac.\ ROUGE-L $\ge 0.5$}} \\
            \multicolumn{2}{l}{GPT-5.5} & 0.003 / 0.00 & 0.003 / 0.00 & 0.004 / 0.00 & 0.002 / 0.00 & 0.000 / 0.00 \\
            \multicolumn{2}{l}{Gemini-3.1-Pro} & 0.000 / 0.00 & 0.000 / 0.00 & 0.005 / 0.00 & 0.000 / 0.00 & 0.000 / 0.00 \\
            \bottomrule
        \end{tabular}
        \caption{\textbf{Frontier memorization probes, by corpus and year.} Label-recall cells report \emph{claim rate (\%) / outcome accuracy when claimed (\%)}; abstract-completion cells report \emph{ROUGE-L mean / fraction with ROUGE-L $\ge 0.5$}. GPT-5.5 is probed on 200 papers per corpus (\dataseticlrshort{} '25: $n{=}93$, '26: $n{=}107$; \datasetarxivname{} '24: $n{=}63$, '25: $n{=}107$, '26: $n{=}30$), Gemini-3.1-Pro on 50 (\dataseticlrshort{} '25: $n{=}26$, '26: $n{=}24$; \datasetarxivname{} '24: $n{=}20$, '25: $n{=}23$, '26: $n{=}7$). GPT-5.5's claim rate tracks its December '25 cutoff, but its claimed decisions are right only about half to three quarters of the time.}
        \label{tab:memory_probe_cues}
    \end{table}

    \begin{table}[H]
        \centering
        \footnotesize
        \setlength{\tabcolsep}{5pt}
        \renewcommand{\arraystretch}{1.08}
        \begin{tabular}{@{}llcccccc@{}}
            \toprule
            & & \multicolumn{2}{c}{\textbf{Recalled}} & \multicolumn{2}{c}{\textbf{Not recalled}} & \multicolumn{2}{c}{\textbf{Full test pool}} \\
            \cmidrule(lr){3-4}\cmidrule(lr){5-6}\cmidrule(lr){7-8}
            \textbf{Corpus} & \textbf{Year} & $n$ & Acc & $n$ & Acc & $n$ & Acc \\
            \midrule
            \dataseticlrshort{} & '25 & 74 & 57 & 18 & 70 & 339 & 61 \\
            \dataseticlrshort{} & '26 & 47 & 62 & 60 & 64 & 495 & 65 \\
            \dataseticlrshort{} & all & 121 & 58 & 78 & 65 & 834 & 63 \\
            \midrule
            \datasetarxivname{} & '24 & 57 & 64 & 6 & 50 & 261 & 72 \\
            \datasetarxivname{} & '25 & 68 & 68 & 38 & 61 & 368 & 66 \\
            \datasetarxivname{} & '26 & 17 & 64 & 13 & 76 & 94 & 64 \\
            \datasetarxivname{} & all & 142 & 67 & 57 & 66 & 723 & 68 \\
            \bottomrule
        \end{tabular}
        \caption{\textbf{GPT-5.5 accuracy on papers it recognizes vs.\ papers it does not.} Accuracy (\%) of GPT-5.5 (vision, tuned as in \Cref{tab:main_results}) on probed papers it claimed to recall under at least one label-recall cue, on probed papers it never claimed, and on the full test pool. Recognized papers are no easier for the model, including on ICLR '25, the only \dataseticlrshort{} year inside its cutoff.}
        \label{tab:memory_probe_tuned}
    \end{table}

\clearpage
\section{Backbone Choice: Qwen2.5 vs.\ Qwen3}\label{appendix:backbone_choice}

    To check whether our backbone choice drives the results, we retrain the text model on \dataseticlrshort{} with a Qwen3-8B backbone \citep{qwen3} using the same SFT recipe as \systemtext{}, and compare against the Qwen2.5-7B backbone used throughout the paper. \Cref{tab:backbone_choice} reports overall \dataseticlrshort{} accuracy and per-class recall together with each model's knowledge cutoff. The newer Qwen3 generation includes public ICLR '25 outcomes (text released April 28, '25; VL on September 22, '25) but not ICLR '26, whereas Qwen2.5 predates both.

    \begin{table}[H]
        \centering
        \footnotesize
        \setlength{\tabcolsep}{6pt}
        \renewcommand{\arraystretch}{1.08}
        \begin{tabular}{@{}llccc@{}}
            \toprule
            Text backbone & Knowledge cutoff & Acc & $R_A$ & $R_R$ \\
            \midrule
            Qwen2.5-7B (ours) & December '24 & 66 & 63 & 69 \\
            Qwen3-8B          & April '25    & 65 & 72 & 59 \\
            \bottomrule
        \end{tabular}
        \caption{\textbf{Text backbone comparison on \dataseticlrshort{}.} Overall accuracy (Acc), accept recall ($R_A$), and reject recall ($R_R$) in percent. Qwen3's cutoff includes ICLR '25 decisions but not '26.}
        \label{tab:backbone_choice}
    \end{table}

    \noindent Accuracy is close ($65\%$ vs.\ $66\%$), and Qwen3 skews toward accept. The backbone matters less than the training data, and Qwen2.5's cutoff predates every test decision, so we use it throughout.

\section{Training Variations}\label{app:training_variations}

\subsection{Evaluation and Training Prior Investigation}\label{appendix:prior-diagnostics}

    The main text reports balanced test metrics because our decision objective is Acc. This is compatible with natural-rate reporting when balanced and natural samples come from the same curated domain distribution. The accept and reject recalls estimated on a balanced test can be reweighted by the natural accept prior. \Cref{tab:natural_prior_extrapolation} checks this directly for matched \dataseticlrshort{} and \datasetarxivname{} 7B text and vision predictors. Across all four cells, the empirical natural-rate accuracy lands within a few percentage points of the prior-extrapolated value, so the balanced test keeps the metric aligned with the objective.

    \begin{table}[H]
    \centering
    \footnotesize
    \setlength{\tabcolsep}{4pt}
    \renewcommand{\arraystretch}{1.1}
    \begin{tabular}{@{}llrrrr@{}}
        \toprule
        \textbf{Domain} & \textbf{Predictor} & \textbf{Balanced Acc} & \textbf{Extrapolated natural} & \textbf{Empirical natural} & $\boldsymbol{\Delta}$ \\
        \midrule
        ICLR & Vision & 68.3 & 68.0 & 65.6 & -2.4 \\
        ICLR & Text & 66.0 & 67.2 & 64.1 & -3.1 \\
        arXiv & Vision & 73.4 & 74.2 & 75.1 & +1.0 \\
        arXiv & Text & 71.3 & 71.7 & 72.0 & +0.3 \\
        \bottomrule
    \end{tabular}
    \caption{\textbf{Balanced-to-natural extrapolation.} For matched ICLR$\rightarrow$ICLR and arXiv$\rightarrow$arXiv diagnostics, balanced-test accept and reject recalls are converted to raw accuracy under the natural test prior. $\Delta$ is empirical natural-prior accuracy minus this extrapolated value.}
    \label{tab:natural_prior_extrapolation}
    \vspace{-0.55em}
\end{table}

    The training-prior diagnostic in \Cref{tab:training_prior_discrepancy} uses validation thresholds only to reveal class-prior bias. All headline \systemfamily{} results use native decisions unless stated otherwise (e.g., calibration).

\subsection{Ablating Training Distribution and Paper Representation}\label{appendix:data_mixture}

    We explored two families of training ablations on \dataseticlrshort{} to improve decision accuracy. The first filtered training papers to high-quality accepts and rejects by reviewer rating, dropping borderline papers. The second tested additional context suffixes such as paper statistics, agent-generated reviews, and vision-text fusion to support the decision objective. Both are summarized in \Cref{tab:data_mixture}.

    Table~\ref{tab:data_mixture}a shows that training only on pre-'26 data does not transfer well to '26. To improve the quality of the corpus, we tried to filter out borderline cases via \emph{Rating-Agreeing}, which keeps only accepts in $[0.6,1]$ and rejects in $[0,0.4]$ by percentile rating. It helps on '25, but once '26 is included, it does not improve either modality noticeably. Table~\ref{tab:data_mixture}b explains why: \textbf{'26 is fundamentally a different distribution}, since the rebuttal process was cut short, author-reviewer discussion was not allowed, and reviewers could not raise scores, yielding the smallest accept-reject gap. Table~\ref{tab:data_mixture}c shows that additional context helps text, especially for '26 transfer, but harms vision, our strongest model. We therefore train on the simplest paper-only pool, \textbf{'20--'23, '25--'26$^\ast$}, which gives the strongest vision results without sacrificing text performance.

    \begin{table}[H]
        \centering
        \footnotesize
        \setlength{\tabcolsep}{2.2pt}
        \renewcommand{\arraystretch}{1.05}
        \begin{minipage}[c]{0.37\textwidth}
            \centering
            \begin{tabular}{@{}lcccc@{}}
                \toprule
                \multirow{2}{*}{\textbf{Training Data}} & \multicolumn{2}{c}{\textbf{'25}} & \multicolumn{2}{c}{\textbf{'26}} \\
                \cmidrule(lr){2-3} \cmidrule(lr){4-5}
                & \textbf{Vision} & \textbf{Text} & \textbf{Vision} & \textbf{Text} \\
                \midrule
                \underline{'20--'23, '25} & 68.1 & 68.3 & 62.6 & 63.5 \\
                + Rating-Agreeing & {\color{gaingreen}+1.6} & {\color{gaingreen}+0.3} & {\color{lossred}-0.9} & {\color{lossred}-1.7} \\
                \midrule
                \textbf{\underline{'20--'23, '25--'26$^\ast$}} & 70.9 & 69.4 & 66.5 & 63.7 \\
                + Rating-Agreeing & {\color{lossred}-1.0} & {\color{lossred}-2.2} & {\color{lossred}-1.2} & {\color{gaingreen}+0.9} \\
                \bottomrule
            \end{tabular}

            \vspace{1.5ex}
            \textbf{(a) Training Data}
        \end{minipage}
        \hfill
        \begin{minipage}[c]{0.17\textwidth}
            \centering
            \begin{tabular}{@{}lccc@{}}
                \toprule
                \textbf{Year} & \textbf{Acc.} & \textbf{Rej.} & \textbf{Gap} \\
                \midrule
                '20 & 6.25 & 3.67 & 2.58 \\
                '21 & 6.53 & 5.03 & 1.50 \\
                '22 & 6.62 & 4.67 & 1.95 \\
                '23 & 6.60 & 4.67 & 1.93 \\
                '25 & 6.47 & 4.80 & 1.67 \\
                \textbf{'26} & \textbf{5.40} & \textbf{3.96} & \textbf{1.44} \\
                \bottomrule
            \end{tabular}

            \vspace{1.5ex}
            \textbf{(b) Rating Gap}
        \end{minipage}
        \hfill
        \begin{minipage}[c]{0.37\textwidth}
            \centering
            \begin{tabular}{@{}lcccc@{}}
                \toprule
                \multirow{2}{*}{\textbf{Additional Context}} & \multicolumn{2}{c}{\textbf{'25}} & \multicolumn{2}{c}{\textbf{'26}} \\
                \cmidrule(lr){2-3} \cmidrule(lr){4-5}
                & \textbf{Vision} & \textbf{Text} & \textbf{Vision} & \textbf{Text} \\
                \midrule
                \textbf{\underline{'20--'23, '25--'26$^\ast$}} & 70.9 & 69.4 & 66.5 & 63.7 \\
                + Simple Features & {\color{lossred}-0.4} & {\color{gaingreen}+0.6} & {\color{lossred}-1.5} & {\color{gaingreen}+0.7} \\
                + Qwen Reviews & {\color{lossred}-1.3} & {\color{lossred}-2.1} & {\color{lossred}-0.4} & {\color{gaingreen}+1.0} \\
                + Gemini Reviews & {\color{lossred}-3.8} & {\color{lossred}-1.1} & {\color{lossred}-2.0} & {\color{gaingreen}+0.7} \\
                \bottomrule
            \end{tabular}

            \vspace{1.5ex}
            \textbf{(c) Additional Context}
        \end{minipage}
        \caption{\textbf{Data mixture breakdown.} We report fixed-cost data and context ablations, with underlined rows as bases and following rows as deltas. Rating-agreeing data helps '25 but not the final '26-oriented setup; '26 has the smallest accept-reject rating gap; and extra feature/review context helps text modestly but not the strongest vision model.}
        \label{tab:data_mixture}
    \end{table}

    We also tested several fixed-compute design choices beyond the paper-only input. As Table~\ref{tab:data_mixture}c shows, suffixing text with paper statistics or agent-generated Qwen/Gemini reviews can help text modestly, especially on '26, but none of these additions improves the strongest vision model. The same is true for other fixed-cost variants such as suffixing text with full vision, vision with full text, or prepending 1k \texttt{<think>} tokens. None of these design choices consistently improves performance, so we simply use paper~$\to$~decision prediction.

    \textbf{Paper statistics features.}
    We appended structured metadata such as page count, number of references, inline math count, cross-reference density, and other surface-level features to each training sample. While these features are individually predictive, adding them to the SFT prompt did not improve over the paper-only baseline, suggesting that the language model already captures these signals implicitly from the paper text.

    \textbf{Thinking tokens.}
    Inspired by chain-of-thought prompting, we prepended dummy reasoning tokens before the decision token during SFT, allowing the model additional computation before committing to a prediction. This did not improve accuracy, consistent with our finding that the decision task benefits more from pattern recognition over the full paper than from explicit multi-step reasoning.

    \textbf{Multimodal fusion.}
    We tested providing both the full extracted text and screenshots of all pages simultaneously (text + vision), as well as the reverse (vision + text). Neither configuration outperformed the single-modality baselines, likely because the two representations are largely redundant for the decision task, and the additional context length increases noise without adding a complementary signal.

    \textbf{Synthetic review augmentation.}
    Appending model-generated reviews (Qwen3.5-122B-A10B or Gemini-2.5-Flash) to training prompts added ${\sim}1.8$k--$4.8$k tokens per sample. These reviews occasionally helped on ICLR~'25 but hurt on ICLR~'26, suggesting that review language is more year-specific than paper content and does not transfer reliably.

\subsection{Reverse-Order Ablation}\label{appendix:reverse_order}

    To test whether SFT models rely on positional shortcuts, we reversed the paper order while keeping labels fixed --- text examples reverse section blocks, and vision examples reverse page order. Training and evaluating in the same reversed condition only loses about 1--1.3\% relative to the forward baseline, and forward-trained checkpoints remain usable on reversed validation inputs (Table~\ref{tab:reverse_order}). This suggests the models are not merely keying on early-section or early-page cues; vision is especially order-invariant.

    \begin{table}[H]
        \centering
        \footnotesize
        \setlength{\tabcolsep}{4pt}
        \renewcommand{\arraystretch}{1.08}
        \begin{tabular}{@{}llcc@{}}
            \toprule
            \textbf{Training} & \textbf{Evaluation} & \textbf{Accuracy} & \textbf{n} \\
            \midrule
            \multicolumn{4}{@{}l}{\textit{Matched training and evaluation order}} \\
            Forward (vision) & Forward   & 68.4\% & 2{,}498 \\
            Reversed (vision)& Reversed  & 67.2\% & 2{,}498 \\
            Forward (text)   & Forward   & 67.0\% & 2{,}495 \\
            Reversed (text)  & Reversed  & 65.7\% & 2{,}495 \\
            \midrule
            \multicolumn{4}{@{}l}{\textit{Forward-trained checkpoints, reversed-order evaluation}} \\
            Forward (vision) & Reversed  & 66.1\% & 1{,}253 \\
            Forward (text)   & Reversed  & 64.9\% & 1{,}252 \\
            \bottomrule
        \end{tabular}
        \caption{\textbf{Reverse-order ablation.} Reversing text section blocks or vision page order during training and evaluation costs only $1$--$1.3$ points of accuracy. Forward-trained checkpoints also stay close to their forward accuracy when evaluated on reversed inputs, indicating that vision in particular is largely order-invariant.}
        \label{tab:reverse_order}
    \end{table}

\section{Cross-Source Generalization}\label{app:generalization}
    How does a model trained on one source behave on the other? We evaluate the \dataseticlrshort{}-trained
    and \datasetarxivname{}-trained 7B vision models on each source's test set and on \datasetarxiviclrname{},
    the \dataseticlrshort{} subset of \datasetarxivname{} (Table~\ref{tab:generalization}).

    \begin{table}[h!]
        \centering
        \small
        \setlength{\tabcolsep}{5pt}
        \begin{tabular}{@{}l cccc@{}}
\toprule
\textbf{Train $\to$ test (7B vision)} & $R_A$ & $R_R$ & Acc & $n$ \\
\midrule
\multicolumn{5}{@{}l}{\emph{Generalizes: accept and reject recall stay balanced}} \\
\quad \textit{ICLR} $\to$ \dataseticlrname{} & 69.2 & 67.4 & 68.3 & 1670 \\
\quad \textit{ICLR} $\to$ \datasetarxivname{}-ICLR & 78.3 & 69.5 & 73.9 & 105 \\
\addlinespace[2pt]
\multicolumn{5}{@{}l}{\emph{Reject-predictive off-domain: $R_R$ holds, $R_A$ collapses}} \\
\quad \datasetarxivname{} $\to$ \dataseticlrname{} & 52.5 & 68.9 & 60.7 & 1670 \\
\quad \textit{ICLR} $\to$ \datasetarxivname{} (all) & 50.1 & 78.2 & 64.2 & 1414 \\
\bottomrule
\end{tabular}

        \caption{\textbf{Cross-source generalization (7B vision).} Accept recall ($R_A$), reject recall
        ($R_R$), and accuracy (Acc), in \%, over $n$ test papers. \dataseticlrshort{} training reaches
        \datasetarxiviclrname{} without loss. But faced with the other source's full set, a model
        still flags its rejects and recovers only about half of its accepts.}
        \label{tab:generalization}
    \end{table}

    The \dataseticlrshort{}-trained model transfers to \datasetarxiviclrname{} with no
    measurable loss (Table~\ref{tab:generalization}). With only $105$ papers, though, its accuracy there
    has a $95\%$ confidence interval that overlaps its in-domain accuracy, so we read this as equal
    performance, not a real gain. A much larger effect appears when either model is run on the full
    distribution of the other source. Both become reject-predictive, recovering only about half of the
    true accepts, so each defaults to rejection on papers that look unfamiliar.

    This is not a threshold problem. We retuned the accept/reject threshold on a small labeled
    sample from the target source, and it only swapped the error. Accepts then rose above $88\%$ while
    rejects fell below $40\%$, and accuracy stayed flat. Even the best threshold chosen directly
    on the test labels barely changed it. The model's ranking of papers, not the threshold, is what
    fails under the shift.

\section{Prompting and Extended Experiments}\label{app:prompting}

    \Cref{tab:prompting_configs} summarizes the prompts, decoding settings, and context limits used across our evaluation suite. Frontier Gemini/GPT models use a single venue-generic rating-plus-decision prompt with \emph{high reasoning} effort and temperature $1.0$. Our RL prompt mirrors this setup with temperature $0.7$. After SFT we decode greedily (temperature~$0$). Open-source Qwen baselines use the same prompt family as SFT/RL but at higher temperature to explore prompt corrections, since the base models predict acceptance for almost every paper out of the box. \Cref{appendix:prompting} documents each prompt verbatim, \Cref{appendix:frontier_reasoning} reports reasoning-token usage across frontier evaluation cells, and \Cref{appendix:prompt_ablations} reports the prompt-correction experiments on Qwen and Gemini-2.5-Flash.

    \begin{table}[H]
        \centering
        \footnotesize
        \begin{tabular}{@{}>{\raggedright\arraybackslash}p{0.22\textwidth}>{\raggedright\arraybackslash}p{0.34\textwidth}>{\raggedright\arraybackslash}p{0.36\textwidth}@{}}
            \toprule
            \textbf{Family} & \textbf{Prompt/input protocol} & \textbf{Decoding and limits} \\
            \midrule
            Frontier Models (Gemini/GPT) & Single venue-generic rating-plus-decision prompt. & Gemini-3-Flash, GPT-5.4-Mini, Gemini-3.1-Pro, and GPT-5.5; high reasoning/thinking effort, temperature $1.0$, max tokens $32{,}000$. \\
            Coding agents & Same rating-plus-decision prompt in single-paper workspaces with markdown and screenshots. & Codex (GPT-5.5) and Claude Code (Sonnet-4.6) with high reasoning effort; web tools disabled. \\
            Qwen Base Prompting & Answer-only prompt matching SFT, or reasoning prompt matching RL. Target-dataset venue wording. & temperature $0.95$, top-$p=0.7$, top-$k=50$, context $24{,}480$. Few-shot runs use context $60{,}000$ with 4x YaRN. \\
            SFT & Venue-specific answer-only prompts. & temperature $0$ (greedy decoding), context $24{,}480$. \\
            GRPO / RL & Venue-specific reasoning prompt. & temperature $0.7$. Text max response $4096$, max prompt $25{,}480$. Vision max response $2046$, max prompt $24{,}480$. \\
            \bottomrule
        \end{tabular}
        \caption{\textbf{Prompting and decoding configurations.} Prompts, decoding settings, and context limits across external reviewers, direct Qwen prompting, SFT, and GRPO.}
        \label{tab:prompting_configs}
    \end{table}

    \subsection{Prompts}\label{appendix:prompting}
    In this section we denote the prompts and decoding settings used during prompting, SFT, and RL. We omit the paper body from the prompts for brevity.

    \subsubsection{Base Open-Source Prompting}

    \textbf{No Reasoning Prompt.}
    Open-source Qwen and SFT runs share a minimal answer-only prompt. The ICLR variant tells the model that ICLR accepts around 30\%, while the \datasetarxivname{} variant uses venue-generic wording because the source-derived dataset spans multiple ML venues. For cross-source evaluation, the wording follows the target dataset rather than the training source.
    \begin{Verbatim}[fontsize=\scriptsize,frame=single]
[System]
You are an expert academic reviewer tasked with evaluating research papers.

[User --- ICLR]
I am giving you a paper. I want to predict its acceptance outcome at ICLR.
 - Your answer will either be: \boxed{Accept} or \boxed{Reject}
 - Note: ICLR generally has a ~30% acceptance rate

[User --- arXiv]
I am giving you a paper submitted to a top machine-learning venue. Predict
its acceptance outcome.
 - Your answer will either be: \boxed{Accept} or \boxed{Reject}
 - Note: typical top-tier ML venues have ~25-30% acceptance rates
    \end{Verbatim}

    \textbf{Reasoning Prompt.}
    Reasoning prompts move the instruction into the system turn and leave the user turn as the paper body. The full \dataseticlrshort{} variant is shown below. The \datasetarxivname{} variant replaces the first sentence with the venue-generic top-ML wording. Vision reasoning prompts add that the paper is provided as Markdown text plus screenshots of each page, and ask the model to inspect page layouts.
    \begin{Verbatim}[fontsize=\scriptsize,frame=single]
[System]
You are an expert academic reviewer for the ICLR conference, predicting whether a
paper will be accepted or rejected. ICLR generally has a ~30% acceptance rate.

Your response must include two parts:

1. **Reasoning**: A detailed, free-flowing chain of thought enclosed in
`<think>` and `</think>` tags.
2. **Final Answer**: A single boxed verdict enclosed in `<answer>` and
`</answer>` tags.

* The reasoning should resemble a stream of consciousness: explore, test
  hypotheses, backtrack if necessary, reflect, and refine.
* Use reasoning strategies such as planning, exploration, evaluation, and
  reflection.

* The answer section must be enclosed in `<answer>` and `</answer>` tags.
* The answer is exactly `\boxed{Accept}` or `\boxed{Reject}`, with no other text.

[User]
<paper body as Markdown; vision variant also attaches per-page screenshots>
    \end{Verbatim}

    \subsubsection{Frontier Model Prompting}

    \textbf{Frontier Reasoning Prompt.}
    Frontier Gemini and GPT models (Gemini-3.1-Pro and GPT-5.5) use a single venue-generic rating-plus-decision prompt for both \dataseticlrshort{} and \datasetarxivname{}. We chose this prompt after finding no consistent gain from separate \dataseticlrshort{}- and \datasetarxivname{}-specific variants. In practice, the venue is usually recoverable from the paper content and formatting.
    \begin{Verbatim}[fontsize=\scriptsize,frame=single]
[System]
You are an expert academic ML researcher and reviewer tasked with evaluating
research papers. I am giving you a paper submitted to a top machine-learning venue.
Predict a rating from 1-10 and the paper's acceptance outcome.
 - Your answer will be of the form: \boxed{rating,decision} where rating is an
   integer in [1,10] and decision is either {Accept, Reject}
 - Note: typical top-tier ML venues have ~25-30% acceptance rates

[User]
<paper body as Markdown; vision variant also attaches per-page screenshots>
    \end{Verbatim}
    We use the generated rating as the score, with a higher rating indicating a higher predicted probability of acceptance, and the generated decision as the hard label. In our main results we additionally calibrate the model's decision using its rating and accuracy performance on the held-out validation set, which generally improves performance.

    \textbf{Frontier Reasoning Agent Workspace.}
    Coding-agent baselines use the same Frontier Reasoning Prompt above (carried in \texttt{AGENTS.override.md}), but place each paper in a self-contained workspace that the agent can inspect with normal file and image tools. The paper body is stripped into \texttt{paper.md}, rendered pages are copied into \texttt{images/}, and the final answer must be written to \texttt{PREDICTIONS.json}. Both Codex (GPT-5.5) and Claude Code (Sonnet-4.6) run with high reasoning effort, no web tools, and a validator that rejects malformed outputs before grading.
    \begin{Verbatim}[fontsize=\scriptsize,frame=single]
agent_workspaces/<sample>/<domain>/<paper_id>/
|-- AGENTS.override.md        # rating + accept/reject instruction
|-- CLAUDE.md -> AGENTS.override.md
|-- id.txt                    # paper identifier
|-- paper.md                  # stripped markdown paper body
|-- images/
|   |-- page_1.png
|   |-- page_2.png
|   `-- ...
|-- main.py                   # calls the schema validator
`-- results/
    |-- codex/
    |   |-- PREDICTIONS.json
    |   `-- session.jsonl
    `-- claude/
        |-- PREDICTIONS.json
        `-- session.jsonl
    \end{Verbatim}
    Grading joins each valid \texttt{PREDICTIONS.json} with the held-out ground-truth manifest and reports accuracy, accept/reject recall, and rating correlation. The workspace tree is intentionally one paper at a time, matching the single-shot API setup while letting the agent choose whether to read markdown, rendered pages, or both modalities together.

    \subsection{Frontier Model Reasoning Usage}\label{appendix:frontier_reasoning}

    \Cref{tab:frontier_reasoning_tokens} reports reasoning-token usage across the four \dataseticlrshort{}/\datasetarxivname{} text/vision evaluation cells. Both frontier models spend a substantial amount of reasoning compute per paper. On text inputs, median reasoning tokens land between $1.3$k and $2.3$k per paper, and on vision inputs between $1.5$k and $2.1$k. P95 reasoning length stretches to $3.6$k--$4.7$k, so a non-trivial tail of papers triggers $2\times$ or more compute than the median. GPT-5.5 has a lower median than Gemini-3.1-Pro at every cell, but its tail is just as heavy: its P95 sits at $3.6$k--$4.5$k against Gemini-3.1-Pro's $4.1$k--$4.7$k, and its P95-to-median ratio is the larger of the two, so its compute varies more from paper to paper. \dataseticlrshort{} cells consistently use more reasoning than the matched \datasetarxivname{} cells (e.g.\ Gemini-3.1-Pro at $2{,}085$ vision-median tokens on \dataseticlrshort{} vs $1{,}825$ on \datasetarxivname{}), suggesting that \dataseticlrshort{} papers either look harder or hit prompts that elicit more deliberation.

    \begin{table}[H]
        \centering
        \footnotesize
        \begin{tabular}{@{}llrrrr@{}}
            \toprule
            \textbf{Model} & \textbf{Cell} & \textbf{$n$} & \textbf{Median} & \textbf{Mean} & \textbf{P95} \\
            \midrule
            Gemini-3.1-Pro & \datasetarxivname{} vision & 360 & 1825 & 2110 & 4121 \\
            Gemini-3.1-Pro & \datasetarxivname{} text & 354 & 1944 & 2289 & 4594 \\
            Gemini-3.1-Pro & ICLR vision & 401 & 2085 & 2353 & 4618 \\
            Gemini-3.1-Pro & ICLR text & 396 & 2311 & 2611 & 4702 \\
            GPT-5.5 & \datasetarxivname{} vision & 360 & 1536 & 1887 & 4095 \\
            GPT-5.5 & \datasetarxivname{} text & 362 & 1336 & 1576 & 3580 \\
            GPT-5.5 & ICLR vision & 416 & 2044 & 2109 & 4482 \\
            GPT-5.5 & ICLR text & 418 & 1536 & 1929 & 4383 \\
            \bottomrule
        \end{tabular}
        \caption{\textbf{Frontier reasoning-token usage.} Median, mean, and P95 reasoning-token counts for the stratified high-reasoning Gemini and GPT evaluation runs.}
        \label{tab:frontier_reasoning_tokens}
    \end{table}

    \subsection{Correcting Model Biases with Prompting}\label{appendix:prompt_ablations}

    Base open-source models we tested predict acceptance for almost every paper. Out of the box, Qwen2.5-7B with a minimal freeform-boxed prompt accepts more than $99\%$ of papers (\Cref{tab:prompt_ablations_qwen_text}, Freeform Boxed Standard: $R_A = 99.4$, $R_R = 0.9$). We therefore tested a range of prompt corrections, few-shot conditioning, and multi-step peer-review emulation across two open-source models (Qwen2.5-7B text and vision, all $2{,}495$ \dataseticlrshort{} test papers) and a cheap frontier model (Gemini-2.5-Flash text and vision, a $500$-paper stratified subset). All numbers are averages across '25 and '26.

    \textbf{Prompt variants.} We compare four prompt formats. \emph{Freeform Boxed} requests a boxed Accept/Reject answer with no structure. \emph{Structured JSON} requests a full review schema (summary, strengths, weaknesses, score, decision). \emph{Fewshot 1A-1R} prepends one accept and one reject calibration example. \emph{Fewshot 0A-2R} uses two reject examples. Each format can be paired with a \emph{Critical} modifier (an explicit instruction to be wary of overstated claims and to prefer rejection when uncertain). Multi-turn variants include \emph{Majority voting} over 5 generations, \emph{Meta-Review} (the Sequential Refinement of \citet{madaan2025rethinking}), and \emph{PDR Boxed} (a single-generation Parallel-Distill-Reflect prompt that runs an Advocate, Critic, and Calibrator inside one rollout).

    \textbf{Critical variant.}
    Appending a single ``be critical, prefer rejection when uncertain'' instruction is not enough to stop the model from predicting acceptance for nearly every paper. On Qwen2.5-7B text, Freeform Boxed Critical only nudges reject recall from $0.9\%$ to $4.3\%$. Switching to Structured JSON (a full review schema with strengths, weaknesses, score, and decision fields) with the Critical modifier produces a larger swing ($R_R = 31.8\%$ for text, $32.1\%$ for vision), but accuracy stays at chance ($\sim$$50\%$). Gemini-2.5-Flash predicts acceptance less often to begin with ($R_A = 70.2\%$ for text under Freeform Boxed Standard), and adding Critical overshoots, so that it predicts rejection too often ($R_R = 75.7\%$ for text, $67.0\%$ for vision) without changing accuracy. Across both model families, prompting can rebalance accept/reject recall but cannot lift accuracy above the majority baseline.

    \textbf{Multi-step peer-review emulation.}
    Multi-turn variants that simulate a reviewer panel, including 5-generation Majority voting, Sequential Refinement Meta-Review~\citep{madaan2025rethinking}, and the single-generation Parallel-Distill-Reflect (PDR) prompt with Advocate, Critic, and Calibrator roles, move recall but not accuracy. Majority voting on Freeform Critical preserves Qwen's single-turn tendency to accept nearly everything ($R_A = 97.3$, $R_R = 4.9$). Few-shot 1A-1R Standard with Majority voting partially rebalances ($R_A = 47.5$, $R_R = 59.1$), and PDR Boxed reaches $R_A = 65.1$, $R_R = 35.2$. None of these multi-turn prompts breaks the $\sim$$50\%$ ceiling on \dataseticlrshort{}, consistent with our main-text observation that base-model prompting alone does not recover the decision signal that SFT later captures from the same backbones.

    \begin{table*}[h!]
        \centering
        \footnotesize
        \captionsetup[subtable]{skip=4pt, justification=centering, singlelinecheck=false}
        \subcaptionbox{Qwen2.5-7B text prompt ablations.\label{tab:prompt_ablations_qwen_text}}[0.49\textwidth]{
        \setlength{\tabcolsep}{2.5pt}
        \scriptsize
        \begin{tabular}{@{}llccc@{}}
            \toprule
            \textbf{Prompt} & \textbf{Modifier} & \textbf{Acc} & $R_A$ & $R_R$ \\
            \midrule
            \multicolumn{5}{c}{\textit{Single Turn}} \\ \midrule
            Freeform Boxed & Standard & 50.2 & 99.4 & 0.9 \\
            Freeform Boxed & Critical & 51.2 & 97.9 & 4.3 \\
            Structured JSON & Standard & 49.9 & 98.3 & 1.4 \\
            Structured JSON & Critical & 49.5 & 67.3 & 31.8 \\
            Fewshot 1A-1R Boxed & Standard & 50.8 & 37.5 & 63.9 \\
            Fewshot 1A-1R Boxed & Critical & 51.6 & 36.0 & 67.2 \\
            Fewshot 1A-1R JSON & Standard & 50.1 & 90.2 & 9.9 \\
            Fewshot 1A-1R JSON & Critical & 51.6 & 68.3 & 34.9 \\
            Fewshot 0A-2R Boxed & Standard & 50.3 & 2.8 & 97.9 \\
            Fewshot 0A-2R Boxed & Critical & 50.2 & 1.9 & 98.4 \\
            Fewshot 0A-2R JSON & Standard & 48.6 & 29.9 & 67.3 \\
            Fewshot 0A-2R JSON & Critical & 49.2 & 22.8 & 75.8 \\
            \midrule
            \multicolumn{5}{c}{\textit{Multi-turn}} \\ \midrule
            Freeform Crit. & Majority & 51.1 & 97.3 & 4.9 \\
            Freeform Crit. & Meta-Review & 50.5 & 96.0 & 4.9 \\
            Fewshot 1A-1R Stand. & Majority & 53.3 & 47.5 & 59.1 \\
            Fewshot 1A-1R Stand. & Meta-Review & 51.5 & 46.2 & 56.9 \\
            Fewshot 1A-1R Crit. & Majority & 50.7 & 31.2 & 70.0 \\
            Fewshot 1A-1R Crit. & Meta-Review & 50.0 & 32.9 & 67.2 \\
            PDR Boxed & --- & 50.1 & 65.1 & 35.2 \\
            \bottomrule
        \end{tabular}
        }
        \hfill
        \subcaptionbox{Qwen2.5-7B vision prompt ablations.\label{tab:prompt_ablations_qwen_vision}}[0.49\textwidth]{
        \setlength{\tabcolsep}{2.5pt}
        \scriptsize
        \begin{tabular}{@{}llccc@{}}
            \toprule
            \textbf{Prompt} & \textbf{Modifier} & \textbf{Acc} & $R_A$ & $R_R$ \\
            \midrule
            \multicolumn{5}{c}{\textit{Single Turn}} \\ \midrule
            Freeform Boxed & Standard & 50.6 & 97.2 & 4.1 \\
            Freeform Boxed & Critical & 52.5 & 46.6 & 58.5 \\
            Structured JSON & Standard & 49.9 & 99.2 & 0.6 \\
            Structured JSON & Critical & 52.3 & 72.6 & 32.1 \\
            Fewshot 1A-1R Boxed & Standard & 50.3 & 28.2 & 72.3 \\
            Fewshot 1A-1R Boxed & Critical & 50.0 & 0.3 & 99.8 \\
            Fewshot 1A-1R JSON & Standard & 50.5 & 92.7 & 8.4 \\
            Fewshot 1A-1R JSON & Critical & 51.1 & 87.0 & 15.3 \\
            Fewshot 0A-2R Boxed & Standard & 50.0 & 2.0 & 97.9 \\
            Fewshot 0A-2R Boxed & Critical & 50.0 & 0.0 & 100.0 \\
            Fewshot 0A-2R JSON & Standard & 52.2 & 54.4 & 50.1 \\
            Fewshot 0A-2R JSON & Critical & 51.0 & 10.9 & 91.0 \\
            \midrule
            \multicolumn{5}{c}{\textit{Multi-turn}} \\ \midrule
            Freeform Crit. & Majority & 50.2 & 41.0 & 59.4 \\
            Freeform Crit. & Meta-Review & 49.6 & 28.0 & 71.1 \\
            Fewshot 1A-1R Stand. & Majority & 50.9 & 24.8 & 77.0 \\
            Fewshot 1A-1R Stand. & Meta-Review & 51.4 & 19.3 & 83.4 \\
            Fewshot 1A-1R Crit. & Majority & 49.7 & 0.1 & 100.0 \\
            Fewshot 1A-1R Crit. & Meta-Review & 49.7 & 0.1 & 100.0 \\
            PDR Boxed & --- & 51.5 & 80.9 & 22.0 \\
            \bottomrule
        \end{tabular}
        }

        \medskip

        \medskip

        \subcaptionbox{Gemini-2.5-Flash prompt ablations (text and vision).\label{tab:prompt_ablations_gemini}}[0.49\textwidth]{
        \setlength{\tabcolsep}{2pt}
        \scriptsize
        \begin{tabular}{@{}ll ccc ccc@{}}
            \toprule
            & & \multicolumn{3}{c}{\textbf{Text}} & \multicolumn{3}{c}{\textbf{Vision}} \\
            \cmidrule(lr){3-5} \cmidrule(lr){6-8}
            \textbf{Prompt} & \textbf{Mod.} & \textbf{Acc} & $R_A$ & $R_R$ & \textbf{Acc} & $R_A$ & $R_R$ \\
            \midrule
            Freeform Boxed & Stand. & 51.8 & 70.2 & 33.3 & 51.5 & 79.0 & 24.1 \\
            Freeform Boxed & Crit. & 50.9 & 26.1 & 75.7 & 48.4 & 29.9 & 67.0 \\
            Struct. JSON & Stand. & 49.9 & 92.8 & 7.0 & 50.6 & 92.3 & 9.0 \\
            Struct. JSON & Crit. & 51.6 & 63.5 & 39.8 & 54.0 & 58.2 & 50.0 \\
            FS 1A-1R Boxed & Stand. & 50.1 & 84.9 & 15.5 & 49.0 & 79.6 & 18.4 \\
            FS 1A-1R Boxed & Crit. & 47.6 & 62.2 & 33.0 & 51.0 & 55.9 & 46.1 \\
            FS 0A-2R Boxed & Stand. & 50.8 & 75.7 & 25.9 & 48.8 & 78.1 & 19.4 \\
            FS 0A-2R Boxed & Crit. & 48.5 & 55.5 & 41.5 & 50.5 & 50.8 & 50.0 \\
            \bottomrule
        \end{tabular}
        }
        \hfill
        \subcaptionbox{Qwen3.5 prompt-time model scaling on a $500$-paper \dataseticlrshort{} subset under Freeform Boxed Standard.\label{tab:qwen35_metrics}}[0.49\textwidth]{
        \setlength{\tabcolsep}{2pt}
        \scriptsize
        \begin{tabular}{@{}ll ccc ccc@{}}
            \toprule
            \multirow{2}{*}{\textbf{Model}} & \multirow{2}{*}{\textbf{Mod.}} & \multicolumn{3}{c}{\textbf{'25}} & \multicolumn{3}{c}{\textbf{'26}} \\
            \cmidrule(lr){3-5} \cmidrule(lr){6-8}
            & & \textbf{Acc} & $R_A$ & $R_R$ & \textbf{Acc} & $R_A$ & $R_R$ \\
            \midrule
            Qwen3.5-9b & text & 52.2 & 92.7 & 11.8 & 52.5 & 96.0 & 9.1 \\
            Qwen3.5-9b & vis. & 50.0 & 98.5 & 1.5 & 51.0 & 100.0 & 2.0 \\
            Qwen3.5-35b-a3b & text & 55.9 & 88.2 & 23.5 & 50.5 & 84.9 & 16.2 \\
            Qwen3.5-35b-a3b & vis. & 55.9 & 98.5 & 13.2 & 50.5 & 97.0 & 4.0 \\
            Qwen3.5-27b & text & 47.8 & 94.1 & 1.5 & 53.0 & 100.0 & 6.1 \\
            Qwen3.5-27b & vis. & 54.4 & 85.3 & 23.5 & 57.1 & 88.9 & 25.3 \\
            Qwen3.5-122b-a10b & text & 53.7 & 83.8 & 23.5 & 52.0 & 96.0 & 8.1 \\
            Qwen3.5-122b-a10b & vis. & 63.2 & 91.2 & 35.3 & 56.6 & 89.9 & 23.2 \\
            \bottomrule
        \end{tabular}
        }

        \caption{\textbf{Prompt-ablation averages across ICLR '25 and '26.} We report accuracy (Acc), accept recall ($R_A$), and reject recall ($R_R$) for each prompt format and modifier on \dataseticlrshort{}. Across all open-source, Gemini, and Qwen3.5 variants, accuracy hovers around chance.}
        \label{tab:prompt_ablations_all}
    \end{table*}

    \textbf{Model size ablation.}
    To check whether, with a stronger base model, scale helps prompting (Qwen3.5 is generally stronger than the Qwen2.5 family we have been using), we evaluated the Qwen3.5 family ($9$B, $27$B, $35$B-MoE, $122$B-MoE) under Freeform Boxed Standard on a $500$-paper \dataseticlrshort{} subset (\Cref{tab:qwen35_metrics}). Scale alone does not consistently lift base prompting above chance. Even at $122$B vision, the strongest cell only reaches $63.2\%$/$56.6\%$ Acc on '25/'26, well below the SFT and RL models reported in the main text.

\section{SFT and RL Method Details}\label{app:training}

    \subsection{SFT}\label{appendix:training_settings}

    SFT uses cosine decay with 5 warmup steps, context length 24,480 tokens, and greedy
    evaluation. The strongest text configuration uses batch size 32 with a learning rate of
    $10^{-6}$. The strongest vision configuration uses batch size 16 with the same learning
    rate. We sweep \dataseticlrshort{} and \datasetarxivname{} training sources, 50/50 versus natural-rate class priors,
    data scale, text versus screenshot inputs, text+vision variants, and suffixes containing
    structured paper statistics or Qwen/Gemini-generated reviews. For vision efficiency, we
    also test the panel setup in Table~\ref{tab:panel_ablation}, which tiles ten pages into one
    2380$\times$1512 image. Unless a row is explicitly marked otherwise, 3B SFT runs train for 4
    epochs and 7B/14B SFT runs train for 2 epochs.

    \vspace{-1.2em}
    \begin{figure}[H]
        \centering
        \includegraphics[width=0.92\columnwidth]{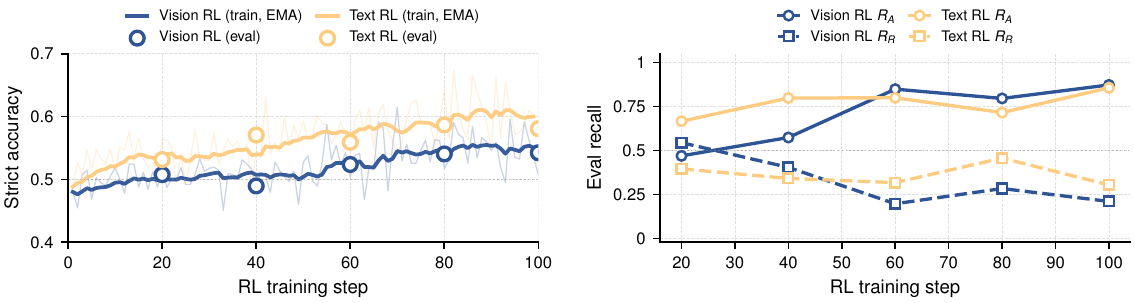}
        \caption{\textbf{RL training curves: strict accuracy and per-class recall.} \emph{Left:} strict-accuracy training estimate (EMA-smoothed; raw curve in faint background) and held-out eval accuracy on 1{,}667 '25/'26 papers. The training estimate is inverted from the graded reward via $\mathrm{acc} \approx (r - 0.2 \cdot b)/(1 - 0.2 \cdot b)$ where the boxed-rate $b \approx 1.0$ on every eval. \emph{Right:} per-class accept ($R_A$, circles, solid) and reject ($R_R$, squares, dashed) recall on the same eval set. Both policies show graded reward improving during training, but their pre-existing accept bias is preserved and tightens further as RL continues.}
        \label{fig:appendix-rl-reward}
    \end{figure}

    \subsection{RL}\label{appendix:rl}

    We train two single-turn GRPO policies following DeepSeek-R1~\citep{Guo_2025}, one text and one vision-language, on \dataseticlrshort{}. Both apply the reasoning prompt variant from \appref{appendix:prompting} on the base Qwen2.5-7B (text) and Qwen2.5-VL-7B-Instruct (vision) models, and use the composite accuracy-plus-format reward described in the main text. \Cref{fig:appendix-rl-reward} shows training trajectories through step $100$ on a 1{,}667-paper held-out '25/'26 \dataseticlrshort{} test set. Both policies show graded reward climbing during training, but the per-class recall panel exposes that the underlying accept/reject mix barely shifts. As the models learn to satisfy the format reward, their pre-existing accept bias is preserved and even tightens further during RL. Neither policy surpasses the strongest SFT model from the main text. The graded reward partially hides the swing because every well-formed prediction already earns a $0.2$ format floor before the decision term contributes anything.

    \textbf{Settings.}
    The text RL policy uses a learning rate of $1.2\times 10^{-6}$ with constant warmup over 4 steps, max gradient norm $0.5$, train batch size $256$, policy mini-batch size $128$, micro-batch size $2$ per GPU, one epoch, $8$ rollouts per prompt, max response length $4096$, and max prompt length $25{,}480$. The vision RL policy uses Qwen2.5-VL-7B-Instruct with a learning rate of $10^{-6}$, train batch size $64$, PPO mini-batch size $32$, micro-batch size $2$ per GPU, one epoch, $8$ rollouts per prompt, max response length $2046$, and max prompt length $24{,}480$. Each paper is represented by roughly $10$ PNG screenshots plus text.

    \subsection{Calibration Details}\label{appendix:calibration}

    Modern neural networks are often poorly calibrated, producing confidence estimates that do not reflect true correctness probabilities~\citep{Guo_2017}. We apply post-hoc temperature scaling to our SFT models and to Jev, with a base-rate term so the probabilities hold at a venue's real acceptance rate.

    \textbf{Expected calibration error.}
    Following \citet{Guo_2017}, we first convert each prediction into confidence in the chosen decision: $c_i=p_i$ when $\hat z_i{=}1$ and $c_i=1-p_i$ when $\hat z_i{=}0$. We partition these confidences into $M$ equal-width bins $B_1,\ldots,B_M$ and define the \emph{expected calibration error} (ECE) as
    \begin{equation}
        \mathrm{ECE} = \sum_{m=1}^{M} \frac{|B_m|}{n} \left| \mathrm{acc}(B_m) - \mathrm{conf}(B_m) \right|,
    \end{equation}
    where $\mathrm{acc}(B_m)=|B_m|^{-1}\sum_{i\in B_m}\mathbf{1}[\hat z_i=z_i]$ is the empirical accuracy and $\mathrm{conf}(B_m)=|B_m|^{-1}\sum_{i\in B_m}c_i$ is the mean confidence in that bin. ECE is therefore the average discrepancy between how confident the model is and how often it is correct; a perfectly calibrated model has $\mathrm{ECE}=0$.

    \textbf{Raw calibration.}
    Before scaling, the text models are close to calibrated, with ECE $0.027$ on \dataseticlrshort{} and $0.043$ on \datasetarxivname{} (balanced test). The vision models are overconfident, with ECE $0.049$ and $0.069$: their confidence exceeds their accuracy.

    \textbf{Temperature scaling with a base-rate term.}
    Temperature scaling uses a single parameter $\tau > 0$. From the decision-token log-probabilities we take the accept log-odds $\ell = \log P(\text{Accept}) - \log P(\text{Reject})$. Our models are trained on a 50/50 mix, so $\ell$ measures the evidence under a balanced prior. To score papers at acceptance rate $\pi$ we add the prior log-odds and rescale:
    \begin{equation}
        p_{\mathrm{cal}} = \sigma\big(\ell / \tau + \log(\pi/(1-\pi))\big),
    \end{equation}
    where $\sigma$ is the sigmoid function. $\tau > 1$ softens overconfident predictions and $\tau < 1$ sharpens them. We fit $\tau$ by minimizing the negative log-likelihood (NLL) on the natural-rate validation split, with $\pi$ set to that split's acceptance rate ($31\%$ for \dataseticlrshort{}, $23\%$ for \datasetarxivname{}), so the shift absorbs the base rate and $\tau$ only sets how sharp the probabilities are. On a balanced set $\pi = 0.5$ and the shift vanishes. Jev's scores already sit near the real acceptance rate, so we fit its $\tau$ without the shift and, on balanced sets, subtract $\log(\pi/(1-\pi))$ instead. Since $p_{\mathrm{cal}} \ge \pi$ exactly when $\ell \ge 0$, neither $\tau$ nor the shift ever changes a decision.

    We do not minimize ECE directly because (i) the binning operation is non-differentiable, (ii) ECE is sensitive to the number of bins, and (iii) NLL is a strictly proper scoring rule whose global optimum is achieved if and only if predicted probabilities equal true probabilities. ECE, by contrast, can be zero even with poorly separated and rank-uninformative predictions.

\begin{figure}[H]
        \centering
        \includegraphics[width=\textwidth]{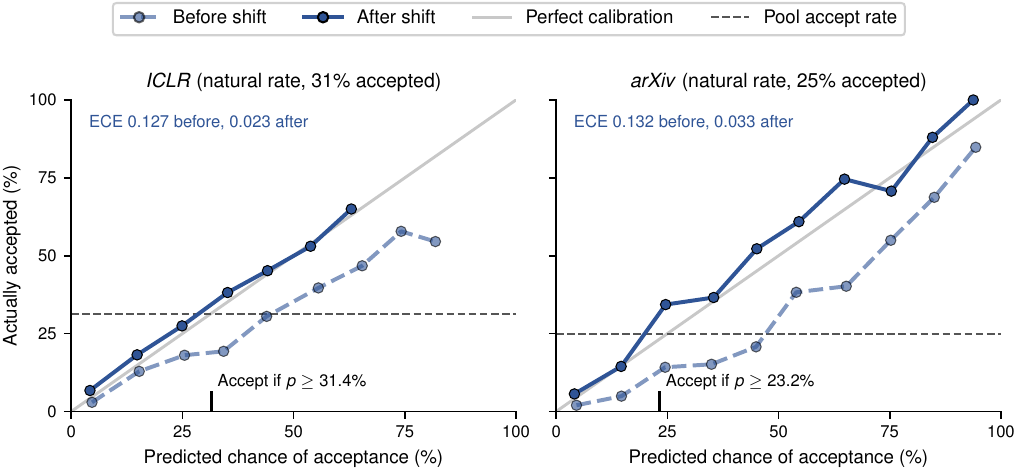}
        \caption{\textbf{Base-rate shift at deployment.} Share of papers actually accepted vs.\ predicted chance of acceptance for \systemvision{} on the natural-rate \dataseticlrshort{} (left) and \datasetarxivname{} (right) test sets, before (dashed) and after (solid) adding $\log(\pi/(1-\pi))$. Ticks mark the decision rule $p_{\mathrm{cal}} \ge \pi$, the same decisions as $\ell \ge 0$.}
        \label{fig:calibration_deployment}
    \end{figure}

    \textbf{Calibration at the real acceptance rate.}
    On a natural-rate set we measure the calibration of the acceptance probability itself: we bin papers by $p_{\mathrm{cal}}$ into ten equal-width bins on $[0,1]$ and take the size-weighted gap between each bin's mean $p_{\mathrm{cal}}$ and the fraction of its papers actually accepted. \Cref{tab:temperature} and Figure~\ref{fig:calibration_deployment} evaluate the shift on the natural-rate test sets. Without it, the probabilities stay centered for a 50/50 world and overstate acceptance (ECE $.13$--$.15$). With it, ECE falls to $.023$--$.042$ for \systemfamily{}. At this rate an \dataseticlrshort{} Accept call by \systemvision{} is right $47\%$ of the time and a Reject call $82\%$ ($56\%$ and $91\%$ on \datasetarxivname{}). The shifted probabilities slightly understate acceptance: their mean is $29.3\%$ where $31.3\%$ of \dataseticlrshort{} papers are accepted, and $21.8\%$ where $24.8\%$ of \datasetarxivname{} papers are.

    \textbf{Robustness.}
    Fitting $\tau$ on a balanced validation split instead, and adding the same shift, gives similar results (ECE $.030$--$.049$). A two-parameter Platt fit~\citep{Platt_1999}, $\sigma(a\ell+b)$ on the natural-rate split, is slightly better still ($0.017$--$0.031$) and also calibrates Jev ($0.018$--$0.021$). Jev's calibrated probabilities, however, never exceed about $60\%$, so unlike \systemfamily{} it cannot single out likely accepts.

    \textbf{Caveats.}
    The \datasetarxivname{} model's $\pi$ is a mix of venue rates (about $23\%$); a single venue's rate can be plugged in as $\pi$. \datasetarxivname{} rejects are inferred, never matched to a proceedings entry, rather than confirmed.

    \begin{table}[H]
        \centering
        \footnotesize
        \setlength{\tabcolsep}{4.5pt}
        \renewcommand{\arraystretch}{1.1}
        \begin{tabular}{@{}llcccccccc@{}}
            \toprule
            & & \multicolumn{2}{c}{$\tau$} & \multicolumn{2}{c}{Balanced test ECE} & \multicolumn{4}{c}{Natural-rate test ECE of $p_{\mathrm{cal}}$} \\
            \cmidrule(lr){3-4}\cmidrule(lr){5-6}\cmidrule(l){7-10}
            \textbf{Source} & \textbf{Model} & Ours & Bal. val & Raw & Ours & No shift & Ours & Bal.-val $\tau$ & Platt \\
            \midrule
            \dataseticlrshort{} & \systemtext{}   & 1.14 & 1.21 & .027 & .023 & .128 & .038 & .038 & .029 \\
            \dataseticlrshort{} & \systemvision{} & 1.25 & 1.59 & .049 & .031 & .127 & .023 & .030 & .022 \\
            \dataseticlrshort{} & Jev             & 0.74 & 2.99 & .166 & .051 & .060 & .060 & .058 & .021 \\
            \datasetarxivname{} & \systemtext{}   & 1.46 & 1.64 & .043 & .029 & .148 & .042 & .049 & .031 \\
            \datasetarxivname{} & \systemvision{} & 1.50 & 1.59 & .069 & .024 & .132 & .033 & .034 & .017 \\
            \datasetarxivname{} & Jev             & 0.66 & 3.15 & .165 & .024 & .025 & .025 & .055 & .018 \\
            \bottomrule
        \end{tabular}
        \caption{\textbf{Temperature scaling with a base-rate term.} $\tau$ is fit on natural-rate validation with the prior term (ours; Jev without it, since its scores already sit at the real rate) or on balanced validation (for comparison). Balanced test: ECE of confidence before and after scaling. Natural-rate test: ECE of the acceptance probability without the shift, with it (ours), with the shift but $\tau$ fit on balanced validation, and with a two-parameter Platt fit.}
        \label{tab:temperature}
    \end{table}

    \subsection{Hardware Details}\label{appendix:hardware}

    All SFT and RL experiments reported in the paper run on a single node with $4\times$ NVIDIA H200 GPUs. Per-eval inference for the held-out checkpoints uses the same node.

\section{Statistical Uncertainty}\label{appendix:statistical_significance}

    For uncertainty estimates, we use paired bootstrap resampling so each resample compares models on the same papers. We resample paper IDs 10,000 times with replacement and recompute Acc and AUC on each resample. \Cref{tab:bootstrap_significance} reports percentile 95\% two-sided CIs for the four headline \systemvision{} comparisons. On \dataseticlrshort{}, the Acc gap against Gemini-3.1-Pro is $+3.8\%$ with CI $[+0.4, +7.3]$, and the AUC gap against GPT-5.5 is $+.066$ with CI $[+.034, +.098]$. On \datasetarxivname{}, the Acc gap against Gemini-3.1-Pro is $+8.7\%$ (CI $[+4.5, +12.7]$) and the AUC gap against GPT-5.5 is $+.097$ (CI $[+.061, +.135]$). Since our claim is directional, we test it one-sided; all four are significant ($p\le.014$). The same holds for every large frontier configuration: against Gemini-3.1-Pro and GPT-5.5 with either text or vision input, with decisions thresholded as in Table~\ref{tab:main_results}, the one-sided $p$ is below $.05$ for both Acc and AUC on both datasets (largest $p=.040$, Acc against GPT-5.5 with text input on \dataseticlrshort{}).

\begin{table}[h!]
\centering
\footnotesize
\setlength{\tabcolsep}{3.4pt}
\renewcommand{\arraystretch}{1.08}
\begin{tabular}{@{}llcccccc@{}}
\toprule
\textbf{Dataset} & \textbf{Metric} & \textbf{\systemvision{}} & \textbf{Large frontier baseline} & \textbf{Baseline score} & \textbf{$\Delta$} & \textbf{95\% CI} & \textbf{$p$} \\
\midrule
\dataseticlrshort{} & Acc & 68.9 & Gemini-3.1-Pro & 65.1 & \textbf{+3.8} & [+0.4, +7.3] & .014 \\
\dataseticlrshort{} & AUC & .759 & GPT-5.5 & .693 & \textbf{+.066} & [+.034, +.098] & $<$.001 \\
\datasetarxivname{} & Acc & 77.1 & Gemini-3.1-Pro & 68.5 & \textbf{+8.7} & [+4.5, +12.7] & $<$.001 \\
\datasetarxivname{} & AUC & .841 & GPT-5.5 & .743 & \textbf{+.097} & [+.061, +.135] & $<$.001 \\
\bottomrule
\end{tabular}
\caption{\textbf{Paired bootstrap significance for headline vision comparisons.} Intervals are 95\% percentile intervals from 10,000 paired bootstrap resamples over paper IDs; $p$ is one-sided (\systemvision{} better).}
\label{tab:bootstrap_significance}
\end{table}

    \Cref{tab:vision_text_significance} compares vision with text on each dataset using the same bootstrap framework. Vision beats text in five of the six comparisons. We test this formally only for \systemfamily{}: its vision and text models are scored on the same papers, and vision is significantly better on both datasets ($p=.012$ and $.005$, one-sided, corrected for the two tests with Holm--Bonferroni). For the frontier models we report the gaps for reference only. Once their decisions are thresholded as in Table~\ref{tab:main_results}, the gaps range from $-1.0$ to $+2.3$, and none is statistically significant.

\begin{table}[h!]
\centering
\footnotesize
\setlength{\tabcolsep}{3.8pt}
\renewcommand{\arraystretch}{1.08}
\begin{tabular}{@{}llccccc@{}}
\toprule
\textbf{Dataset} & \textbf{System} & \textbf{Vision Acc} & \textbf{Text Acc} & \textbf{$\Delta$ Acc} & \textbf{95\% CI} & \textbf{$p$} \\
\midrule
\dataseticlrshort{} & Gemini-3.1-Pro & 65.1 & 62.8 & +2.3 & [-1.94, +6.67] & .164$^\dagger$ \\
\dataseticlrshort{} & GPT-5.5 & 63.4 & 64.4 & -1.0 & [-5.29, +3.32] & .657$^\dagger$ \\
\dataseticlrshort{} & \systemfamily{} & 68.3 & 66.0 & \textbf{+2.3} & [+0.34, +4.16] & .012 \\
\datasetarxivname{} & Gemini-3.1-Pro & 68.5 & 67.7 & +0.8 & [-3.78, +5.38] & .359$^\dagger$ \\
\datasetarxivname{} & GPT-5.5 & 68.3 & 66.0 & +2.3 & [-2.22, +6.84] & .168$^\dagger$ \\
\datasetarxivname{} & \systemfamily{} & 77.5 & 74.2 & \textbf{+3.3} & [+1.07, +5.52] & .005 \\
\bottomrule
\end{tabular}
\caption{\textbf{Dataset-stratified modality comparisons.} Intervals are 95\% bootstrap intervals for the vision-minus-text Acc difference. For \systemfamily{}, $p$ tests whether vision beats text, paired on the same papers and corrected for the two tests (Holm--Bonferroni). $^\dagger$Frontier rows are for reference only: uncorrected one-sided $p$ (stratified randomization), with decisions thresholded as in Table~\ref{tab:main_results}.}
\label{tab:vision_text_significance}
\end{table}

    \textbf{Correlative analysis.}
    Binary accept/reject training yields a continuous score that also tracks reviewer ratings and later citations. On \dataseticlrshort{}, \systemvision{} has the strongest \systemfamily{} rating alignment ($\rho_R=+.48$), while \systemtext{} at 14B is strongest within \systemfamily{} for '25 citation alignment ($\rho_C=+.31$). On \datasetarxivname{}, reviewer ratings and citations are available for fewer papers than decisions are, so we treat them as diagnostics and leave them out of the headline comparisons. On \dataseticlrshort{}~'25, the one slice with both full reviewer ratings and citations ($n=678$), \systemvision{}'s $p_{\mathrm{accept}}$ tracks later citations about as well as the human review score ($\rho\approx.25$ vs.\ $.26$), and accepted papers outcite rejected ones (citation AUC $.64$). We cannot run this check on \datasetarxivname{}, whose citations come only from proceedings-matched accepts, leaving too few rejected papers to compare.

    \textbf{Frontier model repeatability.}
    We repeated a subset of the frontier-model inference calls on the same held-out \datasetarxivname{}-text papers to estimate run-to-run noise. GPT-5.4 is almost deterministic under the same prompt, while Gemini-3.1-Pro has meaningful stochasticity. Decision agreement is 89.4\% and the second pass is 4.5\% lower on the same papers (Table~\ref{tab:frontier_repeatability}). We therefore treat GPT-5.4 point estimates as stable, while Gemini estimates should be read with several points of implicit run noise unless repeated passes are averaged. This repeatability audit was run on GPT-5.4, the frontier model used before we re-ran the main GPT baseline on GPT-5.5; we did not separately re-audit GPT-5.5's run-to-run noise.

    \begin{table}[H]
        \centering
        \footnotesize
        \setlength{\tabcolsep}{4pt}
        \renewcommand{\arraystretch}{1.08}
        \begin{tabular}{@{}lcc@{}}
            \toprule
            \textbf{Metric on repeated \datasetarxivname{}-text papers} & \textbf{Gemini-3.1-Pro} & \textbf{GPT-5.4} \\
            \midrule
            Paired papers & 199 & 208 \\
            Decision agreement & 89.4\% & 98.1\% \\
            Pass-1 accuracy & 64.8\% & 57.7\% \\
            Pass-2 accuracy & 60.3\% & 57.7\% \\
            Pass-1 right, pass-2 wrong & 7.5\% & 1.0\% \\
            Pass-1 wrong, pass-2 right & 3.0\% & 1.0\% \\
            Mean absolute rating delta & 0.61 & 0.40 \\
            Rating within $\pm$1 & 89.4\% & 98.6\% \\
            \bottomrule
        \end{tabular}
        \caption{\textbf{Frontier repeatability audit.} We repeat \datasetarxivname{}-text frontier inference on the same held-out papers and measure decision, accuracy, and rating stability across two passes.}
        \label{tab:frontier_repeatability}
    \end{table}

    \section{Applications to Scientific Systems}
    \label{sec:applications}
    \label{sec:application}

    \systemfamily{} produces a ranking signal that is empirically proven: its high AUC and rating
    correlation show that the score ranks accepted papers above rejected ones and agrees with reviewer ratings, and temperature scaling makes that score
    calibrated after the fact. This invites two applications where a calibrated estimate of acceptance probability
    matters. In literature retrieval, we want to surface work likely to be accepted among topically similar
    candidates. In agentic reviewing, a generated critique relies on a well-calibrated prior over
    the final decision. We study both settings below.

    \begin{figure*}[h!]
        \centering
        \includegraphics[width=0.82\textwidth]{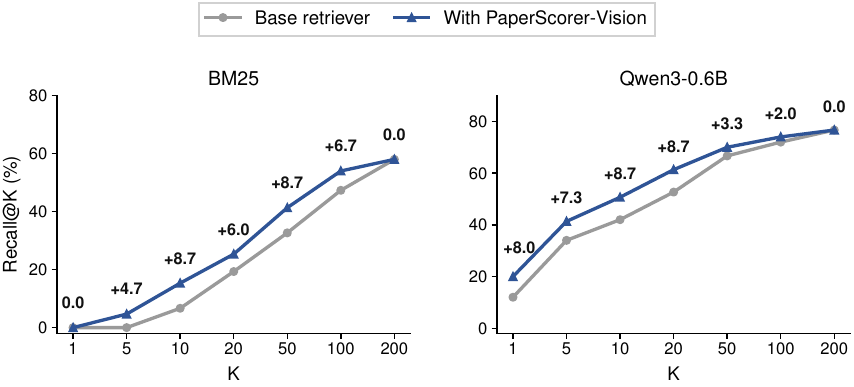}
        \vspace{-0.35em}
        \caption{\textbf{LitSearch reranking.} We evaluate Recall@K on 148 hard LitSearch
        queries whose gold papers are missed by BM25 at K=5, using each retriever's top-200
        candidates. Labels show the Recall@K gain from adding \systemvision{}.}
        \label{fig:litsearch_rerank}
        \vspace{-0.75em}
    \end{figure*}

    \subsection{Ranking Scientific Literature for Retrieval}
    \label{sec:litsearch_application}

    Scientific search systems must retrieve a target paper from many plausible topical matches.
    LitSearch \citep{ajith2024litsearch} captures this setting with natural-language literature
    queries paired with one or more gold cited papers. We evaluate the subset linked to our \datasetarxivname{}
    pool: 478 queries have at least one gold \datasetarxivname{} paper, and 250 have all gold \datasetarxivname{} IDs inside
    the \systemfamily{} \datasetarxivname{} corpus. To isolate the regime where reranking matters, we use
    the 148 hard queries for which BM25 has zero Recall@5, covering 150 gold papers.

    We use \systemfamily{} as a query-independent reranker based on its estimated acceptance probability. For each hard query, BM25
    and Qwen3-Embedding-0.6B \citep{zhang2025qwen3embeddingadvancingtext} first retrieve their
    own top-200 candidate papers. We then score each candidate once with \datasetarxivname{}-trained \systemvision{} (7B checkpoint,
    matching the main results); the 3B checkpoint (not shown) reranks almost as well, within a few percent at
    every $K$, consistent with our finding that decision accuracy
    is largely insensitive to model scale (\Cref{sec:main_results}). Let $s(q,d)$ be the retriever score for candidate
    $d$ under query $q$, let $z(\cdot)$ denote $z$-normalization within the query's candidate list,
    and let $p_{\mathrm{accept}}(d)$ be the \systemfamily{} acceptance score. We rank candidates by a fixed
    blend of the two, with the weight on $p_{\mathrm{accept}}$ chosen on a held-out validation split of queries:
    \begin{equation*}
        0.3\,z(s(q,d)) + 0.7\,p_{\mathrm{accept}}(d).
    \end{equation*}

    Retrieval supplies the topical neighborhood, while \systemvision{} reorders candidates by their
    estimated acceptance probability. \Cref{fig:litsearch_rerank} shows that this helps both retrievers:
    whether candidates come from sparse BM25 or a dense Qwen embedding, reranking pulls more of the
    gold cited papers toward the top of the list. The gain is largest near the top, where retrieval
    has already gathered topically similar work and the only open question is which paper is more likely to be accepted.
    Deeper down, where topical relevance still does most of the sorting, the prior adds little, so
    \systemvision{} acts as a prior on acceptance layered on retrieval, not a substitute for it.

    \subsection{Grounding Agentic Reviewers}
    \label{sec:agent_review_application}

    Prior work shows that LLM reviews can yield actionable suggestions for improving papers \citep{thakkar2026llmfeedback}, but our prompt-tuning results (\Cref{inf_results}) suggest that the predicted decisions poorly track ground-truth outcomes. As noted in \Cref{sec:probform}, papers $X$ are transformed into reviews $R$ before reaching a decision $Z$, so the joint distribution over reviews and decisions factors as $p(R, Z \mid X)$. While the autoregressive flow places the review before the decision, now that we have built a shortcut that predicts the decision without the review, we can ask whether we can generate more \emph{decision-calibrated} reviews by rearranging the joint:
    \begin{equation*}
        p(R, Z \mid X) = p(Z \mid R, X)\,p(R \mid X) \;\Longrightarrow\;
        \overbrace{p(R \mid Z, X)}^{\text{Frontier Model}}\,\overbrace{p(Z \mid X)}^{\text{\systemfamily{}}}.
    \end{equation*}
    Both factorizations are valid. Conceptually, if our conjecture is correct --- that much of the decision is determined by overall presentation quality --- then reviews may serve as a means-to-an-end to justify the reviewer's initial reaction. This follows from psychology research on human decision-making, whereby we often rely on initial judgments and post-hoc rationalization more than we consciously attribute \citep{nisbett1977telling,kunda1990motivated,haidt2001emotional}, with low-level feelings of ``goodness'' or ``badness'' driving longer-form judgments and reasoning \citep{slovic2007affect}.

    We therefore ask whether conditioning frontier coding agents on the \systemfamily{} signal can (1) improve their decision-prediction and (2) improve their generated reviews' alignment with human reviews. We use \systemvision{} as the calibrated prior, and let Claude Code write the review.

    \begin{table*}[h!]
        \centering
        \captionsetup[subtable]{skip=8pt}
        \subcaptionbox{\textbf{Reviewer accuracy and human-review alignment.} `Cn.', `Wk.', `Nov.', and `Ql.' denote contribution, weakness, novel-concern, and overall-quality alignment.\label{tab:agent_results_alignment}}[0.60\textwidth]{%
            \parbox[c][1.15in][c]{\linewidth}{%
                \centering
                \footnotesize
                \setlength{\tabcolsep}{2pt}
                \renewcommand{\arraystretch}{1.12}
                \begin{tabular}{p{0.29\linewidth}cccccccc}
                    \toprule
                    & \multicolumn{3}{c}{\textbf{Acc.\,(\%)}} & \multicolumn{5}{c}{\textbf{Alignment (1--5)}} \\
                    \cmidrule(lr){2-4}\cmidrule(lr){5-9}
                    \textbf{Method} & \textbf{'25} & \textbf{'26} & \textbf{All} & \textbf{Cn.} & \textbf{Wk.} & \textbf{Nov.} & \textbf{Ql.} & \textbf{Mean} \\
                    \midrule
                    \systemvision{} & 72.0 & 68.0 & 69.6 & -- & -- & -- & -- & -- \\
                    \midrule
                    Claude Code & 57.5 & 66.7 & 63.0 & 4.96 & 4.08 & 3.98 & 4.11 & 4.28 \\
                    ~+\systemvision{} & \textbf{70.5} & \textbf{70.7} & \textbf{70.6} & 4.87 & \textbf{4.23} & \textbf{4.10} & \textbf{4.14} & \textbf{4.33} \\
                    \midrule
                    $\Delta$ & \textcolor{green!50!black}{+13.0} & \textcolor{green!50!black}{+4.0} & \textcolor{green!50!black}{+7.6} & -0.09 & \textcolor{green!50!black}{+0.15} & \textcolor{green!50!black}{+0.11} & \textcolor{green!50!black}{+0.03} & \textcolor{green!50!black}{+0.05} \\
                    \bottomrule
                \end{tabular}
            }%
        }
        \hfill
        \subcaptionbox{\textbf{Decision transitions with \systemvision{}.} How agent decisions shift under the prior.\label{tab:agent_results_transitions}}[0.36\textwidth]{%
            \parbox[c][1.15in][c]{\linewidth}{%
                \centering
                \footnotesize
                \setlength{\tabcolsep}{2.5pt}
                \renewcommand{\arraystretch}{1.12}
                \begin{tabular}{@{}lcccccc@{}}
                    \toprule
                    & \multicolumn{2}{c}{\textbf{'25}} & \multicolumn{2}{c}{\textbf{'26}} & \multicolumn{2}{c}{\textbf{Overall}} \\
                    \cmidrule(lr){2-3}\cmidrule(lr){4-5}\cmidrule(lr){6-7}
                    \textbf{Transition} & \textbf{\%} & \textbf{Acc.} & \textbf{\%} & \textbf{Acc.} & \textbf{\%} & \textbf{Acc.} \\
                    \midrule
                    No change & 75.0 & 68.7 & 80.7 & 73.1 & 78.4 & 71.4 \\
                    \rowcolor{green!10} A $\to$ R & 21.0 & \textbf{81.0} & 8.7 & 73.1 & 13.6 & \textbf{77.9} \\
                    R $\to$ A & 4.0 & 50.0 & 10.7 & 50.0 & 8.0 & 50.0 \\
                    \midrule
                    All flips & 25.0 & 76.0 & 19.3 & 60.3 & 21.6 & 67.6 \\
                    \bottomrule
                \end{tabular}
            }%
        }
        \caption{\textbf{Decision-conditioned review generation.} Claude Code is evaluated on 500 \dataseticlrshort{} papers under a 30/70 accept/reject split, with and without the calibrated \systemvision{} prior. Conditioning improves accuracy from 63.0\% to 70.6\% and review alignment, with the prior catching Claude's remaining incorrect acceptance predictions.}
        \label{tab:agent_results}
        \vspace{-0.5em}
    \end{table*}

    We evaluate Claude Code (Sonnet-4.6) on 500 \dataseticlrshort{} papers. The agent receives \systemvision{}'s decision, $p_{\mathrm{accept}}$, and confidence bands, and can inspect text, rendered pages, and the temporally filtered \datasetarxivname{} retrieval. It emits a schema-constrained review and decision. Codex (GPT-5.5, medium reasoning) grades reviews against human reviews (review generation and alignment grading prompts in \appref{appendix:claude_review_prompts}).

    \Cref{tab:agent_results_alignment} shows that Claude alone \emph{underperforms} \systemvision{}, but adding the calibrated prior improves overall accuracy from 63.0\% to 70.6\%, surpassing both individually. This matches the factorization above. \systemvision{} anchors the decision while the reviewer supplies the critique. The gain comes mainly from a 13.6\% subset of papers flipping Accept$\to$Reject at 77.9\% accuracy (\Cref{tab:agent_results_transitions}). Reject$\to$Accept flips are rarer and much less reliable. Although Claude predicts rejection more often than acceptance on the balanced evaluation (\Cref{tab:main_results}), on this 30/70 subset its remaining incorrect acceptance predictions dominate the correction signal, which the calibrated accept probability catches by pulling those papers below the decision threshold the agent then respects.

    Conditioning on \systemvision{} also improves alignment with human reviews across every factor except contributions (\Cref{tab:agent_results_alignment}). Because the prior corrects Claude's accept-bias, the agent emits weakness and novelty critiques at a higher rate, bringing the generated review closer to the human ground truth. Contribution alignment was already saturated under Claude alone, and is preserved under conditioning.

    \subsection{Agentic Reviewer Prompts}\label{appendix:claude_review_prompts}

    This subsection records the prompt scaffolding for the discussion experiment. Claude Code (Sonnet-4.6) performs the review, while review grading is performed by Codex (GPT-5.5) with medium reasoning.

    \textbf{Reviewer agent (no prior).}
    \begin{Verbatim}[fontsize=\scriptsize,frame=single]
    You are an ICLR reviewer. Review the paper with submission_id=<SID>.

    STEPS:
    1. Read the paper content from papers/<SID>/content.json
    2. Examine the page images listed in the content JSON under img_pages
    3. Write a structured ICLR review as JSON with all 11 fields:
       - decision: "accept" or "reject"
       - rating: 0-10
       - confidence: 1-5
       - soundness: 1-4
       - presentation: 1-4
       - contribution: 1-4
       - summary: 2-3 sentence summary
       - strengths: numbered list citing specific sections/figures/theorems
       - weaknesses: numbered list citing specifics (no generic criticisms)
       - questions: numbered list of questions for authors
       - missing_references: prior work explicitly mentioned in the paper

    4. Write the review JSON to agent_reviews/<SID>.json using the Write tool.
    5. Return ONLY "Done: <SID>" -- do NOT return the full review text.
    \end{Verbatim}

    \textbf{Reviewer agent (with \systemvision{} prior).}
    \begin{Verbatim}[fontsize=\scriptsize,frame=single]
    You are an ICLR reviewer. Review the paper with submission_id=<SID>.

    PAPERSCORER PRIOR:
    - Model prediction: <accept|reject>
    - Model confidence: <0.xxx>

    STEPS:
    1. Read the paper content from papers/<SID>/content.json
    2. Examine the page images listed in the content JSON under img_pages
    3. Use the model prediction as a Bayesian prior -- weight it by the
       confidence (high confidence -> trust strongly, low -> use own judgment).
    4. Write a structured ICLR review as JSON with all 11 fields:
       - decision: "accept" or "reject"
       - rating: 0-10
       - confidence: 1-5
       - soundness: 1-4
       - presentation: 1-4
       - contribution: 1-4
       - summary: 2-3 sentence summary
       - strengths: numbered list citing specific sections/figures/theorems
       - weaknesses: numbered list citing specifics (no generic criticisms)
       - questions: numbered list of questions for authors
       - missing_references: prior work explicitly mentioned in the paper

    5. Write the review JSON to agent_reviews/<SID>.json using the Write tool.
    6. Return ONLY "Done: <SID>" -- do NOT return the full review text.
    \end{Verbatim}

        \textbf{Judge prompt (Codex GPT-5.5, medium reasoning).}
        \label{app:codex_judge}
        \begin{Verbatim}[fontsize=\scriptsize,frame=single]
    You are an expert meta-reviewer evaluating AI-generated paper reviews against
    real ICLR human reviews and the Area Chair (AC) metareview.

    For each paper you will see:
    - "human_reviews": individual ICLR reviewer assessments
    - "metareview": the AC's decision comment summarizing why the paper was
      accepted/rejected
    - "agent_review": the AI-generated review to evaluate

    EVALUATION PROTOCOL -- for EACH dimension, you MUST:
    1. QUOTE the relevant text from human reviews / metareview (use "..." for
       excerpts)
    2. QUOTE the relevant text from the agent review
    3. Explain your reasoning comparing the two
    4. Assign a score 1-5

    DIMENSIONS:

    1. contribution_identification (1-5):
       What is the primary contribution according to human reviewers?
       Did the agent identify the same core contribution?
       - 5: Agent identifies the exact same core contribution with similar framing
       - 4: Same contribution but framed differently or missing a key aspect
       - 3: Related but focuses on a secondary contribution instead of the primary
            one
       - 2: Vaguely related to the actual contribution
       - 1: Completely different contribution identified, or contribution not
            discussed

    2. top_weakness_identification (1-5):
       What is the #1 weakness from human reviews? (The one most reviewers agree
       on, or that the AC highlights in the metareview.)
       Did the agent identify this specific weakness?
       - 5: Agent identifies the same top weakness with comparable specificity and
            detail
       - 4: Identifies the weakness but less precisely, or conflates it with other
            issues
       - 3: Mentions the general area of concern but misses the specific technical
            point
       - 2: Identifies only minor or tangential weaknesses, misses the critical one
            entirely
       - 1: No meaningful overlap with human-identified weaknesses

    3. novel_concerns (1-5):
       List any concerns the agent raises that are NOT present in human reviews or
       metareview.
       Are these valid and substantive?
       - 5: Raises 2+ valid, substantive technical concerns that humans missed
       - 4: Raises 1 valid substantive concern that humans missed
       - 3: Raises concerns that are valid but minor, obvious, or not very useful
       - 2: Novel concerns are mostly invalid, confused, or based on
            misunderstanding the paper
       - 1: No novel concerns raised, or all novel points are factually wrong

    4. overall_review_quality (1-5):
       Holistic quality considering: summary accuracy, weakness depth, strength
       identification, constructiveness, specificity.
       - 5: Comparable to a good human ICLR reviewer -- specific, evidence-based,
            actionable feedback
       - 4: Good review with paper-specific points but missing some depth or nuance
       - 3: Adequate but generic -- observations could apply to many papers in the
            area
       - 2: Superficial, repetitive, or contains factual errors about the paper's
            content
       - 1: Useless -- generic template text, wrong paper content, or incoherent
        \end{Verbatim}

\section{What is \systemfamily{} Looking At?}\label{app:signal}

    \subsection{Section-Attention Map}\label{appendix:where_signal}

    \begin{figure}[H]
        \centering
        \includegraphics[width=\linewidth]{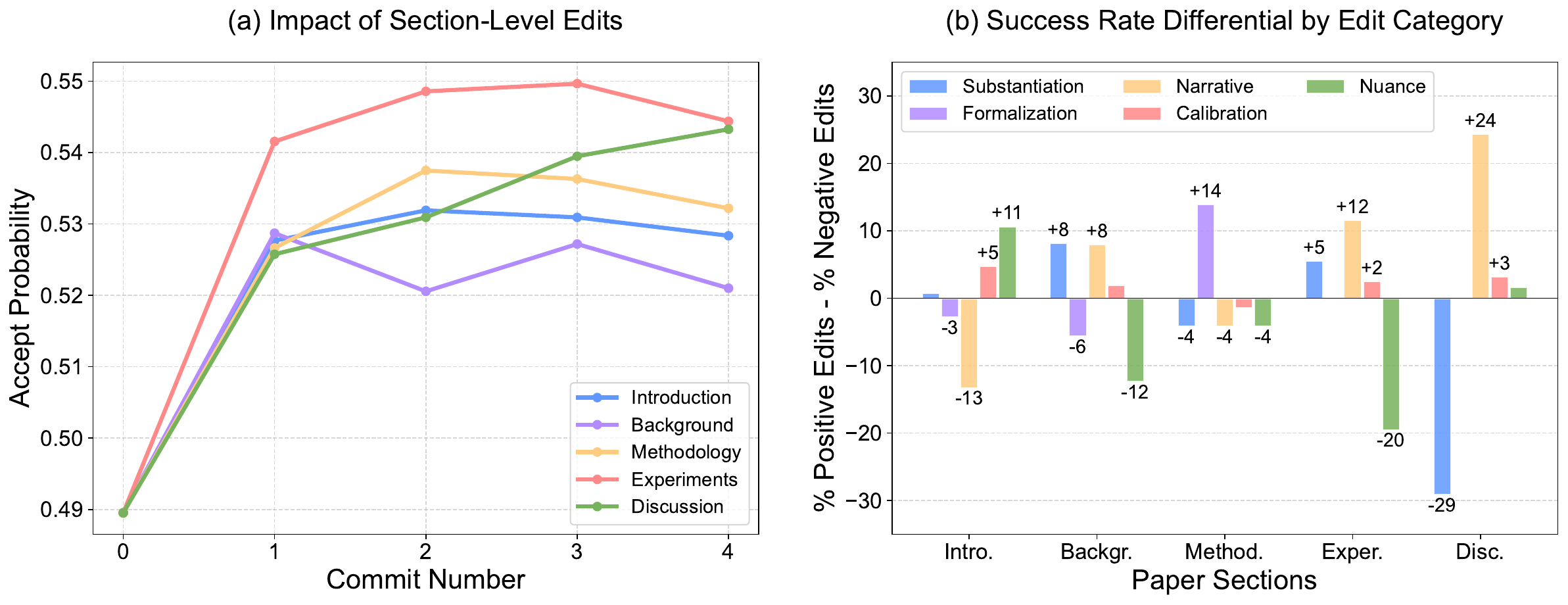}
        \caption{\textbf{Section edits.} (a) Later sections that receive more attention also show a larger maximum observed change in $p_{\mathrm{accept}}$ across four Git commits. (b) The most effective edit type depends on the section: narrative helps Experiments and Discussion, formalization helps Methodology, and nuance helps the Introduction. Details in \appref{appendix:interpretability}.}
        \label{fig:sec_edit}
        \vspace{-0.5em}
    \end{figure}
\begin{figure}[H]
        \centering
        \includegraphics[width=\columnwidth]{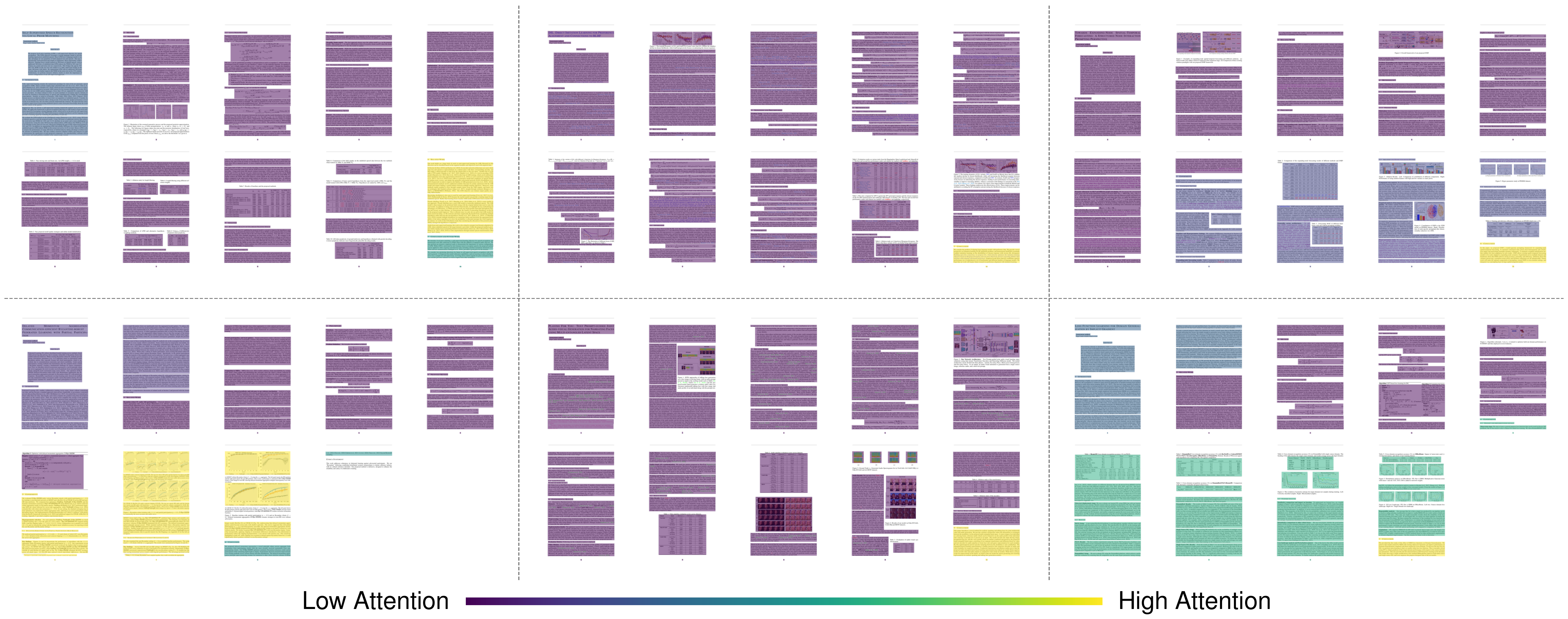}
        \caption{\textbf{Full-paper attention heatmaps.} Example attention overlay for the SFT vision model. Regions highlighted in lighter colors receive more of the final decision token's last-layer attention mass. The model concentrates most strongly on technical and evaluative material in the latter half of the paper rather than on the opening framing alone.}
        \label{fig:attn_vis}
    \end{figure}

    \begin{wraptable}[8]{r}{0.28\textwidth}
        \centering
        \vspace{-0.8em}
        \footnotesize
        \setlength{\tabcolsep}{4pt}
        \renewcommand{\arraystretch}{1.1}
        \begin{tabular}{@{}lc@{}}
            \toprule
            \textbf{Section} & \textbf{Attention} \\
            \midrule
            Discussion & 37\% \\
            Experiments & 28\% \\
            Methodology & 15\% \\
            Introduction & 9\% \\
            Background & 7\% \\
            \bottomrule
        \end{tabular}
        \caption{\textbf{Section attention.}}
        \label{tab:section_attention}
        \vspace{-0.5em}
    \end{wraptable}
    We attribute the decision of \systemname{} to specific input features. Using metadata from MinerU, we divide each paper into 5 sections: Introduction, Background, Methodology, Experiments (including Results), and Discussion (including Conclusion). Using the final decision token's last-layer attention as a descriptive map (described in \appref{appendix:interpretability}), we observe a clear hierarchical preference as seen in \autoref{tab:section_attention}. Generally, attention increases later in the paper. However, the introduction is weighed more heavily than the background. We visualize examples in \autoref{fig:attn_vis}.

    \textbf{Edit causality.}
    To verify that this pattern represents the importance of each section, we perform a causal experiment. We use Codex (\texttt{gpt-5.5}, medium reasoning) to write independent, section-level edits to the markdown paper copy of 30 accepted '26 papers with low presentation scores that \systemname{}-T scores as borderline (accept probability between 0.45 and 0.50). Each Codex instance edits one section at a time, allowing up to four sets of edits to be saved as Git commits, with feedback from \systemname{}-T received after each commit. These changes are constrained to presentation, framing, and style. We analyze the $\sim$600 commits made. Details in \appref{appendix:interpretability}.

    Our causal analysis matches the attention pattern: later sections, especially Experiments and Discussion, show the largest maximum observed changes in $p_{\mathrm{accept}}$ in Figure~\ref{fig:sec_edit}a. We find that the most effective edit type is section-dependent (Figure~\ref{fig:sec_edit}b). Formalization helps most in methodology, narrative improvements help most in the discussion and experiments, whereas substantiation in the conclusion typically hurts acceptance probability. Notably, these edits do not continuously increase $p_{\mathrm{accept}}$, suggesting that \systemtextname{} will only allow style edits to increase $p_{\mathrm{accept}}$ to a certain threshold outside the 0.7 high-confidence region.

    \subsection{Interpretability Details}\label{appendix:interpretability}

    \textbf{Attention score definition.}
    Following \citet{li2026reinforced}, we use the last-layer attention of the final decision token to summarize which paper tokens the model consults most strongly. Let $S = (x_1, \dots, x_T)$ denote the full context, where the paper occupies tokens $x_{P+1}, \dots, x_T$. If $\tau$ is the generated accept/reject token and $\alpha_{\tau,i}$ is the average attention weight from $\tau$ to token $i$ across final-layer heads, then the normalized paper-token attention score is
    \[
        a_{\tau,i} = \frac{\alpha_{\tau,i}}{\sum_{j \in \mathcal{J}} \alpha_{\tau,j}},
    \]
    where $\mathcal{J}$ indexes the paper tokens only. This gives a probability distribution over the paper conditioned on the final decision token. We aggregate these scores by section using text bbox metadata from MinerU, then summarize the resulting distribution by section medians in the main-text table.

    \textbf{Perturbation reasoning.}
    The attention map is descriptive rather than causal. A section may receive high last-layer attention because it carries decisive evidence, but it could also inherit importance from earlier-layer computation or from interactions across sections. We therefore complement attention with direct perturbations: if editing one section consistently changes \systemtext{}'s predicted acceptance probability while all other sections remain fixed, that section is more plausibly part of the model's effective decision boundary.

    \textbf{Perturbation setup.}
    We select 30 borderline papers whose baseline accept probabilities lie in the $0.45$--$0.50$ range and spawn 5 section-specific agents per paper, one for each section family: Introduction, Background / Related Work, Methodology, Experiments / Results, and Discussion / Conclusion. The editing agent is Codex on \texttt{gpt-5.5} with medium reasoning and no web search. Each agent is restricted to one section and is not allowed to change any numerical results, but it may improve framing, exposition, ordering, and technical detail within that section. Each commit the Codex agents submit is scored by our fine-tuned \systemtext{}, evaluated greedily at temperature $0$ with logprobs.

    \begin{table}[H]
        \centering
        \footnotesize
        \renewcommand{\arraystretch}{1.3}
        \begin{tabular}{@{} l p{10cm} @{}}
            \toprule
            \textbf{Category} & \textbf{Description} \\
            \midrule
            \textbf{Substantiation} & Introducing concrete evidence from tables and figures into the paper text. \\
            \textbf{Formalization} & Formalizing notation and stating implicit equations, explicit definitions, or proof structures. \\
            \textbf{Narrative} & Restructuring sections to reframe motivation or alter content flow. \\
            \textbf{Calibration} & Ensure that sentiment matches the scope of results by adjusting overclaims, tempering language, and acknowledging weaknesses. \\
            \textbf{Nuance} & Clarifying ambiguous phrasing, defining terms, or improving precision of descriptions. \\
            \bottomrule
        \end{tabular}
        \caption{\textbf{Paper revision strategy taxonomy.} We define the edit categories used to label section-level perturbations.}
        \label{tab:revision_strategies}
    \end{table}

\subsection{Edit Analysis}\label{appendix:edit_analysis}

\textbf{Section attention trends.}
As noted in the main text, the section-level attention averages show a clear ordering that prioritizes later sections. The example heatmap in \autoref{fig:attn_vis} illustrates this qualitative pattern, with attention clustering tightly around technical claims, result tables, and synthesis paragraphs, rather than the opening overview. While previous literature suggests that attention analysis alone may not definitively determine the importance of input tokens, our causal edit analysis demonstrates that the increase in acceptance likelihood follows this exact same ordering.

\textbf{Perturbation protocol.}
For our section-level edit interventions, we select 30 accepted papers from ICLR '26 that specifically have low presentation scores. This ensures the underlying content is of high quality, isolating the impact of stylistic and structural improvements on $p_{\mathrm{accept}}$.

We use Codex with \texttt{gpt-5.5} under a medium reasoning setting to generate edits \citep{chen2021evaluating}. The editing agent is restricted to a single section per trial, with unlimited read and edit tool calls. Crucially, the agent is explicitly forbidden from modifying any numerical results, guaranteeing that changes are strictly limited to framing, clarity, and style.

Once the agent is satisfied, it commits the updated text with a summary statement. This process is repeated 3 more times using a naive agent given the same prompt, without knowledge of the \systemname{} output. After all commits are generated, we use \systemname{} to grade each iteration.

\textbf{Impact of edits by section.}
The rankings of the sections by $\Delta p_{\mathrm{accept}}$ directly reflect the initial attention distribution, with the majority of the score improvement coming from the first edit commit. Edits to the Experiments / Results and Discussion / Conclusion yield the largest gains (5.4--6.1\%), followed by Methodology (4.7\%), and smaller improvements for the Introduction and Background (3.9--4.2\%). Because the maximum difference between final scores across the sections differs by only 2\%, we conclude that the attention weighting reflects a consistent, albeit small, effect size regarding which sections the model prioritizes. We also hypothesize that the effect size is partially due to the authors' allocation of time. The introductions of papers are typically written quite well, as they provide the first impression for a reviewer or reader. Therefore, there may be fewer style changes that may improve the paper. On the other hand, the results themselves are typically more important than the style of reporting them, so there may be more stylistic things to update in later sections.

\textbf{Optimal edit strategies.}
We also examine the types of edits Codex makes in each section and which types our model favors. Using the commit messages, we classify each edit into specific categories (detailed in Table~\ref{tab:revision_strategies}). As illustrated in \autoref{fig:sec_edit}b, the success rate of these edits is highly section-dependent:

\begin{itemize}
    \item \textbf{Introduction:} Benefits most from adding Nuance (+11\%) and Calibration (+5\%). Adjusting the Narrative here decreases the score (-13\%).
    \item \textbf{Background / Related Work:} Improves most from Substantiation (+8\%) and Narrative framing (+8\%), but is penalized by added Nuance (-12\%).
    \item \textbf{Methodology:} Only Formalization is effective (+14\%), whereas all other edit types negatively impact the score.
    \item \textbf{Experiments and Discussion:} Narrative edits yield the highest positive returns in both the Experiments (+12\%) and Discussion (+24\%) sections. Crucially, attempting to introduce new Substantiation into the Discussion serves as the most detrimental edit overall (-29\%), suggesting that this final section should focus on interpreting results rather than reiterating evidence.
\end{itemize}

These edit success distributions reflect general paper guidelines, suggesting that our model has learned the overall structure of a paper.

\end{document}